\pdfoutput=1
\documentclass[11pt]{amsart}
\usepackage[margin=1in]{geometry}
\usepackage{amssymb,bm}            %
\theoremstyle{plain}
\newtheorem{proposition}{Proposition}
\usepackage{graphicx}
\usepackage{booktabs}
\usepackage{multirow}
\usepackage{xcolor}
\usepackage{caption}
\usepackage{subcaption}
\usepackage[section]{placeins}
\newcommand{\doi}[1]{\href{https://doi.org/#1}{doi:#1}}
\usepackage[colorlinks=true,linkcolor=blue!50!black,citecolor=blue!50!black,urlcolor=blue!50!black]{hyperref}

\newcommand{\R}{\mathbb{R}}
\DeclareMathOperator{\diag}{diag}
\DeclareMathOperator{\dist}{dist}
\DeclareMathOperator{\Lip}{Lip}
\DeclareMathOperator{\softplus}{softplus}
\DeclareMathOperator{\Span}{span}
\DeclareMathOperator{\RePart}{Re}
\newcommand{\bx}{\mathbf{x}}\newcommand{\by}{\mathbf{y}}\newcommand{\bc}{\mathbf{c}}\newcommand{\bd}{\mathbf{d}}
\newcommand{\bh}{\mathbf{h}}\newcommand{\bs}{\mathbf{s}}\newcommand{\bz}{\mathbf{z}}\newcommand{\bb}{\mathbf{b}}
\newcommand{\bv}{\mathbf{v}}\newcommand{\bw}{\mathbf{w}}\newcommand{\br}{\mathbf{r}}
\newcommand{\bU}{\mathbf{U}}\newcommand{\bV}{\mathbf{V}}\newcommand{\bB}{\mathbf{B}}\newcommand{\bW}{\mathbf{W}}
\newcommand{\bC}{\mathbf{C}}\newcommand{\bQ}{\mathbf{Q}}\newcommand{\bJ}{\mathbf{J}}
\newcommand{\bomega}{\bm{\omega}}\newcommand{\btheta}{\bm{\theta}}\newcommand{\balpha}{\bm{\alpha}}
\newcommand{\bbeta}{\bm{\beta}}\newcommand{\bu}{\mathbf{u}}\newcommand{\bH}{\mathbf{H}}\newcommand{\ba}{\mathbf{a}}\newcommand{\be}{\mathbf{e}}\newcommand{\bn}{\mathbf{n}}\newcommand{\bpsi}{\bm{\psi}}
\colorlet{todored}{red!70!black}\colorlet{TODORED}{red!70!black}

\graphicspath{{figures/}}

\title[Multi-layer displacement with LRNRs]{Multi-layer displacement: interpolating moving fronts
from few snapshots with low-rank neural representations}
\author{Swarup Dhar}
\address[S.~Dhar]{Department of Mathematics, Washington University in St.~Louis, One Brookings Drive, St.~Louis, MO 63130, USA}
\email{d.swarup@wustl.edu}

\author{Donsub Rim}
\address[D.~Rim]{Department of Mathematics, Washington University in St.~Louis, One Brookings Drive, St.~Louis, MO 63130, USA}
\email{rim@wustl.edu}

\author{Xiaoyue Yuan}
\address[X.~Yuan]{Department of Mathematics, Washington University in St.~Louis, One Brookings Drive, St.~Louis, MO 63130, USA}
\email{xiaoyuey@wustl.edu}
\date{\today}

\begin{document}

\begin{abstract}
Interpolating a parametrized family of fields $u(\cdot;\mu)$ from a few snapshots is difficult when its features
move.  When only a few snapshots are available, commonly used classical and neural methods, such as linear reduced
bases, Hyper-SIRENs and DeepONets, interpolate such families by cross-fading.
We show that low-rank neural representations (LRNRs) instead interpolate by multi-layer displacement,
that is, by moving the level sets of their units, in the first layer and, where needed, in deeper layers.  A LRNR is a network whose weights and biases are low-rank combinations
of fixed bases, with coefficients produced by a hypernetwork of $\mu$.  As
a diagnostic, we define a fade ratio that compares an
interpolant with the cross-fade of its neighboring snapshots.  For seven one- and two-dimensional families
(translating, splitting, spiraling and oscillatory features, and a control that does not move), we use
it to determine the smallest number of snapshots $N^\ast$ from which a LRNR displaces rather than fades.  At $N^\ast$
the LRNR is as accurate as or more accurate than the most accurate of eleven neural and classical baselines on every
family with moving features, except two oscillatory families whose carriers dilate smoothly (rings and a wave packet),
on which registration and sPOD given the exact shifts are more accurate.  On the non-oscillatory families the LRNR
also has the smallest worst-case error over these families.  On a linear wave and a Burgers
family, where the fade ratio is not informative, we compare held-out errors only, and the LRNR has the lowest
mean held-out error on both.  In the trained models the latent
path is nearly straight, and on translating families 89--98\% of the change with $\mu$ comes from the bias
coefficients, most of it from the first layer.
\end{abstract}

\maketitle

\section{Introduction}\label{sec:intro}

Many parametrized problems produce families of fields whose dominant variation is the motion of fronts: a pulse that
is advected or splits, a vortex that moves with the flow, a wave that propagates.  Linear reduced-order models represent the family in a
fixed basis, $u(\bx;\mu) \approx \sum_k c_k(\mu)\phi_k(\bx)$, and need many modes to represent such families (the slow decay of
the Kolmogorov $n$-width~\cite{kolmogorov1936nwidth,pinkus1985nwidths}).  Interpolation in $\mu$ between a few snapshots shows this problem directly.  Since the
interpolant is a $\mu$-dependent combination of fixed spatial patterns, the cheapest way to pass through the snapshots
is to blend them, so that a feature at two positions appears as two copies, one fading out and one fading in.

Interpolants that move rather than blend have received significant attention over the years.  Displacement interpolation~\cite{mccann1997convexity}
moves mass along optimal-transport geodesics; it has been used to interpolate distributions and
images~\cite{bonneel2011displacement} and, by transport reversal and monotone rearrangement, solutions of hyperbolic
problems for model reduction~\cite{rim2018transport,rim2018displacement}.  Optimal transport itself has a large
literature~\cite{villani2009ot,santambrogio2015ot,peyre2019cot,cuturi2013sinkhorn}.  Advection modes~\cite{iollo2014advection}
use optimal-transport maps between snapshots to advect a reduced basis.  Transformed snapshot
interpolation~\cite{welper2017transformed,welper2020highres} composes the snapshots with parameter-dependent
transformations of the spatial domain that align their features, such as jumps, before interpolating them in the
parameter.  Front transport reduction~\cite{krah2023ftr} writes the field as a fixed front profile applied to a
level-set function and approximates the level-set function in a low-rank linear basis, so that the fronts move while
the basis stays linear.  For signed fields and for
waves, mass is not the right quantity to transport,  but note that the HV geometry~\cite{han2024hv} removes
this restriction for signals.  Its Riemannian metric on the space of signals measures horizontal deformations, as the
Wasserstein metric does, together with vertical deformations of the values, and it admits signed and discontinuous
signals; its geodesics interpolate between signals.  In the terms of this paper, a horizontal deformation displaces
fronts and a vertical deformation blends values.  For waves, plane waves are the natural description.  The Radon transform~\cite{natterer2001ct} turns a translation into a shift of every plane-wave
component, $\mathcal R[f(\cdot - \bc)](\theta,s) = \mathcal Rf(\theta, s-\bomega_\theta\cdot \bc)$, and the wave equation into
a one-dimensional wave equation in $s$ for every direction.  In this representation, the Lax--Phillips translation
representation~\cite{laxphillips1967}, wave propagation is a translation of plane-wave fronts.  Dimensional
splitting and displacement interpolation of hyperbolic problems follow from it~\cite{rim2018radon,rim2018displacement}.
Sliced and Radon optimal transport~\cite{bonneel2015sliced,kolouri2016radoncdt}, reduced-order models based on
it~\cite{long2023radoncdt}, and frames whose elements are displaced along the Hamiltonian flow of a wave propagator, such as
curvelets~\cite{candes2005curvelet} and wave atoms~\cite{demanet2007waveatoms}, rely on the same idea with fixed
elements.  We call interpolation of this kind \emph{displacement}.  The field is
interpolated by displacing the fronts of its components (the level sets $\{\bomega\cdot\bx = b\}$ of ridge functions
$\rho(\bomega\cdot\bx - b)$, displaced by changing $b$) rather than by blending its values.  It covers translation,
splitting and wave propagation alike.  The word ``transport'' is too narrow for these cases, because a wave is a
superposition of displaced plane-wave fronts rather than a field moved by a velocity.

We distinguish three levels.  A
\emph{linear} interpolant changes only the coefficients of fixed elements and cross-fades (proper orthogonal
decomposition (POD)~\cite{ly2001pod}, DeepONet~\cite{lu2021deeponet}).  This level corresponds to a static low-rank
neural representation (LRNR)~\cite{cho2023hyper,cho2025fastlrnr,rim2026entropy,cho2025hyperbolic}, one with all its internal coefficients frozen, so that only the outer coefficients
change with $\mu$; it is a two-layer LRNR whose inner layer is frozen (\S\ref{sec:snapcount}).  \emph{Single-layer
displacement} moves planar fronts of fixed shape: the Radon-domain methods above, and transported
subspaces, which are two-layer LRNRs (\S\ref{sec:invbias}).  A LRNR with several layers adds a third level,
\emph{multi-layer displacement}: its deeper units are ridge functions of learned features, their fronts are curved,
and changing their biases and weights displaces and reshapes these curved fronts.  On the families that split
or translate, the trained models move almost only their first-layer fronts (planar in the input features $\bpsi$); a carrier whose wavelength changes
also needs the weights, and a winding arc the deeper layers as well (\S\ref{sec:biases}).

LRNRs
are implicit neural representations (INRs), also called coordinate networks~\cite{sitzmann2020siren}, whose weight
matrices are low-rank, with fixed bases and coefficients produced by a hypernetwork of $\mu$.  The spatial patterns available to the network then change with $\mu$ through its
nonlinearities.  LRNRs extend a classical nonlinear reduction method, transported
subspaces~\cite{rim2023transported}, in which the family is represented by a few fixed profiles that are moved by
$\mu$-dependent transports and then combined linearly.  This connection is made in~\cite{rim2026entropy} through what is called
the \emph{inverse-bias trick} (\S\ref{sec:invbias}): a transported subspace, whose members contain the inverse of a
transport map, is rewritten without the inverse, with the transport acting on the biases of a two-layer network, and is
then exactly a two-layer LRNR whose coefficients depend affinely on the transported subspace's coefficients.  Every
transported subspace is therefore a special case of a LRNR, and the LRNR drops the constraints the classical
construction needs (monotone, invertible transports).  The same construction gives LRNR approximations of classical
and entropy solutions of scalar conservation laws whose coefficients are linear or constant in time~\cite{rim2026entropy}.  Where
transported subspaces need the transports to be specified or computed separately, the LRNR represents the profiles and
their motion within one network, trained from data~\cite{cho2025hyperbolic}.

The theory of~\cite{rim2026entropy} identifies a low-dimensional structure underlying hyperbolic problems: their solutions have LRNR
representations of small rank whose coefficients are simple (linear or constant) functions of time and are
switched on and off by the activation.  This is an existence result; it does not imply that
such a representation can be found by gradient descent.  It was shown empirically in~\cite{cho2025hyperbolic} that LRNRs of hyperbolic
wave dynamics can be learned from data.  We posit that LRNR interpolation exploits this structure: a coefficient
path that is simple in $\mu$ suffices to move features, so the model can be determined by few snapshots and does not
need to blend them.  In the LRNRs trained here all $\mu$-dependence passes through 400 coefficients in each layer of the network
(2000 in total, against 2.4M fixed basis parameters); about half stay inactive for every $\mu$, and the rest are affine in $\mu$ up to a switch on or off.
The architecture, however, has no component specific to hyperbolic problems (no characteristics and no prescribed
motion).  Because of this generality, the same kind of structure is learned from phenomena that the theory does
not cover: features that split into copies, follow a spiral at varying speed, or have an oscillation whose frequency
changes (\S\ref{sec:latent}, \S\ref{sec:biases}).  The results of this paper therefore support two points.  First, the LRNR
displaces fronts without being given the motion.  Unlike shifted POD (sPOD)~\cite{reiss2018spod},
registration~\cite{taddei2020registration} and optimal transport (OT), it needs no shifts, maps or matched features, and it
represents the whole family as one smooth model.  Classical methods that are given more can be more accurate on
average, and we report where, but each of them has large errors on some family, whereas the LRNR has the smallest
worst-case error over all the non-oscillatory families.  Second, one approach covers all these kinds of motion: the same
architecture, trained the same way, displaces the fronts of translating, splitting and deforming features, of
oscillatory wave packets and of a linear wave, and for hyperbolic problems the theory guarantees that such a representation
exists.  Oscillatory packets are more difficult, because the phase of their carrier is undersampled at small $N$ and
dominates the relative errors and fade ratios, so we report them separately (\S\ref{sec:osc2d}).  On a linear wave and on shocks (a
2D Burgers family) the fade ratio is not informative: the wave changes in amplitude and shape as well as in position,
and on the Burgers family the fade ratio cannot separate a fade from grid-scale smearing of the shock.  We report these
two families as a third separate group, by held-out error against the neural baselines (Appendix~\ref{app:nofade}).  This paper studies whether, and from how many snapshots, such a model
interpolates these families by displacing their fronts rather than fading them.  Its contributions are:
\begin{itemize}
  \item the notion of multi-layer displacement, which treats translation, splitting and wave propagation in
        one framework and distinguishes linear interpolants, single-layer displacement and multi-layer (learned,
        curved-front) displacement
        (\S\ref{sec:intro}, \S\ref{sec:invbias});
  \item a fade ratio and a pass rule that measure whether an interpolant fades, and a snapshot-count transition study
        on four families, including a control whose fronts do not move, and, with an adapted rule, on three oscillatory
        families (\S\ref{sec:metrics}, \S\ref{sec:transition}, \S\ref{sec:osc2d});
  \item studies of snapshot placement, training length, cell size, and regularization (\S\ref{sec:spiral}--\ref{sec:ablation});
  \item parameter-matched comparisons with Hyper-SIREN (a hypernetwork that produces the weights of a SIREN
        INR~\cite{sitzmann2020siren}) and DeepONet, each trained on whole snapshots and on the LRNR's cells, and with
        a Fourier neural operator (FNO)~\cite{li2021fno} and a convolutional autoencoder (CAE)~\cite{lee2020cae}
        (\S\ref{sec:baselines});
  \item a comparison with classical interpolants (POD, registration, sPOD, optimal transport and dynamic mode
        decomposition (DMD)~\cite{schmid2010dmd}) and a discussion of why such comparisons are not like-for-like:
        sPOD needs the transport (its shift parameters) to be provided rather than learned, registration and optimal
        transport are based on independent two-snapshot maps, and DMD assumes linear dynamics in $\mu$
        (\S\ref{sec:classical});
  \item two analyses of the trained interpolants as dynamical systems in $\mu$ (\S\ref{sec:dynamics}).
\end{itemize}

\section{Test families}\label{sec:families}

\begin{figure}[!htbp]
  \centering
  \includegraphics[width=\linewidth]{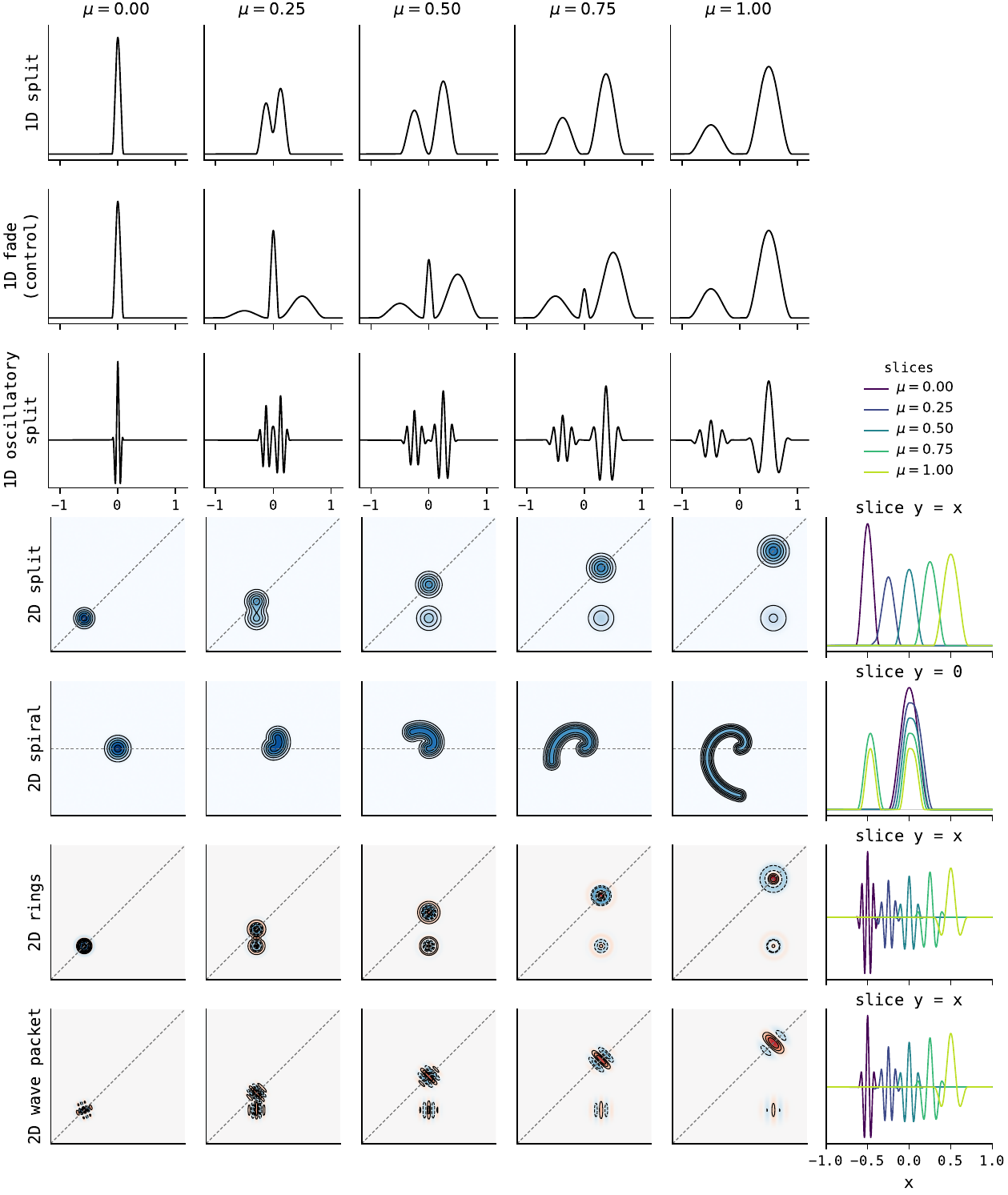}
  \caption{The test families at $\mu = 0, 0.25, 0.5, 0.75, 1$.  2D fields: blue = non-negative, red/blue = signed; contours at
  0.1, 0.3, ..., 0.9 of the panel's maximum ($\pm0.25, \pm0.5, \pm0.75$ for signed fields, negative dashed).  Right column:
  the five 2D fields along the dashed line, overlaid (the diagonal $y = x$: the path of the upper copy for the split
  families; $y = 0$ for the spiral).  The 2D wave and 2D Burgers families are shown in Figure~\ref{fig:familiesnofade}.}
  \label{fig:families}
\end{figure}

\begin{figure}[!htbp]
  \centering
  \includegraphics[width=\linewidth]{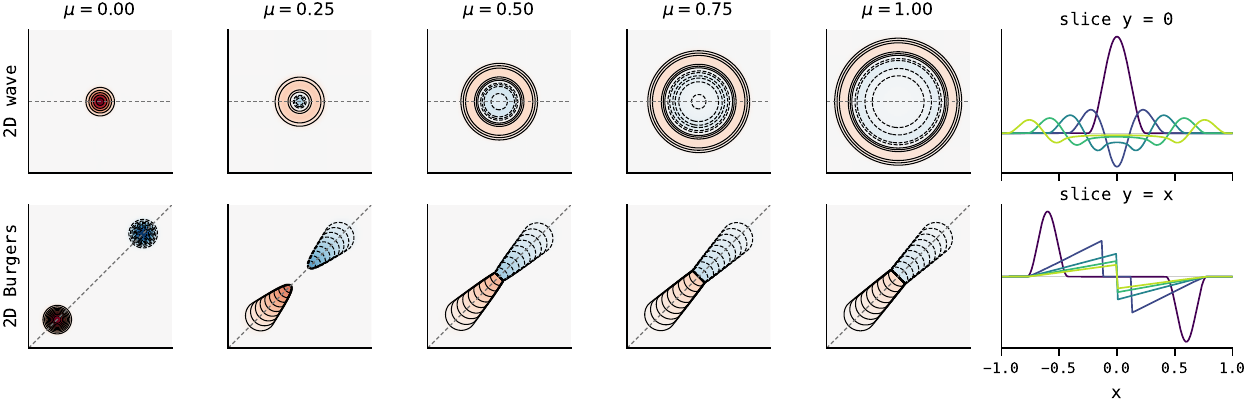}
  \caption{The 2D wave and 2D Burgers families at $\mu = 0, 0.25, 0.5, 0.75, 1$, as in Figure~\ref{fig:families}
  (Burgers: contours at $\pm0.1, 0.2, ..., 0.9$ of the panel's maximum, negative dashed).  Right column: the five
  fields along $y = 0$ (wave) and along the diagonal $y = x$, across the standing shock (Burgers).}
  \label{fig:familiesnofade}
\end{figure}

All families are constructed from the cosine hump $h(d; r) = \tfrac12\bigl(1+\cos(\pi d/r)\bigr)\mathbf{1}_{\{d<r\}}$ and are
known in closed form, except the 2D wave and 2D Burgers families, whose truths are reference solutions on grids two and
eight times finer than the data; every interpolant can be scored against the truth at any $\mu\in[0,1]$ (Figures~\ref{fig:families}
and~\ref{fig:familiesnofade}).

\begin{itemize}
  \item \emph{1D split.} On $x\in[-2,2]$ (2000 cell centers), a hump at the origin splits into two copies that move
        apart while they widen and change height:
        \[ u = (0.5-0.25\mu)h(|x+0.5\mu|; r) + (0.5+0.25\mu)h(|x-0.5\mu|; r), \quad r = 0.1+0.3\mu. \]
  \item \emph{1D fade (control).} The cross-fade $u = (1-\mu)u_0 + \mu u_1$ of the end states $u_0$, $u_1$ of the
        1D split family.  Its truth is a fade, so a model that fades is correct here.  This family checks that the
        metrics do not reward motion for its own sake.
  \item \emph{1D oscillatory split.} The 1D split family with a carrier $\cos(k\pi(x-c))$ on each copy, centered at
        the copy's center $c$, so that a copy is $h(|x-c|; r)\cos(k\pi(x-c))$.  The frequencies change with $\mu$:
        $k_L = 20-10\mu$ and $k_R = 20-15\mu$.  The crests must move with their copy.
  \item \emph{2D split.} On $[-1,1]^2$ ($160\times160$ cell centers), a hump at $\bc_0=(-0.5,-0.5)$ splits into two
        copies that move to $(0.5,0.5)$ and $(0.5,-0.5)$.  The heights are $0.5\pm0.25\mu$ and the radius is
        $0.2+0.1\mu$.
  \item \emph{2D spiral.} A hump at the center stretches into an arc that winds up as $\mu$ grows:
        \[ u = a(\mu)h\bigl(\dist(\bx, C_\mu); w(\mu)\bigr),\quad
           C_\mu = \bigl\{0.7\mu s(\cos 1.5\pi\mu s, \sin 1.5\pi\mu s): s\in[0,1]\bigr\}, \]
        with $w = 0.25-0.13\mu$ and $a = 1-0.5\mu$.  The tail of the arc moves fastest, near $\mu=1$.
  \item \emph{2D rings and 2D wave packet.} The 2D split family with an oscillatory carrier on each copy: radial
        rings $\cos(k\pi|\bx-\bc|)$, or a plane wave $\cos(k\pi (\bx-\bc)\cdot \bd)$ whose crests are perpendicular to
        the direction of travel $\bd$ of the copy.  The frequencies are $k_U = 20-15\mu$ and $k_L = 20-10\mu$
        (\S\ref{sec:osc2d}).
  \item \emph{2D wave.} The wave equation $u_{tt} = \Delta u$ on $[-1,1]^2$ up to $T = 0.7$, with $\mu = t/T$, initial
        data $u(\bx,0) = h(|\bx|; 0.25)$ and $u_t(\bx,0) = 0$.  The hump splits into an outgoing ring, whose amplitude
        decays like $|\bx|^{-1/2}$, and a negative wake behind it, so the data are signed.  The outgoing front reaches
        $|\bx| = 0.95$ at $t = T$, so the boundary does not affect the solution.  The truth is computed spectrally,
        exactly in time, on a grid twice as fine as the data grid.
  \item \emph{2D Burgers.} The scalar Burgers equation $q_t + (q^2/2)_x + (q^2/2)_y = 0$ on $[-1,1]^2$ up to $T = 4$,
        with $\mu = t/T$ and initial data $q(\bx,0) = h(|\bx-\bc_0|; 0.25) - h(|\bx+\bc_0|; 0.25)$,
        $\bc_0 = (-0.6,-0.6)$: a positive hump at $\bc_0$ and a negative one at $-\bc_0$.  The humps move toward each
        other along the diagonal, each steepening into a shock at its leading edge ($t\approx0.3$) with a
        rarefaction behind it.  They collide at the center ($t\approx1.3$), and a standing shock forms along
        $x+y=0$; its jump is largest near $t=2$ (0.69) and decays to 0.36 at $t=4$.  The support stays inside
        $[-0.85,0.85]^2$.  The truth is computed by finite volumes on a $1280\times1280$ grid
        (MUSCL--Hancock~\cite{vanleer1979muscl,toro2009riemann} with minmod slopes, the exact Godunov
        flux~\cite{godunov1959}, Strang splitting~\cite{strang1968splitting}, CFL $\le0.45$), block-averaged to the
        data grid at 601 equally spaced times and interpolated linearly in time between them.
\end{itemize}

Every family except the fade control has fronts that move (they translate, split, wind up or propagate), and we call
these the \emph{moving-front families}; the fade control is the one whose fronts stay in place.
The last two families, the 2D wave and the 2D Burgers family, are reported separately in Appendix~\ref{app:nofade}:
on them the fade ratio is not informative, and we compare held-out errors only.  A dataset is $N$ snapshots at evenly spaced $\mu_k = k/(N-1)$ unless stated otherwise; each snapshot is the field on
the full grid.

\section{Low-rank neural representation}\label{sec:model}

\subsubsection*{Neural net with low-rank parametrized weights and biases}  We use the LRNR architecture of~\cite{cho2023hyper,cho2025fastlrnr}.  It is a fully connected network on input features $\bpsi(\bx)$ (monomials of degree
2, or Chebyshev polynomials of degree 9 for the oscillatory families) with $L=5$ layers of width $M=1000$, whose
weight matrices and biases are rank-$r$ combinations of fixed bases,
\begin{equation}
  \bh_{\ell+1} = \sigma\left(\bU_\ell\diag\bigl(\bs^{W}_\ell(\mu)\bigr)\bV_\ell^\top \bh_\ell
             + \bB_\ell \bs^{b}_\ell(\mu)\right),\quad \ell = 0,...,L-1,
  \label{eq:lrnr}
\end{equation}
with $\bU_\ell, \bV_\ell, \bB_\ell$ of rank $r=200$, $\sigma=\tanh$ on hidden layers and the identity on the output.  The
bases hold about $2.4$M parameters and do not depend on $\mu$.

\subsubsection*{Hypernetwork}  All $\mu$-dependence is in the $2rL = 2000$ coefficients $\bs(\mu)$, produced by a
small hypernetwork,
\begin{equation}
  \bz(\mu) = \tanh\bigl(\bH_2\tanh(\bH_1\mu+\ba_1)+\ba_2\bigr)\in\R^4,\quad \bs(\mu) = \softplus_{10}\bigl(\bC\bz(\mu)+\ba_3\bigr),
  \label{eq:hyper}
\end{equation}
where $\softplus_{10}(t) = \tfrac{1}{10}\log(1+e^{10t})$.  The trained $\bz(\mu)$ is almost a straight line, as shown
in Section~\ref{sec:latent}.

\subsubsection*{Relation to transported subspaces: the inverse-bias trick}\label{sec:invbias}
The following is from~\cite[\S4]{rim2026entropy}, specialized to one space dimension.  A transported
subspace~\cite{rim2023transported} is a family of functions $\bar h = f\circ g^{-1}$, where $f = \sum_i \beta_i\psi_i$ is
a combination of fixed profiles and $g = \sum_i \alpha_i\phi_i$ is a monotone transport map, a combination of fixed
maps; the coefficient vectors $(\balpha,\bbeta)$ play the role of $\bs(\mu)$.  The inverse $g^{-1}$ makes the construction
awkward: $g$ has to stay monotone and invertible, so the coefficients $\balpha$ are constrained, and $g^{-1}$ has to be
computed.  If the profiles are written as superpositions of jumps, $f(x) = c_0 + \sum_k c_k\rho(x - x_k)$ with $\rho$
the Heaviside function, monotonicity of $g$ gives $\rho(g^{-1}(x) - x_k) = \rho(x - g(x_k))$, so
\begin{equation}
  f\circ g^{-1}(x) = c_0 + \sum_k c_k\rho\bigl(g^{-1}(x) - x_k\bigr) = c_0 + \sum_k c_k\rho\bigl(x - g(x_k)\bigr).
  \label{eq:invbias}
\end{equation}
The inverse has disappeared: the transport now acts only through the biases $-g(x_k)$, which are linear in
$\balpha$, and the profile only through the outer weights $c_k$, linear in $\bbeta$.  With $\rho$ replaced by the
difference of two shifted ReLUs, $\rho_\varepsilon(x) = \varepsilon^{-1}[\sigma_{\rm ReLU}(x+\varepsilon/2) - \sigma_{\rm ReLU}(x-\varepsilon/2)]$,
the right-hand side of~\eqref{eq:invbias} is a two-layer network $\bw_2\cdot\sigma_{\rm ReLU}(\bw_1x+\bb_1)+b_2$ whose weights and
biases are rank-$r$ combinations of fixed vectors.  It is therefore a two-layer LRNR, with coefficients that are an affine
function of $(\balpha,\bbeta)$.  This approximates any rank-$(r-1)$ transported subspace uniformly in $L^1$~\cite[Thm.~4.5]{rim2026entropy}.
The right-hand side makes sense for every coefficient vector, invertible transport or not, so the LRNR needs neither
the monotonicity constraint nor the inverse.  In several dimensions the same identity applies to each ridge function
$\rho(\bomega\cdot\bx - b)$ of the network~\cite{rim2026entropy}.

Three points of this construction are relevant below.  (i) Moving a feature costs the LRNR a change of biases, the
$\bB_\ell \bs^b_\ell(\mu)$ terms in~\eqref{eq:lrnr}, while blending two copies costs a change of outer weights; a
linear reduced basis or DeepONet has only the second.  A change of the bias $b$ of a unit $\rho(\bomega\cdot\bx - b)$
displaces its front, the level set $\{\bomega\cdot\bx = b\}$.  A two-layer LRNR therefore performs single-layer displacement
of planar fronts (\S\ref{sec:intro}), and in a deeper LRNR the same change at layer $\ell$ displaces the curved level
sets of a learned feature.  (ii) For a transported family the required coefficients are
affine in the parameter.  For the classical solution of a scalar conservation law, a rank-3 two-layer LRNR with
coefficients linear or constant in time suffices~\cite[Thm.~4.6]{rim2026entropy}.  The trained LRNRs here have
coefficients that are affine in $\mu$ up to ReLU-like switching (\S\ref{sec:latent}), consistent with this picture.
(iii) The transport is not given to the network.  A transported subspace, like sPOD (\S\ref{sec:classical}),
needs its transport maps $g$ to be supplied or computed, whereas the LRNR's biases and their $\mu$-dependence are
trained along with everything else.  Our networks are deeper, use $\tanh$, and are fitted by gradient descent, so the
construction explains what the architecture can represent cheaply.  It does not describe what training finds.  On
the split families, the first layer's bias coefficients of the trained models produce almost all of the motion, as
shown in Section~\ref{sec:biases}.

\subsubsection*{Spatial projectors (training only)}  For spatial mini-batching each cell $k$ of the domain decomposition
(\S\ref{sec:training}) has rank-$300$ projectors $\bQ_{k,\ell}$ applied to the hidden activations of a second,
\emph{filtered} forward pass.  They are used only in the training loss (\S\ref{sec:loss}); the model's output, and every
result in this paper, uses the unfiltered pass~\eqref{eq:lrnr}.

\section{LRNR coefficient count}\label{sec:counts}

The inverse-bias trick turns every family of \S\ref{sec:families} into an explicit LRNR, so for each family an upper
bound on the LRNR \emph{coefficient count} of~\cite[Def.~3.1]{cho2025hyperbolic} can be read off.  For a network with
layer weights $\bW_\ell(\mu)$ and biases $\bb_\ell(\mu)$ we count
\begin{itemize}
  \item the \emph{ranks} $r^W_\ell = \min\{r : \bW_\ell(\mu) = \sum_{j=1}^r s_j(\mu)\bu_j\bv_j^\top \text{ for all } \mu\}$, the
        smallest number of fixed rank-one matrices whose $\mu$-dependent combinations give the weights (for weights
        that do not depend on $\mu$, the matrix rank), and $r^b_\ell = \dim\Span\{\bb_\ell(\mu)\}_\mu$,
        collected in the rank vector $\br = (r^W_0, r^b_0; r^W_1, r^b_1;...)$.  The coefficient count is
        $\|\br\|_1$, minimized over the LRNRs that reach a given approximation rate in the width~\cite{cho2025hyperbolic};
        a construction gives an upper bound.  A LRNR~\eqref{eq:lrnr} with $r\ge\max_\ell(r^W_\ell, r^b_\ell)$
        represents the network exactly, with coefficients $\bs(\mu)$;
  \item the \emph{latent dimension} $m = \dim\Span\{\btheta(\mu)-\btheta(\mu')\}_{\mu,\mu'}$ of the
        parameter vector $\btheta = (\bW_\ell, \bb_\ell)_\ell$: the number of scalar functions of $\mu$ the hypernetwork
        has to produce.  $m=1$ with $\btheta$ affine in $\mu$ means $\btheta(\mu) = \btheta_0 + \mu\btheta_1$.
\end{itemize}
Throughout, $\rho$ is the Heaviside function, which the LRNR realizes as the limit of $\rho_\varepsilon$
(\S\ref{sec:invbias}) or of a steep $\tanh$, and $H(t) = \tfrac12(1+\cos\pi t)\mathbf 1_{|t|<1}$ is the hump of
radius one, so $h(d;r) = H(d/r)$.

\begin{proposition}[transported profiles]\label{prop:transported}
Let $\Omega\subset\R^d$ be bounded and
\[ u(\bx;\mu) = \sum_{i=1}^P a_i(\mu) f_i\Bigl(\frac{\bx - \bc_i(\mu)}{r_i(\mu)}\Bigr),\quad r_i(\mu) > 0, \]
with fixed profiles $f_i$.  Suppose each $f_i$ is approximated on the bounded set
$Y = \bigcup_{i,\mu}(\Omega - \bc_i(\mu))/r_i(\mu)$ by a ridge sum
$f_i^M(\by) = \beta_i + \sum_{k=1}^M \gamma_{ik}\rho(\bomega_k\cdot \by - \eta_k)$ with
$\|f_i - f_i^M\|_{L^\infty(Y)}\le\varepsilon_i$.  Then the two-layer network
\[ u^M(\bx;\mu) = \sum_{i=1}^P a_i(\mu)\Bigl[\beta_i + \sum_{k=1}^M \gamma_{ik}
   \rho\bigl(\bomega_k\cdot \bx - \bomega_k\cdot \bc_i(\mu) - r_i(\mu)\eta_k\bigr)\Bigr] \]
satisfies, for every $\mu$,
\begin{equation}
  \|u(\cdot;\mu) - u^M(\cdot;\mu)\|_{L^\infty(\Omega)} \le \sum_{i=1}^P |a_i(\mu)|\varepsilon_i .
  \label{eq:transported}
\end{equation}
If $\bc_i$, $r_i$ and $a_i$ are affine in $\mu$, so is the whole parameter vector of $u^M$: $\br\le(d, 2; 2, 0)$, a
coefficient count of at most $d+4$, and $m = 1$.
\end{proposition}

\noindent\emph{Proof.}  $\rho$ is invariant under positive scaling, so
$\rho(\bomega_k\cdot(\bx-\bc_i)/r_i - \eta_k) = \rho(\bomega_k\cdot \bx - \bomega_k\cdot \bc_i - r_i \eta_k)$, and
$u^M(\bx;\mu) = \sum_i a_i f_i^M((\bx-\bc_i)/r_i)$ with $(\bx-\bc_i)/r_i\in Y$, which gives~\eqref{eq:transported}.
The rank statements are read off the parameters, as detailed below; affine $\bc_i, r_i, a_i$ make every parameter
affine in $\mu$.  \hfill$\square$

\smallskip
In detail, the first-layer weights of $u^M$, $\bomega_k$, do not depend on $\mu$
($r^W_0\le d$), its biases $-(\bomega_k\cdot \bc_i + r_i \eta_k)$ depend on $\mu$ only through $\bc_i$ and $r_i$, and its
outer weights $a_i\gamma_{ik}$ and output bias $\sum_i a_i\beta_i$ only through $a_i$ ($r^W_1\le P$, $r^b_1\le1$; the
output bias can be moved into a hidden neuron with constant bias, giving $r^b_1 = 0$).

Proposition~\ref{prop:transported} is a special case of the inverse-bias construction of~\cite[\S4]{rim2026entropy},
with the affine transports $g_i(\by) = \bc_i(\mu) + r_i(\mu)\by$ applied to each ridge function; we state it with the
ranks and the coefficient count used below.

\smallskip
The accuracy is set by the width $M$ alone: the fixed bases (here $\bomega_k$, $\eta_k$, $\gamma_{ik}$) represent the
profile, and the number and form of the coefficients do not change as $M\to\infty$.  In this respect the LRNR differs
from a linear method, whose number of coefficients must grow as the accuracy improves: the snapshots of the 1D
split need 15 POD modes for 1\% error, and more modes for smaller errors (Table~\ref{tab:counts}).

\subsubsection*{The split families and the fade control}  The 1D and 2D split families are two copies of the hump with
centers, common radius and heights all affine in $\mu$; the fade control is three humps at fixed positions with
amplitudes $1-\mu$, $\mu/4$, $3\mu/4$.  Proposition~\ref{prop:transported} gives $m = 1$ for all three, with ranks
at most 2.  In 1D the profile is approximated by a staircase of $M$ steps, $\|H - H^M\|_{L^\infty}\le\pi/(2M)$.  In 2D, the
radial hump has an exact plane-wave decomposition
\begin{equation}
  H(|\by|) = \int_0^\pi Q\bigl(\bomega_\theta\cdot \by\bigr)d\theta,\quad \bomega_\theta = (\cos\theta,\sin\theta),
  \label{eq:planewave}
\end{equation}
with $Q$ the ramp-filtered Radon transform of $H(|\cdot|)$ (the filtered back-projection formula); a quadrature in
$\theta$ and a staircase of $Q$ give the ridge sum.

\begin{proposition}[linear wave~{\cite[Thm.~3.2]{cho2025hyperbolic}}]\label{prop:wave}
Let $u$ solve $u_{tt} = \Delta u$ in $\R^2$ with $u(\cdot,0) = u_0$, $u_t(\cdot,0) = 0$, and let
$u_0(\by) = \int_0^\pi Q_\theta(\bomega_\theta\cdot \by)d\theta$.  Then
\[ u(\bx,t) = \int_0^\pi \tfrac12\bigl[Q_\theta(\bomega_\theta\cdot \bx - t) + Q_\theta(\bomega_\theta\cdot \bx + t)\bigr]d\theta . \]
Discretized as above, $t$ appears only in the first-layer biases, additively: $\br = (2, 2; 1, 0)$, a coefficient
count of at most 5, with bias coefficients $\bs^b_0(t) = (1, t)$, and $m = 1$.
\end{proposition}

This is the two-dimensional case of~\cite[Thm.~3.2]{cho2025hyperbolic}, which gives $\br = (d, 2; 1, 0)$ and a
coefficient count of at most $d+3$ in $d$ dimensions, with an approximation rate in the width from a ReLU
discretization of the planar-wave decomposition~\cite[\S5.F]{folland1995pde}.  The identity itself is quickly
checked: each $Q_\theta(\bomega_\theta\cdot \bx\mp t)$ solves the wave equation since $|\bomega_\theta| = 1$, the sum equals
$u_0$ at $t=0$ and has zero time derivative there, and the solution is unique.

\smallskip
The wave family thus needs a single varying coefficient, the time, which appears in the biases of a fixed set of plane-wave
ridges.  Its interpolation is a single-layer displacement of planar fronts, as in the Lax--Phillips translation
representation~\cite{laxphillips1967,rim2018radon}, and it does not involve a transport of the field by a velocity.
The decomposition is not specific to radial data ($Q_\theta$ may depend on $\theta$) and extends to nonzero initial
velocity by adding the ridges of its decomposition, with the time-integrated profile.  The LRNR does not
represent the outgoing ring and its wake as separate objects; they arise as a fixed superposition of traveling plane
waves.

\subsubsection*{Oscillatory families}  Here the profile in the moving frame $\by = (\bx-\bc_i)/r$ changes with $\mu$: a copy is
$a_i H(|\by|)\cos(\pi\kappa_i\phi(\by))$ with $\kappa_i = k_i r$ and $\phi(\by) = y$ (1D), $|\by|$ (rings) or
$\by\cdot \bd_i$ (wave packet).  With $\kappa_i\in[\bar\kappa_i - \Delta,\bar\kappa_i + \Delta]$ and $|\phi|\le1$ on the support,
\[ \cos(\pi\kappa_i\phi) = \sum_{n<n_\varepsilon}\frac{\bigl(\pi(\kappa_i-\bar\kappa_i)\bigr)^n}{n!}
   \RePart\bigl[(\mathrm i\phi)^n e^{\mathrm i\pi\bar\kappa_i\phi}\bigr] + R,\quad
   |R|\le\frac{(\pi\Delta)^{n_\varepsilon}}{n_\varepsilon!}\Bigl(1-\frac{\pi\Delta}{n_\varepsilon+1}\Bigr)^{-1}, \]
for $n_\varepsilon+1>\pi\Delta$, so each copy is a combination of $n_\varepsilon$ fixed profiles, and Proposition~\ref{prop:transported} applies to
each: the biases are as before ($r^b_0\le2$) and the outer weights lie in a space of dimension at most
$Pn_\varepsilon$.  The bound is crude; for any fixed linear fit of the profiles, the rank of the outer weights is at
most the $\varepsilon$-rank of the moving-frame profile family $\{a_i H(|\by|)\cos(\pi\kappa_i(\mu)\phi(\by))\}_\mu$, which
is computed directly (Table~\ref{tab:counts}).  A deeper LRNR needs fewer coefficients.  With the carrier frequency as a
weight coefficient and a fixed subnetwork computing $(\by, z)\mapsto H(|\by|)\cos(\pi z)$ from the channels
$\by = (\bx - \bc_i)/r$ and $z = \kappa_i\phi(\by)$ (the weight $\kappa_i$ multiplies the channel $\phi(\by)$, itself computed
by a fixed subnetwork unless $\phi$ is linear), all parameters lie in the span of $1/r$, $\mu$, $\mu^2$
for our families ($m\le3$ for every accuracy), at the price of a fixed subnetwork whose size grows with the
accuracy.

\subsubsection*{Spiral}  The arc is a growing piece of one fixed curve:
$C_\mu = \{0.7\mu s(\cos1.5\pi\mu s,\sin1.5\pi\mu s)\} = S([0,\mu])$ with
$S(\sigma) = 0.7\sigma(\cos1.5\pi\sigma,\sin1.5\pi\sigma)$, $\Lip(S)\le\Lambda = 3.4$.  For samples
$\sigma_j = j/J$,
\[ d_J(\bx;\mu) = \min_j\bigl(|\bx - S(\sigma_j)| + \Lambda\sigma_{\rm ReLU}(\sigma_j - \mu)\bigr)
   \quad\text{satisfies}\quad \dist(\bx, C_\mu)\le d_J\le\dist(\bx, C_\mu) + \Lambda/J : \]
a sample beyond the tip, $\sigma_j>\mu$, is penalized by at least the distance it lies beyond $S(\mu)$ along the
curve.  The penalty $\sigma_{\rm ReLU}(\sigma_j-\mu)$ is a neuron with bias $\sigma_j - \mu$ and no input weight;
$|\bx-S(\sigma_j)|$ and the minimum are computed by fixed subnetworks; and $aH(d_J/w) \approx
a\sum_k\gamma_k\rho(d_J - w\eta_k)$ puts $w$ in a bias and $a$ in the outer weights.  The parameters are affine in
$\mu$: $m = 1$, with an error at most $\tfrac{\pi}{2}a\Lambda/(wJ)$ plus the error of the fixed subnetworks, whose
size, and with it the rank $r$, grows with $J$.

\begin{table}[!htbp]
\centering\small
\caption{Explicit LRNR constructions.  Ranks of the first-layer weights and biases and of
the outer weights ($r^b_1 = 0$ throughout), the resulting bound on the coefficient count~\cite{cho2025hyperbolic},
and the latent dimension $m$; error: maximum relative $L^2$ error of the construction over a fine $\mu$
grid on the data grid (1D: $M$ steps per copy; 2D: 256 directions $\times$ 6000 nodes).  Oscillatory families:
$\varepsilon$-rank of the outer weights (1D, explicit construction) or of the moving-frame profiles (2D) for 1\% /
0.1\% extra error, and the a-priori Taylor bound $Pn_\varepsilon$.  Last column: number of POD modes of the snapshots
for 1\% error, the number a linear method in the lab frame needs.}
\label{tab:counts}
\begin{tabular}{lccccccc}
\toprule
family & $r^W_0$ & $r^b_0$ & $r^W_1$ & $\|\br\|_1$ & $m$ & error & \begin{tabular}[b]{@{}c@{}}POD\\(1\% err.)\end{tabular} \\
\midrule
1D split          & 1 & 2 & 2 & 5 & 1 & $9\times10^{-4}$ ($M=2000$) & 15 \\
1D fade (control) & 1 & 1 & 2 & 4 & 1 & $9\times10^{-4}$ ($M=2000$) & 2 \\
2D split          & 2 & 2 & 2 & 6 & 1 & $1.0\times10^{-3}$ & 22 \\
2D wave           & 2 & 2 & 1 & 5 & 1 & $1.1\times10^{-3}$ & 14 \\
2D spiral         & \multicolumn{4}{c}{deep (grows with accuracy)} & 1 & -- & 31 \\
\midrule
1D oscillatory split & 1 & 2 & 7 / 9 (Taylor 24 / 28) & 10 / 12 & & $2.6\times10^{-3}$ ($M=4000$) & 25 \\
2D rings             & 2 & 2 & 6 / 7 (Taylor 28 / 30) & 10 / 11 & & -- & 39 \\
2D wave packet       & 2 & 2 & 5 / 7 (Taylor 28 / 30) & 9 / 11 & & -- & 41 \\
\bottomrule
\end{tabular}
\end{table}

\subsubsection*{Relation to the trained models}  The networks trained here have $L=5$, $r=200$, 2000
coefficients and a four-dimensional latent~\eqref{eq:hyper}; the constructions above need two layers, ranks 1--2
and a one-dimensional affine latent for every non-oscillatory family except the spiral, and ranks of 5--9 for the oscillatory ones.  The
bounds concern existence.  In our runs, gradient descent did not find these sparse solutions, and the bounds give no information
on how many snapshots determine them.  Two features of the bounds are nevertheless visible in the
trained models: the latent path is nearly a straight line (\S\ref{sec:latent}), as $m=1$ with affine coefficients would have
it, and the motion comes from the first layer's biases (\S\ref{sec:biases}), as in
Proposition~\ref{prop:transported}.  For the Burgers family no closed form is available; for scalar conservation
laws in one dimension the analogous construction is given in~\cite[Thm.~4.6]{rim2026entropy}.

\subsection{Coefficient count and the fade--displacement transition}\label{sec:snapcount}

The bounds above concern representation.  This paper measures a different quantity: the number of
snapshots from which an interpolant that fits them also displaces its fronts between them.  A cross-fade fits any set
of snapshots, so the snapshots alone do not exclude it; only a preference for simple interpolants can exclude it.  We take the
coefficient count as the measure of simplicity and ask when the simplest interpolant consistent with the snapshots
displaces its fronts.  This idealizes training (the networks trained here have far more coefficients than they need, and
gradient descent does not minimize the count), but it yields a threshold that can be computed for each family and
compared with the measured $N^\ast$.

Call a two-layer LRNR \emph{static} if its first-layer weights and biases do not depend on $\mu$.  Its output is then
a $\mu$-dependent combination of fixed functions, $u(\cdot;\mu) = \sum_k w_k(\mu)\rho(\bomega_k\cdot \bx - b_k)$: a
static LRNR does not move its fronts and interpolates by blending, like a linear reduced basis, DeepONet, or the
piecewise-linear cross-fade of the snapshots, which is itself a static LRNR.  For snapshots $u_j = u(\cdot;\mu_j)$,
$j=1,...,N$, let $n_\varepsilon(S_N)$ be the smallest dimension of a linear space that comes within relative error
$\varepsilon$ of every snapshot, and for the whole family $\mathcal M = \{u(\cdot;\mu)\}_{\mu\in[0,1]}$ let
$d_n(\mathcal M)$ be its Kolmogorov $n$-width (relative, in $L^2$).

\begin{proposition}[static interpolants]\label{prop:static}
Let $\Omega\subset\R^d$ be a box and let the snapshots be nonzero with compact support in $\Omega$.  There is an
$\varepsilon_0>0$, depending on the snapshots, such that a static two-layer Heaviside LRNR that is within relative
error $\varepsilon<\varepsilon_0$ of the $N$ snapshots has
\[ r^W_0\ge d,\quad r^b_0\ge 1,\quad r^W_1 + r^b_1\ge n_\varepsilon(S_N), \]
hence coefficient count at least $d + 1 + n_\varepsilon(S_N)$, and its error over the family is at least
$\sup_\mu \|u^M(\cdot;\mu)-u(\cdot;\mu)\|/\|u(\cdot;\mu)\|\ge d_{r^W_1+r^b_1}(\mathcal M)$.
\end{proposition}

\noindent\emph{Proof.}  A snapshot $u_j$ is nonzero with compact support in $\Omega$, so every line through a point
of its support leaves the support inside $\Omega$, and $u_j$ is not constant along any direction $\bn$: its relative
$L^2$ distance to the functions on $\Omega$ that are constant along $\bn$ is positive.  This distance depends
continuously on $\bn$, so it has a positive minimum over the unit sphere; the same holds for the functions that are
constant along rays from the origin.  Let $\varepsilon_0$ be the smallest of these minima over the snapshots.  If the
$\bomega_k$ span fewer than $d$ dimensions, the output is constant along a direction $\bn\perp\Span\{\bomega_k\}$,
and if every bias is zero, each $\rho(\bomega_k\cdot \bx)$ is constant along rays from the origin; in both cases the
relative error is at least $\varepsilon_0$.  The outputs lie in the span of the fixed ridges combined with the
outer weights and the output bias, a fixed space of dimension at most $r^W_1 + r^b_1$; it must come within
$\varepsilon$ of every snapshot, and since it is one fixed space for all $\mu$, its worst-case relative distance to
the family is at least $d_{r^W_1+r^b_1}(\mathcal M)$.  \hfill$\square$

\smallskip
For snapshots of a moving feature $n_\varepsilon(S_N) = N$ as long as $N$ is below the linear $\varepsilon$-rank of
the family, so a static interpolant pays one coefficient per snapshot, while the displacing construction of
\S\ref{sec:counts} has a fixed count $n_T$, independent of $N$.  The comparison gives a threshold.

\begin{proposition}[transition]\label{prop:transition}
Let the family have a two-layer LRNR of coefficient count $n_T$ within relative error $\varepsilon<\varepsilon_0$
(Proposition~\ref{prop:static}) at every $\mu$, let $n_\varepsilon(S_N)$ be nondecreasing in $N$, and let
\[ N_c = \min\{N : d + 1 + n_\varepsilon(S_N) > n_T\}. \]
For $N\ge N_c$, every two-layer LRNR of minimal coefficient count that is within $\varepsilon$ of the snapshots moves
its first layer with $\mu$; every static LRNR whose count does not exceed $n_T$ has worst-case error at least
$d_{n_T-d-1}(\mathcal M)$ over the family.  For $N < n_T - d - 1$ (that is, $N < N_c - 1$ when $n_\varepsilon(S_N) = N$) a
static interpolant of lower count than $n_T$ exists whenever the piecewise-linear cross-fade of the snapshots is within $\varepsilon$ of them (it is exact at the
snapshots), so the snapshots and the count alone favor fading.
\end{proposition}

\noindent\emph{Proof.}  The first two statements follow from Proposition~\ref{prop:static} and the existence of the
displacing LRNR, which is consistent with the snapshots and has count $n_T$.  For the third, the cross-fade is a static
LRNR with $r^W_1 = N$ and first-layer ranks those of a representation of the snapshots with fixed ridges, $d$ and
$1$ for the constructions used here.  \hfill$\square$

\begin{table}[!htbp]
\centering\small
\caption{Snapshot counts from coefficient counts ($\varepsilon = 1\%$).  $n_T$: count of the
displacing construction (Table~\ref{tab:counts}); $N_c$: threshold of Proposition~\ref{prop:transition}; lower bound on
the worst held-out error of any static LRNR with count $\le n_T$ (rms over $\mu$ of the best $(n_T-d-1)$-dimensional
space, a lower bound on $d_{n_T-d-1}$); measured $N^\ast$ and, at the largest $N$ below it, the filtered fade
ratios of the two seeds (oscillatory families: envelope fade ratio over $\mu\ge0.1$, or the condition that fails).}
\label{tab:snapcount}
\begin{tabular}{lccccc}
\toprule
family & $n_T$ & $N_c$ & static error $\ge$ & $N^\ast$ (measured) & below $N^\ast$ \\
\midrule
1D split          & 5  & 4 & 0.24 & 5 & $N=4$: 0.87 / 0.91 \\
2D split          & 6  & 4 & 0.61 & 5 & $N=3$: 0.68 / 0.76 \\
1D fade (control) & 4  & -- (static is minimal) & 0 & 3 (fades) & \\
\midrule
1D oscillatory split & 10 & 9 & 0.19 & 9  & $N=7$: s0 gap \\
2D rings             & 10 & 8 & 0.77 & 17 & $N=13$: 0.84 / 0.86 \\
2D wave packet       & 9  & 7 & 0.66 & 21 & $N=17$: s1 fit \\
2D spiral            & \multicolumn{2}{c}{no small two-layer $n_T$} & & 25 & $N=21$: 0.82 / 0.70 \\
\bottomrule
\end{tabular}
\end{table}

\subsubsection*{Comparison with the measured transition}  Table~\ref{tab:snapcount} lists the thresholds.  On every
family the measured $N^\ast$ is at or above $N_c$, as the argument requires if training favors low counts at all.  On
the split families, where the count is smallest, the two are close: at $N=2$ the cross-fade has the lower count
($d+3$ against $d+4$) and every model fades (fade ratios 0.34--0.50); at $N=3$ the counts tie and the trained models
are in between, with fade ratios of 0.68--0.83; from $N=4$ the static interpolant costs more.  On the 1D split we
ran $N=4$ as well: the fade ratios rise to 0.87 / 0.91 and one seed passes; both seeds pass from $N=5$, which is
$N^\ast$ under the all-seeds rule.  This is consistent with the predicted threshold, but the transition is gradual
over $N = 3$--5 rather than a step, and two seeds cannot place it more precisely; the count gives the order of
magnitude of $N^\ast$ but does not determine its exact value.  On the oscillatory
families (\S\ref{sec:osc2d}) the threshold is larger, 7--9.  The 1D oscillatory family meets it ($N^\ast = 9 = N_c$, with one seed already
passing at $N=7$); the 2D families need two to three times more snapshots ($N^\ast = 17$ for the rings, 21 for the wave
packet, whose second seed fails only on its fit at $N = 17$).  The trained models are deeper and smoother than
the construction, and we do not know how close training comes to the count minimum there; the count again gives a
lower bound and the order of magnitude but not the exact $N^\ast$.  The spiral has no small two-layer displacing count (its arc grows rather than moves, and the
deep construction of \S\ref{sec:counts} has a rank that grows with the accuracy), which is consistent with its much
larger $N^\ast$.  The fade control is the case where the truth is static: its snapshots span two dimensions for every $N$, a
static LRNR of count 4 represents it exactly, and the argument does not force motion; the LRNR correctly fades there.

\subsubsection*{Scope of the counting argument}  The argument concerns an idealized learning procedure; it is
not a statement taking gradient descent into account.  It is consistent with the shape of the transition: a fade is the simplest explanation of two
snapshots, the counts tie at three, and displacement is simpler after that, while the trained models move gradually
rather than at a sharp step.  It gives a threshold that depends only on the family, through $n_T$, which can be
computed in advance.  It also separates the two kinds of interpolant quantitatively: a static LRNR
that is as cheap as the displacing one cannot be more accurate than the static-error column of Table~\ref{tab:snapcount}, while
the displacing construction is accurate to $\varepsilon$ at every $\mu$.  It does not guarantee that
training finds the count-minimal interpolant, and it gives no rate in $N$ for families whose coefficient path is not
affine (the spiral); both questions are open problems.

\section{Training with spatial mini-batches}\label{sec:training}

\subsection{Cells and schedule}
Each training step (called an \emph{epoch} below) uses the points of one spatial cell of one snapshot, following
spatially local mini-batching~\cite{choi2026local}.
The cells form two levels: a coarse level of $\lfloor n_d/2\rfloor$ cells per axis and a fine level of $n_d$ cells per
axis on the domain, where $n_d$ is a fixed integer; in 2D this gives $\lfloor n_d/2\rfloor^2 + n_d^2$
cells.  The study uses $n_d = 2$ (Table~\ref{tab:hyper}): one whole-domain cell and four quadrants in 2D (one whole-domain cell and two
halves in 1D), so that about one step in five (one in three in 1D) sees the whole snapshot.  At every step the cell boundaries of each
level are shifted by a random translation of up to $\pm\frac14$ of a cell, with edge cells clamped to the domain, so
the cell boundaries change from step to step.  Cells are visited by cycling a shuffled list.  Every 100 steps an
adaptive boost appends 10 extra entries to the list for each cell whose error is at or above the 0.8 quantile, where
a cell's error is an exponential moving average (weight 0.3) of its last loss divided by its target variance.
Snapshots are visited round-robin (the code's timer curriculum grows and resets the set of admitted snapshots
every few steps, which amounts to uniform visits).

\subsection{Loss}\label{sec:loss}
For the cell $k$ of snapshot $\mu$ with points $X_k$ and data $Y_k$, the loss is
\begin{equation}
\begin{aligned}
  \mathcal{L} ={}& \tfrac{1}{|X_k|}\|u(X_k;\mu)-Y_k\|^2
   + \lambda_{\rm sp}\tfrac{1}{|X_k|}\|u^{(k)}(X_k;\mu)-u(X_k;\mu)\|^2 \\
   &+ \phi(t)\bigl(\lambda_{\perp}R_\perp + \lambda_{s}R_s + \lambda_{\perp,\rm sp}R_{\perp,\rm sp}\bigr),
\end{aligned}
  \label{eq:loss}
\end{equation}
where the first term is the misfit; $u^{(k)}$ is the filtered pass through cell $k$'s projectors; $R_\perp = \sum\|\mathbf A^\top \mathbf A - \mathbf I\|_F^2/(r^2L)$ over the
bases $\mathbf A\in\{\bU_\ell,\bV_\ell,\bB_\ell\}$ (orthogonality); $R_s$ penalizes departures of the coefficients from a geometric
decay along the rank axis (ratio $\gamma = 4$); $R_{\perp,\rm sp}$ is the orthogonality of the projectors; and
$\phi(t) = \frac12(1+\sin(10^{-3}t-\pi/2))$ cycles the three structural terms with period about 6300 steps.  The study
uses $\lambda_\perp = 10^{-3}$, $\lambda_s = 10^{-5}$, $\lambda_{\rm sp} = 10^{-3}$, $\lambda_{\perp,\rm sp} = 10^{-9}$.
Dropout is applied after every layer of the main pass ($p = 0.1$) and of the filtered pass ($p=0.05$); in the second half
of training both rates are multiplied at every curriculum reset by $0.99999$ and $0.999$ respectively, so the main-pass
rate ends near $0.09$ while the filtered-pass rate becomes negligible.  None of these
terms refers to motion; the regularizers available for that in our implementation (a Wasserstein term, constant-speed and
linear-coefficient path penalties) are off.  All of these terms are removed in the ablation of \S\ref{sec:ablation}.

\subsection{Optimization and budget}
Adam~\cite{kingma2015adam} with learning rate $10^{-4}$ (constant in effect; the plateau scheduler's factor of $0.9999$ does not act during training) for the
bases and hypernetwork, and a separate Adam for the projectors.  To compare datasets of different size fairly, the
number of steps is proportional to $N$, so that every snapshot receives the same number of visits:
$60{,}000N/3$ steps for 1D and $100{,}000N/3$ for 2D families.

\subsubsection*{Stopping at fade-ratio saturation}  Learning curves (\S\ref{sec:budget}) showed that the fade ratio
stops improving long before the fit does.  Runs after the initial sweep therefore save a checkpoint of the model every
40,000 steps, score it (\S\ref{sec:metrics}), and stop once the filtered fade ratio of the last three checkpoints varies by
at most $0.05$ and at least 120,000 steps have been taken.

\subsubsection*{Second stage: $L^1$ refinement}  The first stage ends at a learning rate of $10^{-4}$ with the regularizers
and dropout on, a setting that finds the displacing solution but does not converge in it.  The second stage starts
from the first stage's final model, replaces the squared misfit by $\frac{1}{|X_k|}\|u(X_k;\mu)-Y_k\|_1$, lets the
learning rate decay from $2\times10^{-6}$ to $10^{-8}$ by a cosine schedule, divides the four loss weights by ten and
turns dropout off, for a quarter of the first stage's steps (an eighth for the spiral).  The snapshot-count sweep of
\S\ref{sec:transition} uses first-stage models, since whether a model displaces its fronts is decided there (on the
non-oscillatory families the second stage lowers no fade ratio by more than 0.03 and raises none by more than 0.06,
\S\ref{sec:l1}); the pass rule is applied to the first stage, and at each family's $N^\ast$ the reported LRNR is the
final, second-stage model.

\begin{table}[!htbp]
\centering\small
\caption{LRNR settings used throughout (all examples).}
\label{tab:hyper}
\begin{tabular}{lp{0.78\linewidth}}
\toprule
network & 5 layers, width 1000, $\tanh$; input features: degree-2 monomials (Chebyshev degree 9 for oscillatory) \\
low rank & rank 200 (2000 coefficients), bases $\approx 2.4$M parameters \\
hypernetwork & $1\to4\to4\to2000$, $\tanh,\tanh,\softplus_{10}$ \\
spatial cells & $n_d = 2$ (1 + 4 cells in 2D, 1 + 2 in 1D), jitter $\pm\frac14$ cell, adaptive boost every 100 steps \\
projectors & rank 300 per cell and layer (training only) \\
loss weights & $\lambda_\perp=10^{-3}$, $\lambda_s=10^{-5}$ ($\gamma=4$), $\lambda_{\rm sp}=10^{-3}$, $\lambda_{\perp,\rm sp}=10^{-9}$, cyclic schedule \\
dropout & 0.1 (main pass, $\to\approx0.09$), 0.05 (filtered pass, $\to0$) over the second half \\
optimizer & Adam, $10^{-4}$; batch = one cell of one snapshot \\
budget & $60$k$\cdot N/3$ (1D), $100$k$\cdot N/3$ (2D) steps; or saturation stopping (checkpoints every 40k) \\
pass rule & training error $\le0.30$, gap $\le1.5$ (or $\le0.05$ absolute), fade ratio $\ge0.9$; oscillatory families: over $\mu\ge0.1$, envelope fade ratio; applied to the first stage \\
second stage & $L^1$ misfit, Adam $2\times10^{-6}\to10^{-8}$ (cosine), loss weights $\times0.1$, no dropout, $\frac14$ of the steps ($\frac18$ spiral) \\
\bottomrule
\end{tabular}
\end{table}

\section{Metrics}\label{sec:metrics}

\subsubsection*{Errors}  With $e(\mu) = \|u_{\rm model}(\cdot;\mu)-u(\cdot;\mu)\|_2/\|u(\cdot;\mu)\|_2$ on the full grid,
the \emph{training error} is the mean of $e$ over the training $\mu_k$ and the \emph{held-out error} the mean over 101
evenly spaced $\mu$ and the gap midpoints, excluding the training values.  The ratio of held-out to training error is the \emph{interpolation gap}.

\subsubsection*{Fade ratio}  For a gap $[\mu_g,\mu_{g+1}]$, eleven interior points $\mu$ with fractions
$a = (\mu-\mu_g)/(\mu_{g+1}-\mu_g)$, and a region $\Omega_j$ of the domain (the two halves in 1D, the four quadrants in
2D), let $C_f(\mu) = (1-a)f(\mu_g)+af(\mu_{g+1})$ be the cross-fade of a family $f$ between the gap's ends, and
\begin{equation*}
  d_f = \operatorname*{mean}_{\mu}\frac{\|f(\mu)-C_f(\mu)\|_{\Omega_j}}{\|C_f(\mu)\|_{\Omega_j}} .
\end{equation*}
The fade ratio of the pair $(g,j)$ is $d_{\rm model}/d_{\rm truth}$: about $1$ when the model departs from its own
cross-fade as much as the truth does from its own, about $0$ when the model is the cross-fade.  Regions with
less than 10\% of the truth's energy in the gap are skipped.  Where the truth itself is close to a fade
($d_{\rm truth}$ small, as at fine snapshot spacing) the ratio compares two small numbers and is noise, so the
\emph{filtered fade ratio}, reported throughout, is the minimum of the pair ratios over the pairs with
$d_{\rm truth}\ge0.1$.

\subsubsection*{Pass rule and transition point}  A run passes at threshold $\tau$ if its training error is at most 0.30,
its gap at most 1.5 (or its held-out error at most 0.05 above its training error: once the snapshots are fitted to
a fraction of a percent the ratio compares two small errors and no longer indicates overfitting), and its filtered
fade ratio at least $\tau$ ($\tau = 0.9$ unless stated).  The transition point
$N^\ast$ is the smallest $N$ at which every seed passes, over the sweep's $N$ and any further evenly spaced runs of the
family.  We do not also require every larger $N$ to pass: at fine spacing the fade ratio compares two small distances
and becomes noisy (\S\ref{sec:transition}), and confirming it would need ever larger runs.  Two seeds are trained
per $(N,\text{family})$, one for the fade control.  The rule is applied to the first training stage
(\S\ref{sec:training}): a run that passes there counts as passed, and the second stage started from it is the reported
model without being tested again (\S\ref{sec:l1}).  The three oscillatory families need an adapted version of the
rule and are reported separately (\S\ref{sec:osc2d}); everything in \S\ref{sec:transition}--\S\ref{sec:classical}
concerns the other four families (1D split, 2D split, 2D spiral, fade control); the 2D wave and 2D Burgers families are reported in Appendix~\ref{app:nofade}.

The fade ratio measures how far the model departs from a fade; it does not measure whether the model moves in the
right direction.  The gap condition and the analyses of \S\ref{sec:dynamics} cover the direction.

\section{Results: the snapshot-count transition}\label{sec:transition}

\begin{figure}[!htbp]
  \centering
  \includegraphics[width=\linewidth]{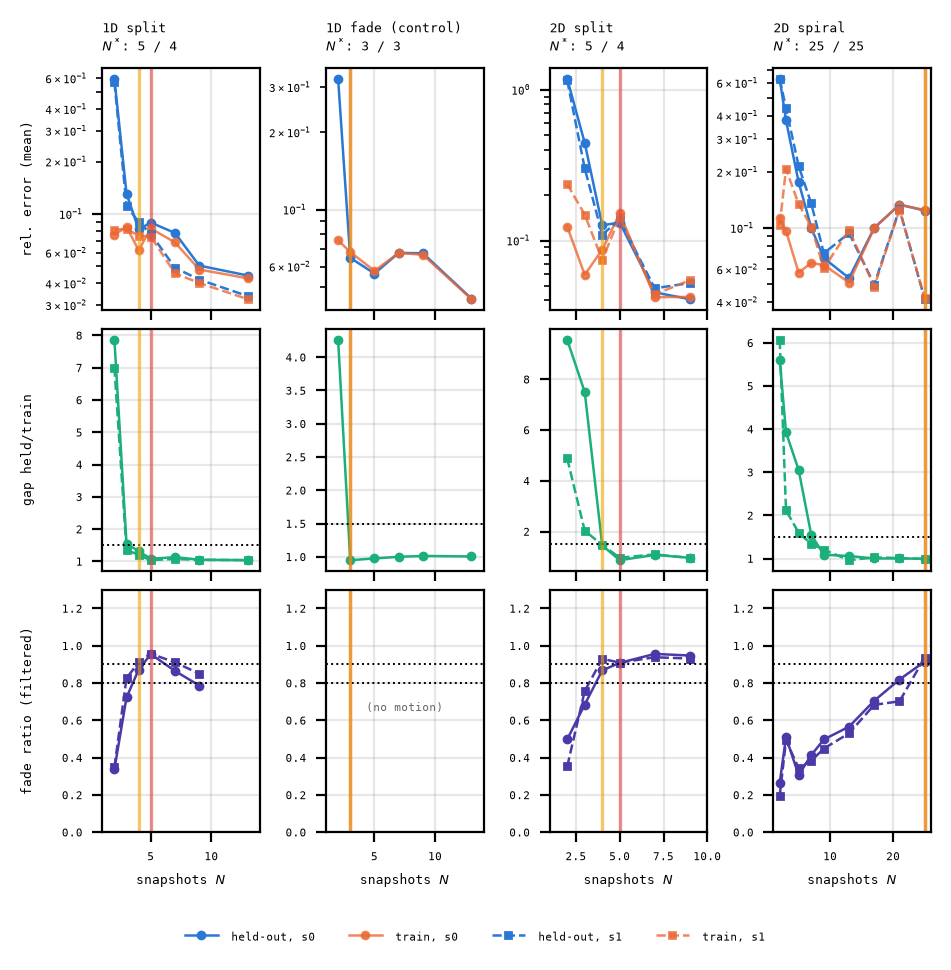}
  \caption{Transition sweep: held-out error, interpolation gap and filtered fade ratio against the number of
  snapshots $N$, per family and seed (first-stage LRNR); vertical lines mark $N^\ast$ at fade threshold 0.9 (red)
  and 0.8 (orange), given in the column titles as 0.9 / 0.8.  The oscillatory families have their own rule and
  figure (Figure~\ref{fig:osctransition}).}
  \label{fig:transition}
\end{figure}

Table~\ref{tab:transition} and Figure~\ref{fig:transition} summarize the sweep over $N\in\{2,3,5,7,9,13\}$ (2D up to
9), extended for the spiral up to 25.  The oscillatory families are in \S\ref{sec:osctransition}.

\begin{table}[!htbp]
\centering\small
\caption{Snapshot-count transition (first-stage LRNR, seeds 0/1).  Held-out error and filtered fade ratio at
selected $N$; $N^\ast$ by the pass rule of \S\ref{sec:metrics}: 1D split 5, 1D fade 3, 2D split 5, 2D spiral 25
(oscillatory families: Table~\ref{tab:osctransition}).  The 2D wave and 2D Burgers families are reported in
Appendix~\ref{app:nofade}.}
\label{tab:transition}
\begin{tabular}{lccccc}
\toprule
family & $N$ & train & held-out & fade (filtered) & passes \\
\midrule
1D split & 3 & 0.084 / 0.082 & 0.131 / 0.112 & 0.72 / 0.83 & no \\
         & 5 & 0.083 / 0.074 & 0.089 / 0.077 & \underline{0.96 / 0.96} & yes \\
         & 7 & 0.069 / 0.046 & 0.078 / 0.049 & 0.86 / 0.91 & s1 \\
1D fade (control) & 3 & 0.068 & 0.065 & -- & yes \\
2D split & 3 & 0.059 / 0.148 & 0.445 / 0.301 & 0.68 / 0.76 & no \\
         & 5 & 0.152 / 0.143 & 0.132 / 0.138 & \underline{0.91 / 0.91} & yes \\
         & 9 & 0.042 / 0.055 & 0.041 / 0.052 & 0.95 / 0.93 & yes \\
2D spiral & 9 & 0.063 / 0.061 & 0.069 / 0.073 & 0.50 / 0.45 & no \\
          & 25$^\dagger$ & 0.125 / 0.042 & 0.124 / 0.041 & \underline{0.92 / 0.94} & yes \\
\bottomrule
\multicolumn{6}{l}{\footnotesize $^\dagger$ spacing study, \S\ref{sec:spiral} ($N = 13, 17, 21$ fail); last first-stage models, stopped at 320k and 680k steps.}
\end{tabular}
\end{table}

\begin{figure}[!htbp]
  \centering
  \includegraphics[width=\linewidth]{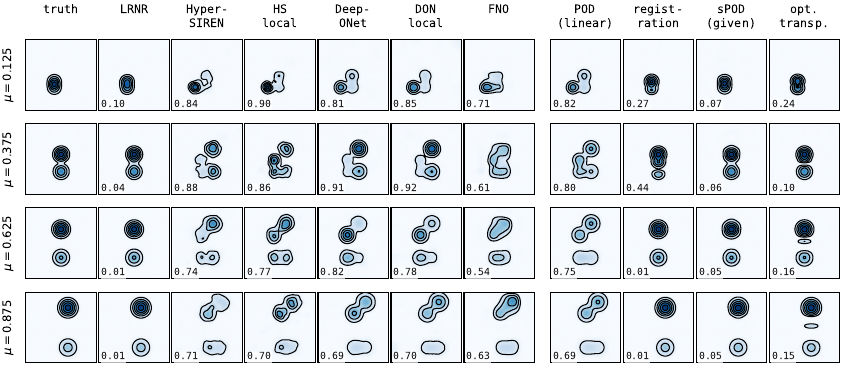}
  \caption{2D split, $N=5$, at the four gap midpoints (unseen $\mu$, rows): truth, final LRNR, five of the neural baselines (Hyper-SIREN, DeepONet, each also trained on the LRNR's cells, and the FNO)
  and four classical interpolants (seed 0 for the trained models); every panel is contoured at the truth's levels (0.1,
  0.3, ..., 0.9 of its maximum) and the numbers are relative errors.  The LRNR moves the two copies; Hyper-SIREN and
  DeepONet (with whole snapshots or the LRNR's cells) and linear POD show a fading copy at the old position and a
  growing copy at the new one.  Optimal transport leaves a filament between the separating copies ($\mu = 0.625, 0.875$) and
  registration folds the hump at the split ($\mu = 0.375$); sPOD with the exact shifts provided is accurate
  throughout, but those shifts are the transport itself, given to it and not to the LRNR.  sPOD with fitted shifts and cubic-spline POD are in Table~\ref{tab:classical}.}
  \label{fig:strip2d}
\end{figure}

\begin{itemize}
  \item \emph{2D split}: $N^\ast = 5$.  At $N\le3$ the copies fade (Figure~\ref{fig:panels2d}); from $N=5$ on both seeds move them
        (Figure~\ref{fig:strip2d}), and the error at unseen $\mu$ equals the training error (gap $\le 1.1$).
  \item \emph{1D split}: $N^\ast = 5$, both seeds at fade 0.96.  At larger $N$ the filtered ratio of one seed dips
        below 0.9 (0.86 at $N=7$) and at $N = 13$ no cell has $d_{\rm truth}\ge0.1$: the gaps are so short that the
        truth itself is nearly a fade and the ratio is no longer informative, while the held-out error keeps falling
        (0.077--0.089 at $N=5$, 0.034--0.044 at $N=13$).  For this reason, $N^\ast$ is the smallest passing $N$.
  \item \emph{1D fade (control)}: passes from $N=3$ with gap $\approx 1$.  The LRNR fades where the truth fades.
  \item \emph{2D spiral}: $N^\ast = 25$.  No $N\le 21$ passes; the fade lies in the arc's tail, where it moves
        fastest (\S\ref{sec:spiral}).
\end{itemize}

\begin{figure}[!htbp]
  \centering
  \includegraphics[width=\linewidth]{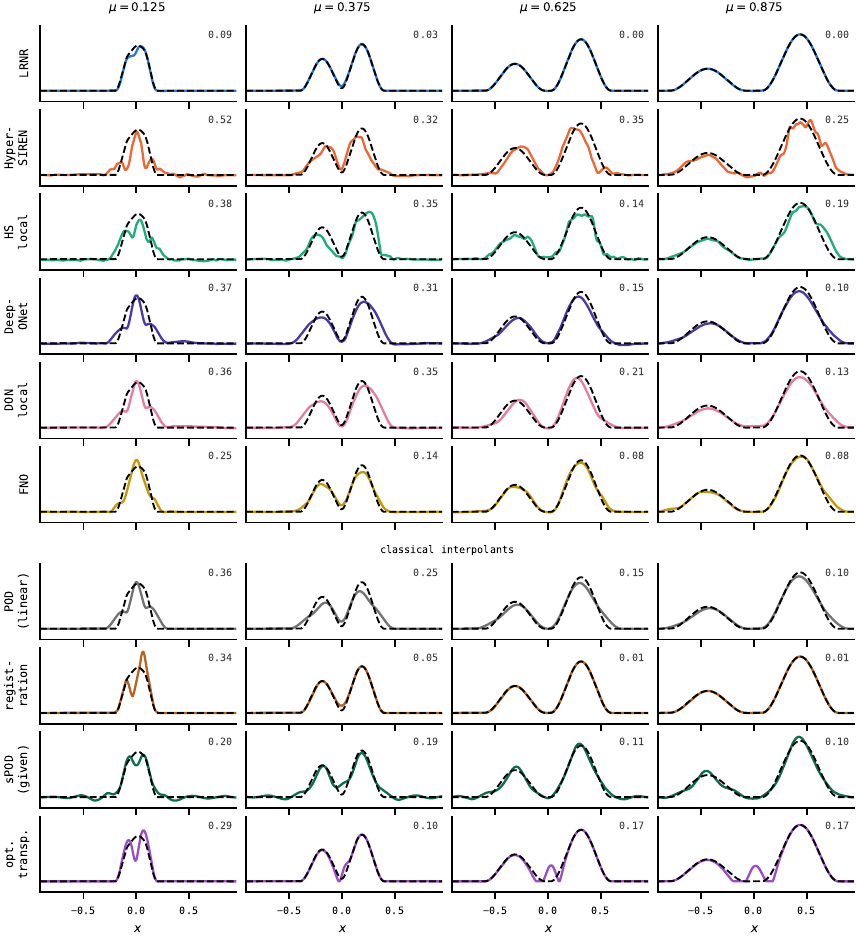}
  \caption{1D split, $N=5$, at the gap midpoints: the final LRNR, five of the neural baselines and four classical
  interpolants (solid, seed 0 for the trained models) against the truth (dashed); numbers are relative errors.}
  \label{fig:strip1d}
\end{figure}

\subsection{Snapshot placement and spacing: the spiral}\label{sec:spiral}

\begin{figure}[!htbp]
  \centering
  \includegraphics[width=\linewidth]{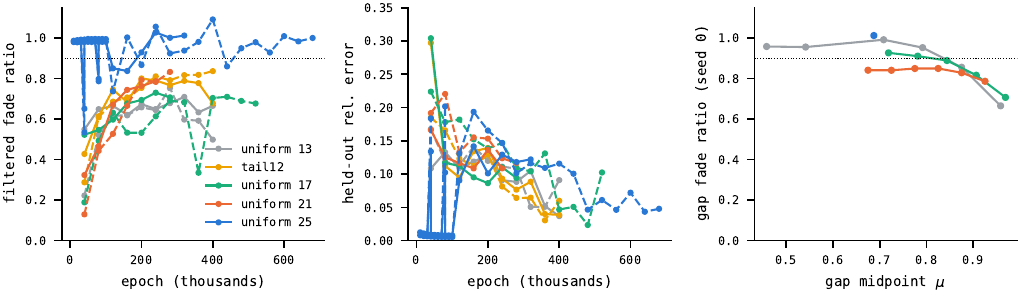}
  \caption{Spiral designs.  Left, middle: filtered fade ratio and held-out error of the model snapshots against
  training steps (solid seed 0, dashed seed 1).  Right: fade ratio per gap at the last snapshot of seed 0, for the gaps with $d_{\rm truth}\ge0.1$; at finer spacing fewer
  gaps move enough to qualify (one at $N=25$).}
  \label{fig:spiral}
\end{figure}

At $N=9$ the spiral fades only in its tail, where the arc moves fastest.  We compared adding snapshots where it fades
with refining the uniform grid (Figure~\ref{fig:spiral}, Table~\ref{tab:spiral}):
\emph{tail12} adds the midpoints of the last three gaps of $N=9$ (spacing $1/16$ on $[0.625,1]$), and uniform $N=13$,
17, 21 and 25 refine everywhere.

\begin{table}[!htbp]
\centering\small
\caption{Spiral: snapshot designs (seed 0 / seed 1; last scored model; filtered fade ratio).}
\label{tab:spiral}
\begin{tabular}{lccc}
\toprule
design & tail spacing & held-out & fade (filtered) \\
\midrule
uniform 9  & 1/8  & 0.069 / 0.073 & 0.50 / 0.45 \\
uniform 13 & 1/12 & 0.037 / 0.091 & 0.67 / 0.50 \\
tail12     & 1/16 & 0.039 / 0.060 & 0.68 / 0.84 \\
uniform 17 & 1/16 & 0.100 / 0.050 & 0.71 / 0.68 \\
uniform 21 & 1/20 & 0.134 / 0.126 & 0.82 / 0.70 \\
uniform 25 & 1/24 & 0.124 / 0.041 & \underline{0.92 / 0.94} \\
\bottomrule
\end{tabular}
\end{table}

\begin{figure}[!htbp]
  \centering
  \includegraphics[width=0.85\linewidth]{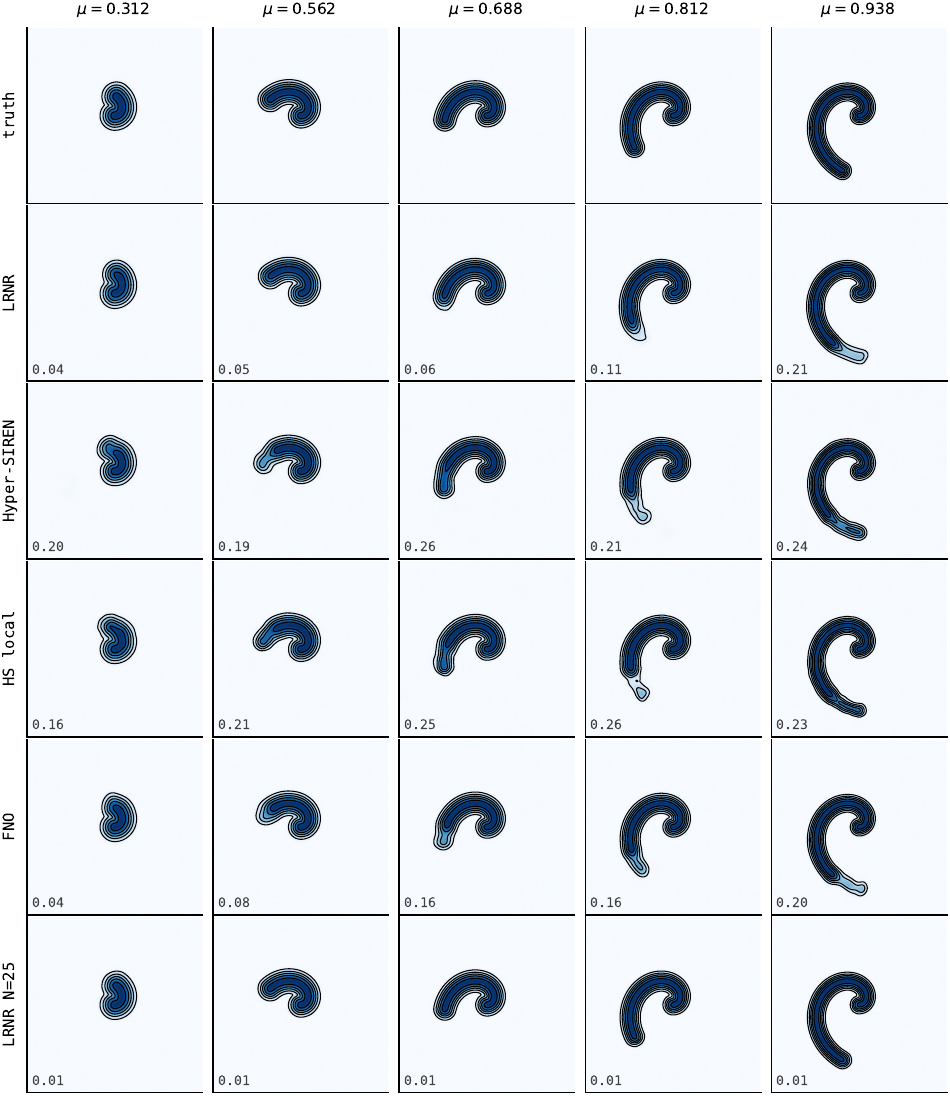}
  \caption{Spiral at unseen $\mu$: LRNR, Hyper-SIREN, local Hyper-SIREN and the FNO at $N=9$, and the LRNR at $N=25$.  At
  $N=9$ the end of the arc is cross-faded between the two neighboring snapshots ($\mu = 0.8125, 0.9375$); at $N=25$
  it is displaced.}
  \label{fig:strip_spiral}
\end{figure}

Up to $N\approx 17$, the placement affects the fade ratio more than the count.  Tail12, with 12 snapshots, matches
uniform 17.  Beyond that the
remaining fade moves into the last two gaps ($\mu\ge0.875$; Figure~\ref{fig:spiral}, right) and only a finer tail
spacing removes it: uniform 25 passes for both seeds (last first-stage models: fade 0.92 and 0.94, held-out 12.4\% and 4.1\%;
the model snapshots along training range over 0.86--1.09), so $N^\ast = 25$ for the spiral.  Figure~\ref{fig:strip_spiral} shows the tail at $N=9$ and $N=25$.

\subsection{Training length}\label{sec:budget}

\begin{figure}[!htbp]
  \centering
  \includegraphics[width=\linewidth]{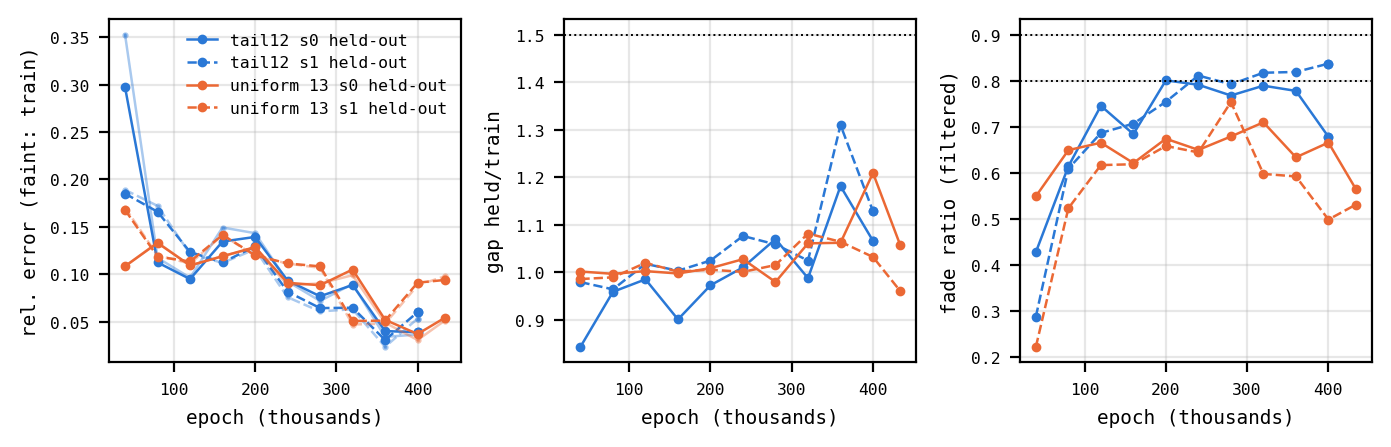}
  \caption{Learning curves from saved models, spiral tail12 and uniform 13 (models saved every 40k steps): held-out and
  training error, gap, and filtered fade ratio against training steps.}
  \label{fig:curves}
\end{figure}

Scoring saved models along training (Figure~\ref{fig:curves}, Table~\ref{tab:splitckpt}) shows that the fade ratio saturates early (by
40--80k steps for the 2D split and about 200k for the spiral), while the fit keeps improving.  For the 2D split at
$N=5$ the fade ratio is already 0.91 at 40k steps for seed 0 and 0.92 at 80k for seed 1 (Appendix,
Table~\ref{tab:splitckpt}); the fixed 167k-step budget mostly improves accuracy, and individual checkpoints vary by a
factor of two in held-out error around the trend (constant learning rate).  This motivated stopping on saturation of
the fade ratio (\S\ref{sec:training}).

\subsection{Spatial mini-batching: cell size}\label{sec:ndomains}

\begin{figure}[!htbp]
  \centering
  \includegraphics[width=0.9\linewidth]{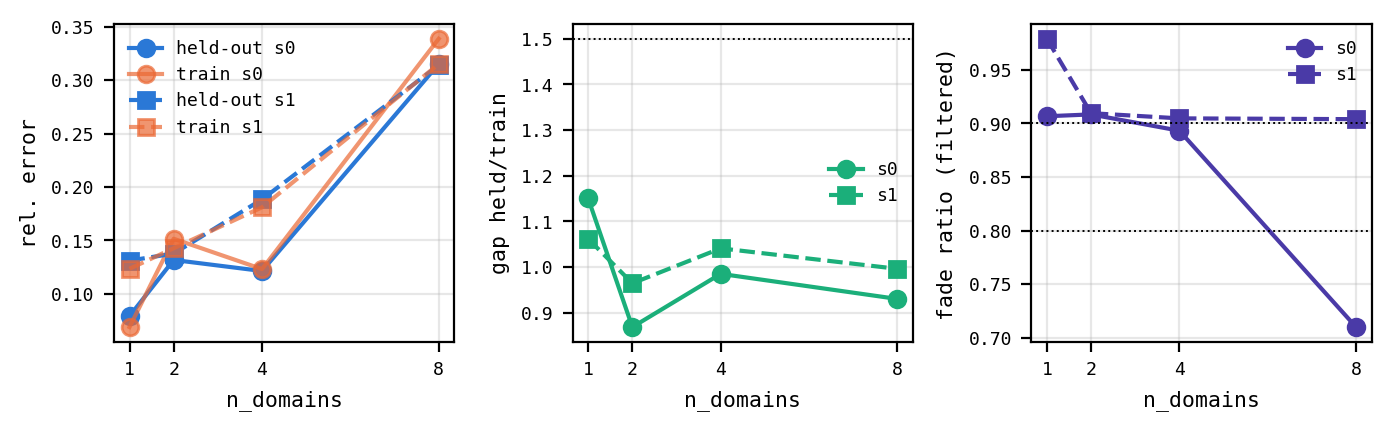}
  \caption{Cell-size study, 2D split $N=5$: errors, gap and filtered fade ratio against $n_d$ (1, 5, 20 and 80 cells per
  snapshot).}
  \label{fig:ndomains}
\end{figure}

Table~\ref{tab:ndomains} and Figure~\ref{fig:ndomains} vary the cell size on the 2D split at $N=5$, from whole snapshots ($n_d=1$) to $16+64$
cells ($n_d = 8$), with saturation stopping.  With $n_d = 2$ and $4$, the fade ratio stays close to the
whole-snapshot reference (0.89--0.91 against 0.91--0.98).  The smallest cells ($n_d=8$, an eighth of the domain per
axis) lose accuracy at this budget (held-out 31\%), and one seed fades.  Wall-clock times
are dominated by the shared GPU and are comparable only within a batch ($n_d = 4$ and 8 ran together: 4.7--6.8 h
and 6.5--6.6 h, respectively).

\begin{table}[!htbp]
\centering\small
\caption{Cell size, 2D split, $N=5$ (seed 0 / seed 1).}
\label{tab:ndomains}
\begin{tabular}{lcccc}
\toprule
$n_d$ & cells per snapshot & held-out & fade (filtered) & passes \\
\midrule
1 & 1 (whole) & 0.079 / 0.131 & 0.91 / 0.98 & both \\
2 & 1 + 4 & 0.132 / 0.138 & 0.91 / 0.91 & both \\
4 & 4 + 16 & 0.121 / 0.188 & 0.89 / 0.90 & s1 \\
8 & 16 + 64 & 0.315 / 0.314 & 0.71 / 0.90 & neither \\
\bottomrule
\end{tabular}
\end{table}

\subsection{Regularization ablation}\label{sec:ablation}

\begin{figure}[!htbp]
  \centering
  \includegraphics[width=\linewidth]{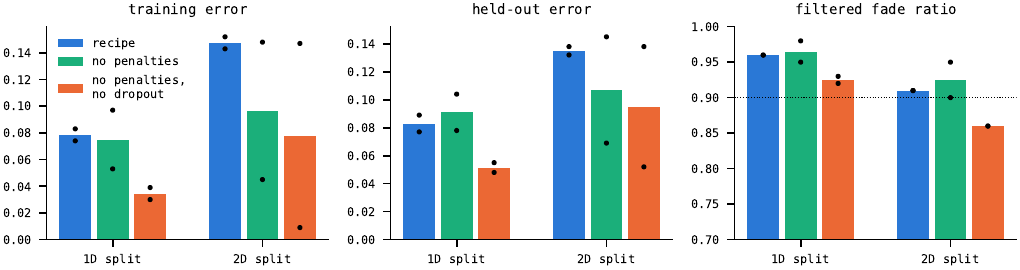}
  \caption{Regularization ablation at $N=5$: bars are seed means, dots individual seeds.}
  \label{fig:ablation}
\end{figure}

At $N^\ast = 5$ of the two split families, we retrained the LRNR (i) with all four loss
weights of~\eqref{eq:loss} set to zero, and (ii) additionally without dropout, i.e., on the misfit alone
(Table~\ref{tab:ablation}, Figure~\ref{fig:ablation}).  The projectors are not on the output path, so (ii) fits the
output by the misfit alone; the training still visits local cells.

\begin{table}[!htbp]
\centering\small
\caption{Regularization ablation at $N=5$ (seed 0 / seed 1).}
\label{tab:ablation}
\begin{tabular}{llcccc}
\toprule
family & variant & train & held-out & gap & fade (filtered) \\
\midrule
1D split & recipe            & 0.083 / 0.074 & 0.089 / 0.077 & 1.08 / 1.04 & 0.96 / 0.96 \\
         & no penalties      & 0.053 / 0.097 & 0.078 / 0.104 & 1.47 / 1.08 & 0.98 / 0.95 \\
         & no penalties, no dropout & 0.030 / 0.039 & 0.048 / 0.055 & 1.60 / 1.39 & 0.92 / 0.93 \\
2D split & recipe            & 0.152 / 0.143 & 0.132 / 0.138 & 0.87 / 0.96 & 0.91 / 0.91 \\
         & no penalties      & 0.045 / 0.148 & 0.069 / 0.145 & 1.54 / 0.98 & 0.90 / 0.95 \\
         & no penalties, no dropout & 0.009 / 0.147 & 0.052 / 0.138 & 5.74 / 0.94 & 0.86 / 0.86 \\
\bottomrule
\end{tabular}
\end{table}

Without the penalty weights, the fade ratio is unchanged (0.90--0.98) and the fit improves, so the penalties are not
needed to keep the LRNR from fading.  Removing dropout as well lowers the fade ratio slightly (to 0.86--0.93) and loosens the coupling
between training and held-out error: the 2D seed-0 run fits its snapshots to 0.9\% but reaches 5.2\% at unseen
$\mu$ (gap 5.7, but within 0.05 in absolute terms) and is the most accurate first-stage 2D model at $N=5$, yet fails on its fade
ratio (0.86).  We posit that the bias against fading comes mostly from the architecture (\S\ref{sec:dynamics}), with
dropout as a small additional aid.

\subsection{The second training stage}\label{sec:l1}

Table~\ref{tab:l1} compares the models at $N^\ast$ after the first stage and after the second ($L^1$ refinement,
\S\ref{sec:training}).  A first version of the second stage, at a constant learning rate of $10^{-5}$ with the
first stage's regularization, lowered the worst-case error slightly and raised the mean error.

\begin{table}[!htbp]
\centering\small
\caption{The two training stages at $N^\ast$ (seed 0 / seed 1): training, mean and worst held-out error, filtered
fade ratio after the first stage and after the second (final model).  The fade control has one seed and no
fade ratio (its truth does not move).}
\label{tab:l1}
\begin{tabular}{llcccc}
\toprule
family & model & train & held-out (mean) & held-out (max) & fade (filtered) \\
\midrule
1D split, $N=5$ & stage 1    & 0.083 / 0.074 & 0.089 / 0.077 & 0.174 / 0.138 & 0.96 / 0.96 \\
                & final      & 0.003 / 0.002 & 0.024 / 0.015 & 0.095 / 0.069 & 0.95 / 0.96 \\
1D fade, $N=3$  & stage 1    & 0.068 / -- & 0.065 / -- & 0.071 / -- & -- \\
                & final      & 0.002 / -- & 0.015 / -- & 0.026 / -- & -- \\
2D split, $N=5$ & stage 1    & 0.152 / 0.143 & 0.132 / 0.138 & 0.285 / 0.272 & 0.91 / 0.91 \\
                & final      & 0.007 / 0.022 & 0.030 / 0.041 & 0.105 / 0.123 & 0.95 / 0.94 \\
2D spiral, $N=25$ & stage 1  & 0.125 / 0.042 & 0.124 / 0.041 & 0.158 / 0.075 & 0.92 / 0.94 \\
                & final      & 0.008 / 0.004 & 0.008 / 0.005 & 0.031 / 0.041 & 0.98 / 1.00 \\
\bottomrule
\end{tabular}
\end{table}

On the non-oscillatory families the second stage lowers the mean held-out error three- to fifteenfold and the fade
ratio drops by at most 0.01 (Table~\ref{tab:l1}).  On the spiral the mean error falls from 0.124 / 0.041 to 0.008 / 0.005, about registration's mean (0.006), without its kinks, and the worst from
0.158 / 0.075 to 0.031 / 0.041.  On the split families
and the spiral the worst-case error halves as well.  We posit
that the first-stage models were already in the basin of a displacing solution and had not converged in it.  The
training error falls further than the held-out error, so the interpolation gap becomes 1.8--8.7, and for this reason the pass
rule also accepts an absolute gap (\S\ref{sec:metrics}).  The training error falls on the spiral as well, but the held-out error falls with it
(interpolation gap 1.0 / 1.2).  On the oscillatory families (\S\ref{sec:osc2d}) the second stage gains less.  Over all $\mu$ the
mean held-out error falls 1.4- to 2.4-fold, the worst error falls only on the 1D family, and the interpolation gap rises to
6.7--19 (\S\ref{sec:oscl1} gives the errors over $\mu\ge0.1$).  On the rings, regularizer weights three and ten times larger in the second stage give the same models.

\section{Comparison with Hyper-SIREN, DeepONet, a Fourier neural operator and a convolutional autoencoder}\label{sec:baselines}

\subsubsection*{Hyper-SIREN}  Following~\cite{sitzmann2020siren}, a SIREN with $\omega_0 = 30$ (width 58, three hidden
layers) whose every weight and bias is produced from $\mu$ by its own ReLU hypernetwork head ($1\to256\to256\to$
numel), initialized so that the generated network starts as a SIREN-initialized SIREN; 2.35M parameters.

\subsubsection*{DeepONet}  Following~\cite{lu2021deeponet}, $u(\bx;\mu) = \sum_{k=1}^{p} b_k(\mu)\varphi_k(\bx) + b_0$ with $p = 200$ (the
LRNR's rank).  The coefficients $b_k$ are given by a ReLU network of $\mu$, $1\to256\to256\to256\to p$, and the
spatial modes $\varphi_k$ by an INR of $\bx$, a $\tanh$ network with four layers of width 800 on random Fourier features
$[\cos 2\pi\bB\bx, \sin 2\pi\bB\bx]$ ($\bB\in\R^{128\times d}$, entries $\mathcal N(0,2^2)$); 2.27M parameters.
Without the Fourier features, the INR of $\bx$ did not fit the 2D split snapshots (training error $\approx 0.9$ after 15k steps at learning rate
$10^{-3}$, 0.31 after 8k at $10^{-4}$); with the Fourier features and learning rate $10^{-4}$ it fits them to 2--4\%.
The output is linear in the $p$ spatial modes $\varphi_k$, so DeepONet is a learned linear reduced basis.

\subsubsection*{Fourier neural operator}  Following~\cite{li2021fno} (their FNO-1d and FNO-2d), the input function is the
constant field $\mu$ on the snapshot grid with the grid coordinates appended, lifted pointwise to width $c$, passed
through four Fourier layers $v\leftarrow \mathrm{GELU}(\mathcal K v + \mathcal W v)$~\cite{hendrycks2016gelu} (no GELU after the last; $\mathcal K$ keeps the lowest $k_{\max}$
frequencies per axis) and projected pointwise by $c\to128\to1$.  The domain is zero-padded by 8 points per axis, since
the fields are not periodic.  1D: $c = 96$, $k_{\max} = 32$, 2.41M parameters; 2D: $c = 32$, $k_{\max} = 12$ (the setting of the 2D
experiments of~\cite{li2021fno}), 2.37M parameters.  The FNO has no coordinate input, and its output is defined on the
training grid, so it is trained on whole snapshots only and evaluated on that grid.

\subsubsection*{Convolutional autoencoder}  Following~\cite{lee2020cae}, a convolutional encoder maps a snapshot on its
grid to a latent code of size 16 and a mirror-image decoder maps the code back; after training, the codes of the $N$
snapshots are interpolated in $\mu$, piecewise linearly (a natural cubic spline changes the held-out errors by at most
0.02), and decoded.  1D: the grid zero-padded from 2000 to 2048 points, six stride-2 convolutions (kernel 5; 32, 64,
128, 256, 288, 352 channels) to $352\times32$, a linear map to the code, and six stride-2 transposed convolutions
(kernel 4) and a final convolution back, 2.34M parameters; 2D: five stride-2 convolutions (kernel 3; 32, 64, 128, 192,
256 channels) to $256\times5\times5$ and the mirror image, 2.33M parameters; GELU after every layer but the code and the
output.  Like the FNO, it is defined on the training grid and is trained on whole snapshots only.

\subsubsection*{Training and sampling}  Adam ($10^{-4}$), $10{,}000N$ steps; each step fits one randomly chosen
snapshot either on all its points (\emph{whole}) or on one cell drawn exactly as for the LRNR (\emph{local}: the
same two-level cells at $n_d=2$, jitter, shuffled cycling and adaptive boost).  The sampling is the only difference
between the two variants.  The FNO follows the schedule of~\cite{li2021fno}: Adam at $10^{-3}$, halved five times,
weight decay $10^{-4}$, relative-$L^2$ loss, the same $10{,}000N$ steps with one whole snapshot per step; so does
the CAE, with the reconstruction of the chosen snapshot as its loss (at learning rates $3\cdot10^{-4}$ and $10^{-4}$ it
did not leave its initial plateau on the 2D split).  The CAE was run with two seeds at every family's $N^\ast$ (Burgers
excepted, the fade control with one seed).
DeepONet was run at every $N$ up to 7 with whole snapshots; Hyper-SIREN and DeepONet, with both samplings and two
seeds, and the FNO with two seeds were run at every family's $N^\ast$ (Burgers excepted; the FNO on the fade control
with one seed), and the FNO also at $N = 9$, 13, 17 and 21 on the spiral and at $N = 9$--29 on the wave packet.

\begin{table}[!htbp]
\centering\footnotesize
\caption{LRNR against parameter-matched baselines (seed 0 / seed 1).  \emph{local} = trained on the LRNR's cells.
At the LRNR's $N^\ast$ the LRNR is the final model; the spiral at $N=9$ is below $N^\ast$ and shows the first-stage
model (\S\ref{sec:training}).  Pass: the run passes the rule of \S\ref{sec:metrics} (for the LRNR, its first stage).  The FNO is trained on
whole snapshots only, as is the CAE (latent codes interpolated linearly).  Oscillatory families:
Table~\ref{tab:oscbaselines}.}
\label{tab:baselines}
\setlength{\tabcolsep}{4pt}
\begin{tabular}{llcccc}
\toprule
case & model & train & held-out & fade (filtered) & pass \\
\midrule
\multirow{7}{*}{2D split, $N=5$}
 & LRNR             & 0.007 / 0.022 & \underline{0.030 / 0.041} & \underline{0.95 / 0.94} & yes \\
 & Hyper-SIREN      & 0.017 / 0.028 & 0.535 / 0.547 & 0.35 / 0.30 & no \\
 & Hyper-SIREN local& 0.020 / 0.074 & 0.542 / 0.550 & 0.35 / 0.38 & no \\
 & DeepONet         & 0.026 / 0.038 & 0.542 / 0.563 & 0.11 / 0.15 & no \\
 & DeepONet local   & 0.064 / 0.088 & 0.550 / 0.555 & 0.16 / 0.15 & no \\
 & FNO              & 0.004 / 0.004 & 0.419 / 0.440 & 0.32 / 0.30 & no \\
 & CAE              & 0.001 / 0.005 & 0.565 / 0.553 & 0.37 / 0.32 & no \\
\midrule
\multirow{7}{*}{1D split, $N=5$}
 & LRNR             & 0.003 / 0.002 & \underline{0.024 / 0.015} & \underline{0.95 / 0.96} & yes \\
 & Hyper-SIREN      & 0.003 / 0.015 & 0.227 / 0.205 & 0.86 / 0.84 & no \\
 & Hyper-SIREN local& 0.007 / 0.022 & 0.180 / 0.191 & 0.82 / 0.83 & no \\
 & DeepONet         & 0.013 / 0.011 & 0.165 / 0.168 & 0.25 / 0.31 & no \\
 & DeepONet local   & 0.004 / 0.010 & 0.179 / 0.167 & 0.25 / 0.34 & no \\
 & FNO              & 0.002 / 0.002 & 0.095 / 0.130 & 0.58 / 0.44 & no \\
 & CAE              & 0.001 / 0.001 & 0.188 / 0.195 & 0.86 / 0.92 & no \\
\midrule
\multirow{6}{*}{1D fade, $N=3$}
 & LRNR             & 0.002 & \underline{0.015} & -- & yes \\
 & Hyper-SIREN / local (seed 0) & 0.027 / 0.002 & 0.142 / 0.172 & -- & no \\
 & DeepONet / local (seed 0) & 0.004 / 0.023 & 0.059 / 0.062 & -- & no / yes \\
 & FNO (seed 0)     & 0.001 & 0.059 & -- & no \\
 & CAE (seed 0)     & 0.001 & 0.205 & -- & no \\
\midrule
\multirow{7}{*}{2D spiral, $N=25$}
 & LRNR             & 0.008 / 0.004 & \underline{0.008} / \underline{0.005} & 0.98 / 1.00 & yes \\
 & Hyper-SIREN      & 0.009 / 0.032 & 0.019 / 0.035 & 0.80 / 0.77 & no \\
 & Hyper-SIREN local& 0.019 / 0.037 & 0.023 / 0.040 & 0.99 / 0.92 & yes \\
 & DeepONet         & 0.012 / 0.020 & 0.034 / 0.038 & 0.50 / 0.49 & no \\
 & DeepONet local   & 0.026 / 0.022 & 0.039 / 0.037 & 0.71 / 0.50 & no \\
 & FNO              & 0.005 / 0.005 & \underline{0.008} / 0.007 & 0.99 / 1.01 & yes \\
 & CAE              & 0.002 / 0.003 & 0.030 / 0.027 & 0.90 / 0.65 & s0 \\
\midrule
\multicolumn{6}{l}{\emph{not at $N^\ast$: first-stage LRNR}} \\
\multirow{4}{*}{2D spiral, $N=9$}
 & LRNR, stage 1    & 0.063 / 0.061 & 0.069 / \underline{0.073} & 0.50 / 0.45 & no \\
 & Hyper-SIREN      & 0.007 / 0.026 & 0.128 / 0.126 & 0.40 / 0.42 & no \\
 & Hyper-SIREN local& 0.012 / 0.018 & 0.123 / 0.108 & 0.52 / 0.49 & no \\
 & FNO              & 0.004 / 0.005 & \underline{0.066} / \underline{0.073} & 0.38 / 0.39 & no \\
\bottomrule
\end{tabular}
\end{table}

\begin{figure}[!htbp]
  \centering
  \includegraphics[width=\linewidth]{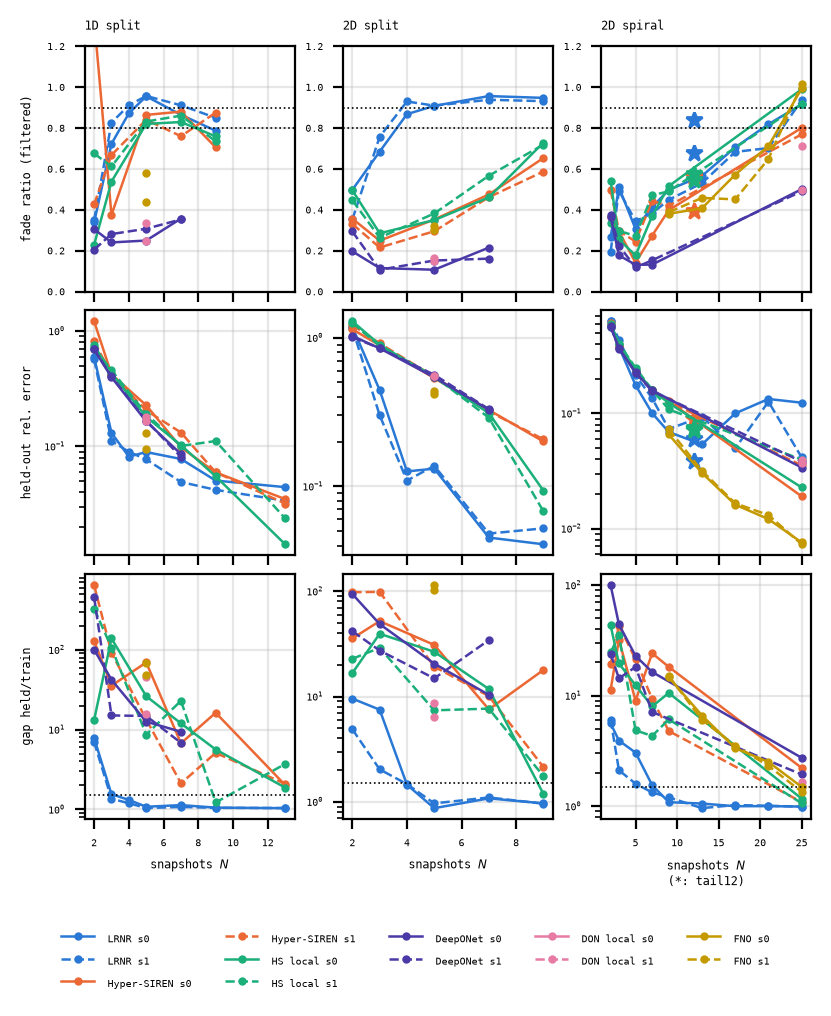}
  \caption{First-stage LRNR against Hyper-SIREN and DeepONet (whole snapshots and LRNR cells) and the FNO across $N$:
  filtered fade ratio, held-out error and interpolation gap; solid seed 0, dashed seed 1; stars: the spiral's tail12
  design.  The FNO was run at the families' $N^\ast$ and on the spiral at $N = 9$--25 (Table~\ref{tab:baselines}).  The
oscillatory families are compared in \S\ref{sec:osc2d} (Table~\ref{tab:oscbaselines}), the 2D wave and 2D Burgers
families in Appendix~\ref{app:nofade}.}
  \label{fig:baselines}
\end{figure}

Hyper-SIREN and DeepONet fit the training snapshots closely (training errors 0.2--10\%), about as tightly
as the final LRNR at $N^\ast$ and more tightly than the first-stage LRNR, and interpolate worse.  At the $N^\ast$ of
the three non-oscillatory moving-front families, 2 of the 24 Hyper-SIREN and DeepONet runs pass the rule: local Hyper-SIREN on the spiral
($N=25$, both seeds).  DeepONet passes on none.  On the split families ($N^\ast=5$) the baselines
memorize their snapshots, with held-out errors 7--18 times the final LRNR's; on the spiral, where local Hyper-SIREN
passes, the final LRNR is still 3--8 times more accurate (0.008 / 0.005 against 0.023 / 0.040), and the other Hyper-SIREN and DeepONet
runs fade there (fade ratios 0.49--0.80).  Below $N^\ast$ their held-out errors are 1.5--3 times the first-stage
LRNR's, and no baseline run passes at any $N$ below the LRNR's $N^\ast$ (Figure~\ref{fig:baselines}, which follows the
first-stage LRNR across $N$).  Of the failing runs, the nearest to passing is Hyper-SIREN on the 1D split, with fade ratios of 0.82--0.88 at $N =
5$--9 but gaps of 5--70; on the 2D split and on the spiral below $N^\ast$ the baselines' fade ratios stay at 0.1--0.6.  Between snapshots they show
the two-copy picture of a fade (Figures~\ref{fig:strip2d}, \ref{fig:strip1d}).  Training on the LRNR's cells changes
neither model's behavior in a consistent way.  On the fade control, where fading is correct, DeepONet (a learned
linear reduced basis) and the FNO are the most accurate baselines (0.06), and the final LRNR is four times more
accurate still (0.015).

The FNO fits its snapshots as tightly as the CAE (training errors 0.1--0.6\%) and, where the fronts move far between
snapshots, interpolates like the other baselines: on the splits at $N^\ast = 5$ its held-out errors are
0.09--0.44, its gaps 48--114 and its fade ratios 0.30--0.58, and on the fade control its gap is 52.  On the spiral it
passes at $N^\ast = 25$ with both seeds and is about as accurate as the final LRNR (0.008 / 0.007 against 0.008 /
0.005); it passes at no smaller $N$ (at $N = 9$--21 its held-out errors fall from 0.07 to 0.012, slightly below the
first-stage LRNR's at $N = 9$, but its fade ratios stay at 0.38--0.71 and its gaps at 2.3--15), so its $N^\ast$ there is the
LRNR's.  The FNO thus reaches the accuracy of the final LRNR once the snapshots are dense enough that little motion falls
between them, but it does not lower the snapshot count at which displacement starts.

The CAE fits its snapshots to 0.1--0.5\% and is among the least accurate of the neural baselines between them.  Its
held-out errors are 4--19
times the final LRNR's on the four non-oscillatory families, with gaps of 9--390.  On the 2D split the
decoded path between two codes fades (fade ratios 0.37 / 0.32); on the 1D split it moves (0.86 / 0.92), but its held-out
error is 0.19.  On the spiral, where little motion falls between snapshots, seed 0 passes (0.030, against the final
LRNR's 0.008) and seed 1 fades (0.65).  Interpolating the codes by a cubic spline instead of piecewise linearly
does not change any of these outcomes.

\subsubsection*{Errors between the snapshots}  A figure with three to five values of $\mu$ can hide or exaggerate what happens in
between.  We therefore scanned every model on a fine grid of $\mu$ (401 values in 1D, 161 in 2D) for its relative error
and its amplitude $\|u\|/\|u_{\rm truth}\|$, and chose the frames to show by a fixed rule: the $\mu$ of
Hyper-SIREN's largest error in each of the four gaps where that error is largest, plus the LRNR's worst frame if it lies in
another gap (Figures~\ref{fig:scan_sp1}--\ref{fig:scan_spr}).  The baselines reproduce their training snapshots almost exactly, and their
error rises to a peak in every gap.  In the
middle half of the gaps, Hyper-SIREN and DeepONet (with either sampling) average 0.71--0.72 on the 2D split and the FNO
0.55, against the final LRNR's 0.04; on the 1D split, 0.22--0.31 and 0.13 against 0.03.  The LRNR's error varies much less across a gap.  The first-stage LRNR's error is nearly
constant in $\mu$, including at the snapshots, which it does not fit exactly, so within a tenth of a gap of a snapshot
it is slightly worse than the baselines (0.08 against 0.02--0.04 on the 1D split); the final LRNR fits the snapshots
(0.008 there against the baselines' 0.02--0.04) and is more accurate everywhere.  Appendix~\ref{app:movies} shows every
comparison at fine resolution in $\mu$: 3D surfaces $u(x,\mu)$ on 241 values of $\mu$ and error waterfalls in 1D,
stacked half-maximum contours in $(x, y, \mu)$ and frame grids with four frames per gap in 2D, for each family (the 1D oscillatory family: \S\ref{sec:oscbetween}).

\section{Comparison with classical interpolation methods}\label{sec:classical}

Transport-dominated families have received significant attention in model reduction, and several classical interpolants were designed
for the failure of \S\ref{sec:intro}.  We implemented six, all deterministic and cheap (seconds to minutes on
a CPU), and scored them with the metrics of \S\ref{sec:metrics} on the same snapshots
(details and the full sweep in Appendix~\ref{app:classical}):
\begin{itemize}
  \item \emph{POD, linear}: full-rank POD with linearly interpolated coefficients, i.e., the cross-fade of neighboring
        snapshots (fade ratio 0 by construction, the floor);
  \item \emph{POD-I}: the same with natural cubic splines in $\mu$ \cite{ly2001pod};
  \item \emph{registration}: per gap, a smooth displacement field aligning the two end snapshots, fitted coarse to fine,
        and the symmetric morph $u(\mu) = (1-s)u_i(\bx - s\bd) + su_{i+1}(\bx + (1-s)\bd)$ (in the spirit of
        registration-based reduced models, e.g.,~\cite{taddei2020registration});
  \item \emph{shifted POD} (sPOD~\cite{reiss2018spod}): the snapshots as a sum of several co-moving frames, each low-rank,
        $u(\bx;\mu) \approx \sum_k T_{\bc_k(\mu)} q_k(\bx;\mu)$, computed by singular-value and soft thresholding as in the
        robust sPOD of~\cite{krah2025robust}; once with the shifts $\bc_k(\mu)$ provided and once with them fitted along with
        the frames;
  \item \emph{optimal transport} (OT): displacement interpolation~\cite{mccann1997convexity} between neighboring snapshots, the positive
        and negative parts transported separately (exact in 1D, an entropic plan~\cite{cuturi2013sinkhorn} in 2D);
  \item \emph{dynamic mode decomposition} (DMD~\cite{schmid2010dmd,tu2014dmd}), with $\mu$ as time: the rank-$r$ linear map
        that best takes each snapshot to the next, its eigenvalues $\lambda_k$ and modes $\phi_k$, and
        $u(\mu) = \RePart\sum_k b_k\phi_ke^{\omega_k\mu}$ with $\omega_k = \log\lambda_k/\Delta\mu$, the
        amplitudes $b_k$ fitted to all snapshots by least squares~\cite{jovanovic2014sparsity}.  Unlike the others it is
        a single model of the whole family, smooth in $\mu$, and it does not reproduce the snapshots.
\end{itemize}
We did not run transported subspaces~\cite{rim2023transported} as a separate baseline, because the LRNR is itself a genuine
extension of that method via the inverse-bias trick (\S\ref{sec:invbias},~\cite{rim2026entropy}), in which the transported profiles, the transports and their
$\mu$-dependence are all learned by one network instead of being prescribed.  The comparison below is therefore with the
classical methods that the LRNR does not contain as a special case.

Every method's hyperparameters (regularization weight, number of frames and rank, thresholds) were chosen per
example and $N$ by the held-out error itself, which is an oracle choice and favors the baselines.

\subsubsection*{Limits of a direct comparison}  The numbers below are not a like-for-like comparison, for six reasons.
The first five favor the classical methods in the numbers or make their errors look smaller than their defects; the
last affects both sides.
\begin{enumerate}
  \item \emph{Shifts provided to sPOD.}  sPOD needs, before it computes anything, the
        number of co-moving frames, which structure each frame follows, and the shift $\bc_k(\mu)$ of every frame at
        every $\mu$, including every held-out $\mu$ at which it is evaluated.  These shift parameters are the
        transport.  Once they are known, the remainder in each frame is a slowly changing profile that any linear method
        interpolates well, provided the snapshots determine how they split into frames; on the 1D
        oscillatory split they do not at $N^\ast$ (Appendix~\ref{app:classical}).  Here we provided them exactly, from the analytic construction of each family (the two
        copies' straight paths for the split families, the arc's center and tip for the spiral) and, for Burgers,
        from the reference solution itself.  The LRNR is given no shifts and learns where features go
        from the snapshots alone.  When sPOD must estimate the shifts as well, the problem becomes non-convex and the
        same model fails (held-out error 0.39 on the 2D split against 0.042 with the shifts provided, Table~\ref{tab:classical}).
        Where no part of the motion is a translation, as for the wave equation's radially expanding ring, there is no
        shift to provide and sPOD reduces to POD; for the spiral, which stretches and rotates, the best available
        shifts describe the motion only partly.  We therefore treat the provided-shift sPOD as an oracle upper
        bound for its class.
  \item \emph{Exact interpolation against regression.}  POD, registration and OT reproduce the training snapshots
        exactly; the LRNR is a regression model with a training error of 5--15\% at small $N$.  The gap test
        $\text{held-out}/\text{train} \le 1.5$ is undefined for an exact interpolant, and a held-out error comparison
        mixes the LRNR's fitting error, which dominates its held-out error at $N^\ast$, with the baselines'
        interpolation error.
  \item \emph{Two-snapshot maps against one model.}  Registration, OT and linear POD build $N-1$ independent maps,
        each from the two snapshots at the ends of one gap.  The maps share no information across the family, the stored model grows
        with $N$, and the information in other snapshots (for example, that a hump which has started to split will end as two
        intact copies) is not used.  The interpolated path is only continuous in $\mu$: its velocity
        $\partial_\mu u$ is discontinuous at every snapshot.  We measure this with a kink index, the relative jump of the
        one-sided $\mu$-derivatives at the interior snapshots; it is about 0.05 for the true family
        and for the LRNR, a single network driven by one smooth latent curve, and 0.5--1.7 for the two-snapshot maps
        (Table~\ref{tab:classical}).
  \item \emph{Defects that a norm hides.}  OT splits a hump by tearing it: between the snapshots it leaves a thin
        filament of mass between the separating copies; registration, whose warps cannot change topology, folds the
        hump at the split instead (Figure~\ref{fig:strip2d}).  Both defects are small in $L^2$ and both are
        qualitatively wrong.
  \item \emph{Method-specific assumptions.}  OT transports non-negative mass: signed fields must be split into parts
        whose masses are interpolated separately, and an entropic plan blurs at the scale of the grid.  Registration
        assumes a diffeomorphic deformation.  sPOD assumes translations.  DMD assumes that the snapshots follow linear
        dynamics in $\mu$, which a translating front does not: a translate is not a linear function of a few snapshots,
        so the rank needed grows with the motion.  Each assumption holds on some of our families
        and fails on others, and the method has to be chosen to match; the LRNR is trained by one recipe on all of them.
  \item \emph{Oracle hyperparameters and unequal cost.}  The baselines' settings were chosen on the test data, the
        LRNR's are fixed across examples.  Conversely, the classical methods cost seconds to minutes on a CPU and a LRNR
        hours on a GPU, so the comparison is not cost-matched.
\end{enumerate}

\subsubsection*{The 2D split at $N=5$}  Table~\ref{tab:classical} and Figure~\ref{fig:strip2d} show the case the
LRNR first passes.  On this family (two copies of a non-negative hump moving along straight lines), sPOD with the
exact shifts matches the LRNR's mean error and has a smaller worst error, and OT and registration have three to four and a
half times the LRNR's mean error.
Their defects are the ones listed above: the tear and the fold at the split, and kinks at the snapshots.  The other
families follow; the full sweep is in Appendix~\ref{app:classical}.

\subsubsection*{Comparison across the families at $N^\ast$}  
\begin{figure}[!htbp]
  \centering
  \includegraphics[width=\linewidth]{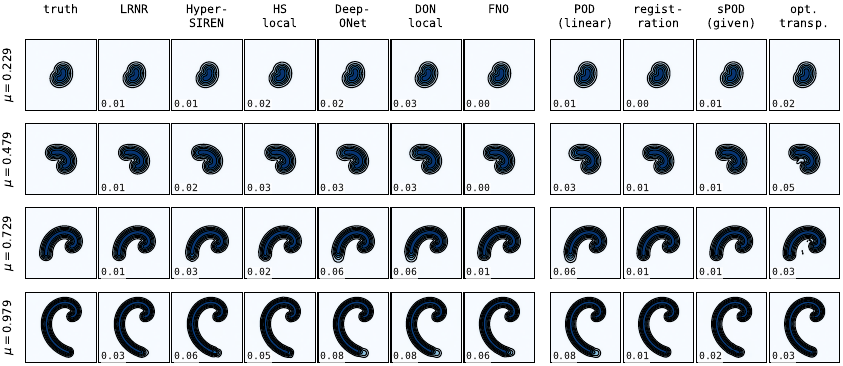}
  \caption{2D spiral, $N^\ast=25$, at four gap midpoints: truth, final LRNR, the neural baselines and the classical
  interpolants (seed 0), contoured at the truth's levels; numbers are relative errors.  At this spacing every method
  follows the arc; the errors differ at the growing tip ($\mu = 0.729, 0.979$).}
  \label{fig:comparespr}
\end{figure}

Table~\ref{tab:minimax} compares the models on the four
non-oscillatory families at the smallest $N$ where the LRNR passes, by mean and by worst held-out error over the fine $\mu$ grid (101 points and the
gap midpoints).  Each classical method has large errors on some family.  Registration is the most
accurate on the spiral, sPOD with provided shifts on the 2D split (where it is given the exact paths of both copies),
and POD and DMD on the fade control.  Registration and OT impose
motion on the fade control, whose truth has none (worst error 1.3 and 1.0); POD cross-fades every moving-front family.  Registration,
the closest classical method on the 1D split, has three to five times the LRNR's mean error there, and its
error peaks inside some gaps at up to 3.6 / 5.0 times the LRNR's worst (seed 0 / seed 1).  The LRNR's mean error is the lowest on the 1D split and level with the best classical
method on the 2D split (sPOD with provided shifts, which the LRNR matches without being given them) and the spiral (registration, Figure~\ref{fig:comparespr}), and its worst error is at most 0.12
on each of these four families, at least six times below the worst case of any classical method that applies to all of them (fitted
sPOD: 0.76; POD, POD-I: 0.82--0.85; registration, OT, DMD: 1.0--1.3).  DMD is exact on the fade control, whose truth is linear in $\mu$, and
fails on every moving family: on the splits it does not fit its snapshots (training error
0.61--1.00), because neighboring snapshots of a moving front are nearly orthogonal and no low-rank linear map takes
one to the next; on the spiral, whose snapshots overlap, it reaches 0.12, fifteen to twenty-four times the LRNR's
error.

\begin{table}[!htbp]
\centering\small
\caption{2D split, $N=5$: final LRNR (seed 0 / seed 1) against the classical interpolants, each at the setting with the lowest
held-out error.  Kink index: relative jump of $\partial_\mu u$ at the interior snapshots (truth: 0.045).  $^\dagger$Given
the exact paths of both copies at every $\mu$; an oracle, not a like-for-like competitor.}
\label{tab:classical}
\begin{tabular}{lcccc}
\toprule
model & train & held-out & fade (filtered) & kink \\
\midrule
LRNR                          & 0.007 / 0.022 & \underline{0.030} / 0.041 & 0.95 / 0.94 & 0.045 / 0.045 \\
POD, linear                   & 0 & 0.529 & 0.00 & 1.70 \\
POD-I (cubic spline)          & 0 & 0.546 & 0.19 & 0.047 \\
registration                  & 0 & 0.122 & 0.98 & 0.49 \\
sPOD, shifts fitted           & 0.287 & 0.393 & 0.21 & -- \\
sPOD, shifts provided$^\dagger$ & 0.001 & 0.042 & \underline{0.98} & 0.053 \\
optimal transport             & 0 & 0.134 & 0.97 & 0.92 \\
DMD                           & 0.998 & 0.996 & 0.14 & 0.017 \\
\bottomrule
\end{tabular}
\end{table}

\begin{table}[!htbp]
\centering\small
\caption{Mean / worst relative held-out error at each family's $N^\ast$ (LRNR: seed 0, seed 1).  Worst over the fine
$\mu$ grid.  LRNR: final models (after the second training stage, \S\ref{sec:l1}).  Underlined: smallest worst-case error per family; last column: worst case over
the families the method applies to.  $^\dagger$Not a like-for-like competitor: given the exact shift of every frame at
every $\mu$, held-out ones included, which is the transport the LRNR learns from the snapshots (\S\ref{sec:classical}).}
\label{tab:minimax}
\setlength{\tabcolsep}{4pt}
\begin{tabular}{lccccc}
\toprule
 & 1D split & 1D fade & 2D split & spiral & worst \\
 & $N^\ast=5$ & $N^\ast=3$ & $N^\ast=5$ & $N^\ast=25$ & \\
\midrule
LRNR, s0          & .024 / .095 & .015 / .026 & .030 / .105 & .008 / .031 & \underline{.105} \\
LRNR, s1          & .015 / \underline{.069} & -- & .041 / .123 & .005 / .041 & .123 \\
POD, linear       & .149 / .362 & .000 / \underline{.000} & .529 / .816 & .028 / .081 & .816 \\
POD-I (cubic)     & .119 / .352 & .000 / \underline{.000} & .546 / .851 & .019 / .083 & .851 \\
registration      & .074 / .346 & .408 / 1.305 & .122 / .438 & .006 / \underline{.012} & 1.305 \\
sPOD, fitted      & .391 / .764 & .180 / .430 & .393 / .590 & .105 / .537 & .764 \\
sPOD, provided$^\dagger$ & .110 / .211 & n/a & .042 / \underline{.073} & .008 / .023 & (.211) \\
OT                & .146 / .344 & .534 / 1.031 & .134 / .247 & .022 / .047 & 1.031 \\
DMD               & .547 / 1.233 & .000 / \underline{.000} & .996 / 1.001 & .121 / .266 & 1.233 \\
\bottomrule
\end{tabular}
\end{table}

\section{The oscillatory families}\label{sec:osc2d}

The 1D oscillatory split, the 2D rings and the 2D wave packet (\S\ref{sec:families}, Figure~\ref{fig:families})
combine the split families' motion with a carrier whose crests must travel with their copy while its frequency
changes.  They are reported in this section, with their own pass rule and their own tables, and are not pooled with
the four non-oscillatory families of \S\ref{sec:transition}--\S\ref{sec:classical}: the problem is harder in ways
that change what the metrics mean, and an error or a fade ratio on these families is not commensurate with the same
number on the others.  The models, the recipe (with Chebyshev degree-9 input features) and the baselines are those of
the preceding sections.

\subsection{Difficulties specific to the oscillatory families}\label{sec:oscchallenge}

\subsubsection*{Undetermined carrier motion at small $N$}  Between neighboring snapshots at $N=5$
each copy travels about 3.5 carrier wavelengths and its phase shifts by about 2.4 rad.  The snapshots sample the
carrier's phase far below its rate of change, so many interpolants agree with them equally well: a packet that moves with its
copy, one that slides its crests the wrong way, and one that fades out and in.  The coefficient count of
\S\ref{sec:snapcount} reflects this: the displacing construction needs about twice as many coefficients as for a plain
hump ($n_T = 9$--10 against 4--6), and the threshold $N_c$ rises from 4 to 7--9 (Table~\ref{tab:snapcount}).

\subsubsection*{Relative error of a small misplacement}  For a carrier $\cos\theta$ displaced by a fraction
$\delta$ of a wavelength, $\|\cos\theta - \cos(\theta + 2\pi\delta)\|/\|\cos\theta\| = 2|\sin\pi\delta|$ over whole
periods: a tenth of a wavelength gives a relative error of 0.62, a sixth gives 1.0 (the error of predicting zero),
and a quarter gives 1.41.  For the families' carriers ($k = 10$--20) and humps of radius 0.2--0.4, shifting a packet
by a tenth of a wavelength costs 0.62--0.64 in relative error, the same shift of the hump without its carrier 0.05--0.18.
Relative $L^2$ errors on the oscillatory families are therefore three to twelve times larger than on the other
families for the same accuracy in position, and are compared only among models on the same oscillatory family.

\subsubsection*{Phase sensitivity of the field fade ratio}  For a carrier that travels with its copy, the cross-fade
of two phase-shifted carriers is far from both, so an interpolant that does not move the packet can score well above
1 on the field fade ratio of \S\ref{sec:metrics}, and one that moves it correctly can score low when its crests are
slightly misplaced.  On the 1D family the field fade ratio of the LRNR does not reach 0.9 at any $N$ up to 25 (0.84 at
best), although from $N = 9$ its packets move with their copies (the final model at $N=9$:
Figure~\ref{fig:oscstrip1d}).

\subsubsection*{Overlap of the copies at small $\mu$}  Below $\mu\approx0.1$ the two copies have not yet separated and their
carriers interfere; the data hardly determine how the hump splits, and every model's largest errors lie there
(\S\ref{sec:oscmeasure}).  We regard errors in this range as a property of the problem rather than of the interpolant.

\subsection{Pass rule and reporting}\label{sec:oscrule}

For these families the fade test is applied to the envelope, the amplitude of the analytic signal (Hilbert transform
in 1D, Riesz transform~\cite{felsberg2001monogenic} in 2D), which moves with the copy and does not oscillate; cells count when the truth's envelope
departs from its own cross-fade by $d_{\rm truth}\ge0.05$.  All three conditions of the pass rule (training error at
most 0.30, gap at most 1.5 or 0.05 in absolute terms, envelope fade ratio at least 0.9) are evaluated over
$\mu\ge0.1$ only, on both seeds, and on the first training stage, as for the other families.  The envelope test is
one-sided: a cross-fade of phase-shifted carriers can have an envelope ratio well above 1 (2.45 for linear POD on the
1D family at $N=9$), so a high ratio is not evidence of displacement, while a ratio below 0.9 still is evidence of
fading; the gap condition rejects such interpolants.  Every number in this section (errors, gaps, worst
cases, fade ratios) is over $\mu\ge0.1$ and uses the envelope, and the tables of this section are kept separate from
those of the other families.

\subsection{The transition}\label{sec:osctransition}

\begin{figure}[!htbp]
  \centering
  \includegraphics[width=\linewidth]{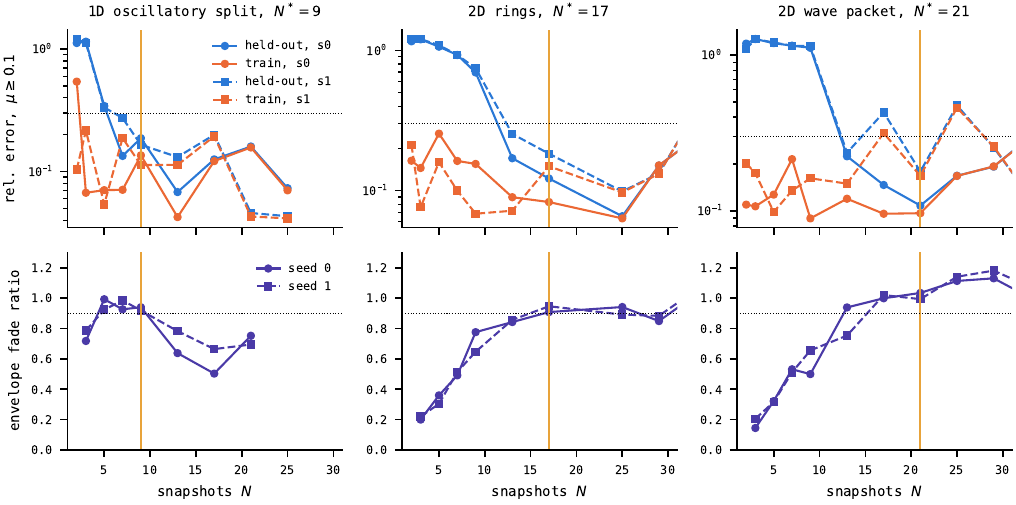}
  \caption{Oscillatory families (first-stage LRNR): training and held-out error over $\mu\ge0.1$ (top; dotted: the 0.30
  fit allowance) and envelope fade ratio (bottom; dotted: 0.9) against $N$; vertical lines mark $N^\ast$.  Missing
  envelope points: no gap moves the truth's envelope by at least 0.05.}
  \label{fig:osctransition}
\end{figure}

Table~\ref{tab:osctransition} and Figure~\ref{fig:osctransition} give the sweep.  For $N\le9$ every 2D run memorizes its
snapshots: training errors of 7--26\%, held-out errors of 70--127\% (close to or worse than predicting zero), gaps of 4--16 and
envelope ratios of 0.14--0.78; the 1D family memorizes up to $N=5$ (held-out 0.33--1.2).

\begin{table}[!htbp]
\centering\small
\caption{Oscillatory families: snapshot-count transition (first-stage LRNR, seeds 0/1).  Errors over $\mu\ge0.1$;
envelope fade ratio.  $N^\ast$: 1D 9, rings 17, wave packet 21.}
\label{tab:osctransition}
\begin{tabular}{lccccc}
\toprule
family & $N$ & train & held-out & envelope fade & passes \\
\midrule
1D oscillatory & 5 & 0.070 / 0.054 & 0.343 / 0.334 & 0.99 / 0.92 & no \\
               & 7 & 0.071 / 0.188 & 0.134 / 0.274 & 0.93 / 0.99 & s1 \\
               & 9 & 0.136 / 0.113 & 0.187 / 0.165 & \underline{0.94 / 0.92} & yes \\
               & 13 & 0.042 / 0.113 & 0.068 / 0.131 & 0.64 / 0.78 & no \\
               & 21 & 0.157 / 0.043 & 0.160 / 0.046 & 0.75 / 0.69 & no \\
2D rings       & 9 & 0.155 / 0.068 & 0.696 / 0.746 & 0.78 / 0.65 & no \\
               & 13 & 0.090 / 0.072 & 0.170 / 0.254 & 0.84 / 0.86 & no \\
               & 17 & 0.083 / 0.149 & 0.122 / 0.183 & \underline{0.91 / 0.95} & yes \\
               & 25 & 0.063 / 0.097 & 0.066 / 0.099 & 0.94 / 0.89 & s0 \\
               & 29 & 0.151 / 0.132 & 0.151 / 0.132 & 0.85 / 0.88 & no \\
               & 33 & 0.237 / 0.363 & 0.238 / 0.366 & 1.02 / 1.05 & s0 \\
2D wave packet & 9 & 0.090 / 0.162 & 1.114 / 1.142 & 0.50 / 0.66 & no \\
               & 13 & 0.120 / 0.150 & 0.224 / 0.236 & 0.94 / 0.75 & no \\
               & 17 & 0.096 / 0.316 & 0.147 / 0.429 & 1.00 / 1.02 & s0 \\
               & 21 & 0.097 / 0.168 & 0.108 / 0.179 & \underline{1.03 / 0.99} & yes \\
               & 25 & 0.168 / 0.460 & 0.168 / 0.475 & 1.11 / 1.14 & s0 \\
               & 29 & 0.193 / 0.259 & 0.192 / 0.255 & 1.13 / 1.18 & yes \\
               & 33 & 0.295 / 0.122 & 0.300 / 0.122 & 1.00 / 1.07 & yes \\
\bottomrule
\end{tabular}
\end{table}

\begin{itemize}
  \item \emph{1D oscillatory}: $N^\ast = 9$, at the count threshold $N_c = 9$.  At $N=5$ the envelope already moves
        (0.99 / 0.92) but the models memorize (gap 4.9 / 6.2); at $N=7$ seed 1 passes and seed 0 misses the gap test
        (held-out 0.134 against training 0.071); at $N=9$ both pass.  At $N = 13$--21 the envelope ratio falls again, to
        0.50--0.78, in single gaps, while the held-out error stays at 0.046--0.20; at $N=25$ no gap moves the
        truth's envelope by 0.05 and the ratio is undefined.  This is the behavior of the 1D split above its $N^\ast$
        (\S\ref{sec:transition}), with a worst-cell statistic at fine spacing.  For this reason $N^\ast$ is the first pass.
  \item \emph{2D rings}: $N^\ast = 17$ (per-gap motion about 0.44 carrier wavelengths).  At $N=25$ seed 1 and at
        $N=29$ both seeds fall just short on the envelope (0.85--0.89) with held-out errors of 0.10--0.15.  At $N=33$ the
        envelope moves fully again (1.02 / 1.05), but seed 1 misses the fit allowance (training error 0.36): the fit
        worsens from $N=25$ (0.06--0.10) to $N=33$ (0.24--0.36), although the budget grows in proportion to $N$.
  \item \emph{2D wave packet}: $N^\ast = 21$.  The envelope moves from $N=17$ on (ratios 0.99--1.18), but at $N=17$
        seed 1 does not fit its snapshots (training error 0.32 against the 0.30 allowance); at $N=21$ both seeds pass.
        Above $N^\ast$ seed 1 misses the fit allowance again at $N=25$ (0.46), and both seeds pass at $N=29$ and 33
        (seed 0 at $N=33$ with a training error of 0.295, just inside the allowance).
\end{itemize}
With an envelope threshold of 0.85 instead of 0.9, no $N^\ast$ changes; only the rings above $N^\ast$ gain seeds (seed 1
at $N=25$ and at $N=29$), while the 1D family's dips at $N=13$--21 stay below either threshold.  Measured against the
count thresholds, the 1D family reaches $N_c$, and the 2D families need two to three times more
snapshots than $N_c = 7$--8.

\subsection{The second stage}\label{sec:oscl1}

\begin{table}[!htbp]
\centering\small
\caption{Oscillatory families: the two training stages at $N^\ast$ (seed 0 / seed 1), over $\mu\ge0.1$; envelope fade
ratio.}
\label{tab:oscl1}
\begin{tabular}{llcccc}
\toprule
family & model & train & held-out (mean) & held-out (max) & envelope fade \\
\midrule
1D oscillatory, $N=9$ & stage 1 & 0.136 / 0.113 & 0.187 / 0.165 & 0.682 / 0.565 & 0.94 / 0.92 \\
                & final      & 0.005 / 0.006 & 0.062 / 0.083 & 0.416 / 0.479 & 0.91 / 0.93 \\
2D rings, $N=17$ & stage 1   & 0.083 / 0.149 & 0.122 / 0.183 & 0.559 / 0.736 & 0.91 / 0.95 \\
                & final      & 0.009 / 0.020 & 0.070 / 0.081 & 0.581 / 0.661 & 0.89 / 0.88 \\
2D wave packet, $N=21$ & stage 1 & 0.097 / 0.168 & 0.108 / 0.179 & 0.264 / 0.467 & 1.03 / 0.99 \\
                & final      & 0.010 / 0.010 & 0.026 / 0.034 & 0.189 / 0.408 & 1.01 / 1.00 \\
\bottomrule
\end{tabular}
\end{table}

The second stage (\S\ref{sec:training}) lowers the mean held-out error 1.7- to fivefold (Table~\ref{tab:oscl1}).  On
the wave packet ($N=21$) the training error falls to 1.0\% and the held-out error to 0.026 / 0.034, so held-out minus
training error is 0.016 / 0.024 and the final model would pass the rule again.  On the 1D family ($N=9$) and the rings ($N=17$) the training error falls to
0.5--2\% while the held-out error stays at 6--8\%, so held-out minus training error is 0.056--0.077, just above the
absolute gap allowance of 0.05; on the rings the envelope fade ratio also falls slightly, from 0.91 / 0.95 to 0.89 /
0.88.  Models saved every 10,000 steps show that both errors reach their final values within the first 10,000 steps and
do not change afterward.  The held-out error does not rise, so the second stage does not overfit in the usual sense; it
removes the fitting error at the snapshots and leaves the interpolation error between them, which at these spacings is
the larger part.  Raising the second stage's loss weights from a tenth of the first stage's to three tenths or to the
full first-stage values changes none of these numbers in the third decimal; the structural penalties stay at their
first-stage values throughout.  Since the pass rule is applied to the first stage, these two families keep their
$N^\ast$, and we report the second-stage model as the more accurate interpolant.  The worst-case error falls less than
the mean: by 15--40\% on the 1D family, by 13--28\% on the wave packet and by 10\% on the rings for seed 1 (seed 0 rises slightly), at $\mu =
0.10$--0.17, the edge of the overlap region.

\subsection{Hyper-SIREN, DeepONet, the FNO and the CAE}\label{sec:oscbaselines}

\begin{table}[!htbp]
\centering\footnotesize
\caption{Oscillatory families at $N^\ast$: final LRNR against the parameter-matched baselines (seed 0 / seed 1), over
$\mu\ge0.1$; envelope fade ratio.  Pass: the rule of \S\ref{sec:oscrule} (for the LRNR, its first stage).  The FNO and the CAE are trained on whole snapshots only.  $^\ddagger$The CAE's output stayed at zero (training error
1.00) for the whole run, all 90,000 steps on the 1D family and 210,000 on the wave packet.}
\label{tab:oscbaselines}
\setlength{\tabcolsep}{4pt}
\begin{tabular}{llccccc}
\toprule
case & model & train & held-out & gap & envelope fade & pass \\
\midrule
\multirow{7}{*}{1D oscillatory, $N=9$}
 & LRNR             & 0.005 / 0.006 & \underline{0.062 / 0.083} & 11.5 / 13.0 & 0.91 / 0.93 & yes \\
 & Hyper-SIREN      & 0.012 / 0.065 & 0.578 / 0.550 & 49.1 / 8.4 & 2.02 / 2.12 & no \\
 & Hyper-SIREN local& 0.106 / 0.063 & 0.567 / 0.578 & 5.3 / 9.2 & 1.94 / 1.78 & no \\
 & DeepONet         & 0.007 / 0.020 & 0.537 / 0.545 & 74.5 / 27.4 & 2.09 / 1.85 & no \\
 & DeepONet local   & 0.010 / 0.029 & 0.519 / 0.533 & 53.4 / 18.7 & 1.80 / 1.98 & no \\
 & FNO              & 0.003 / 0.007 & 0.516 / 0.495 & 160.7 / 72.4 & 1.87 / 1.82 & no \\
 & CAE              & 1.000 / 1.000 & 1.000 / 1.000 & -- & -- & no$^\ddagger$ \\
\midrule
\multirow{7}{*}{2D rings, $N=17$}
 & LRNR             & 0.009 / 0.020 & \underline{0.070 / 0.081} & 7.8 / 4.1 & 0.89 / 0.88 & yes \\
 & Hyper-SIREN      & 0.054 / 0.075 & 0.445 / 0.298 & 8.3 / 4.0 & 0.95 / 0.93 & no \\
 & Hyper-SIREN local& 0.108 / 0.259 & 0.335 / 0.268 & 3.1 / 1.0 & 0.88 / 0.87 & no \\
 & DeepONet         & 0.076 / 0.084 & 0.547 / 0.548 & 7.2 / 6.5 & 1.05 / 1.15 & no \\
 & DeepONet local   & 0.117 / 0.044 & 0.567 / 0.538 & 4.8 / 12.3 & 1.06 / 1.24 & no \\
 & FNO              & 0.013 / 0.015 & 0.179 / 0.115 & 13.6 / 7.6 & 0.87 / 0.93 & no \\
 & CAE              & 0.065 / 0.005 & 0.598 / 0.525 & 9.1 / 98.0 & 1.30 / 1.25 & no \\
\midrule
\multirow{7}{*}{\shortstack[l]{2D wave packet,\\ $N=21$}}
 & LRNR             & 0.010 / 0.010 & \underline{0.026} / 0.034 & 2.5 / 3.6 & 1.01 / 1.00 & yes \\
 & Hyper-SIREN      & 0.026 / 0.116 & 0.036 / 0.116 & 1.4 / 1.0 & 1.00 / 1.01 & yes \\
 & Hyper-SIREN local& 0.085 / 0.112 & 0.221 / 0.111 & 2.6 / 1.0 & 1.28 / 0.95 & s1 \\
 & DeepONet         & 0.042 / 0.098 & 0.624 / 0.652 & 14.9 / 6.7 & 3.26 / 3.33 & no \\
 & DeepONet local   & 0.083 / 0.197 & 0.635 / 0.627 & 7.6 / 3.2 & 3.19 / 4.36 & no \\
 & FNO              & 0.011 / 0.013 & 0.043 / \underline{0.023} & 3.8 / 1.8 & 0.96 / 0.97 & yes \\
 & CAE              & 1.000 / 1.000 & 1.000 / 1.000 & -- & -- & no$^\ddagger$ \\
\bottomrule
\end{tabular}
\end{table}

\begin{figure}[!htbp]
  \centering
  \includegraphics[width=\linewidth]{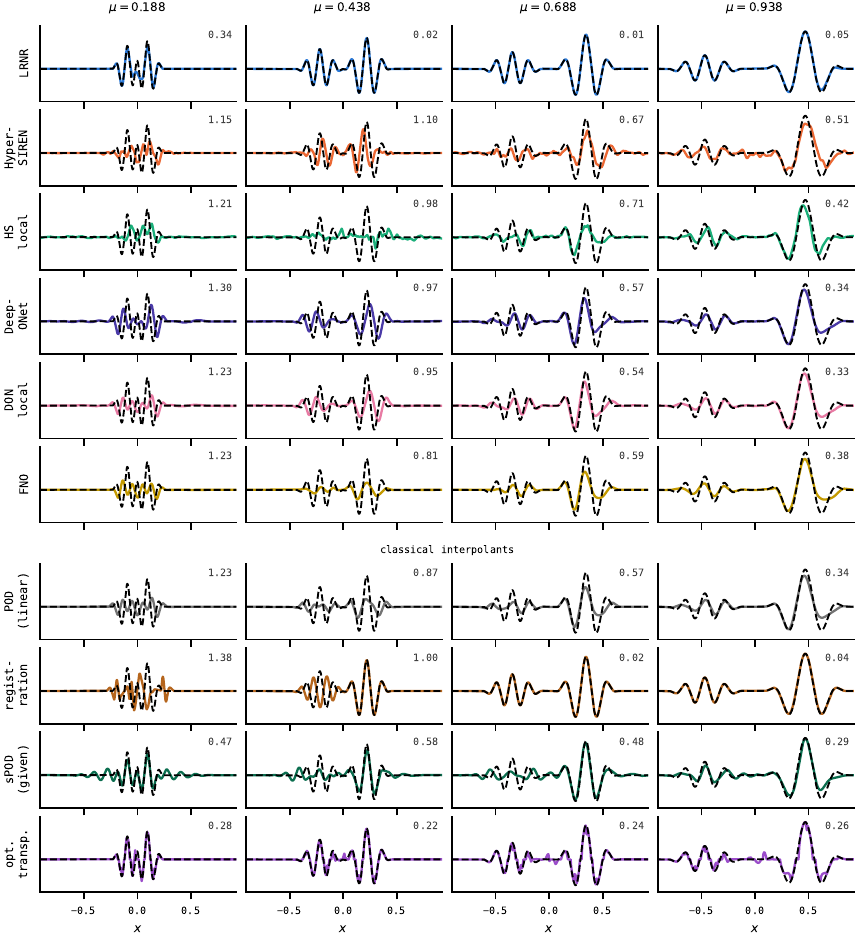}
  \caption{1D oscillatory split, $N^\ast = 9$, at four gap midpoints with $\mu\ge0.1$: final LRNR, the five neural
  baselines other than the CAE and four classical interpolants (solid, seed 0 for the trained models) against the truth (dashed); numbers
  are relative errors.  Registration misaligns the carrier at $\mu = 0.188$ and $0.438$ (errors 1.38, 1.00); OT and sPOD with
  provided shifts move the packets with misphased crests (errors 0.22--0.58).}
  \label{fig:oscstrip1d}
\end{figure}

\begin{figure}[!htbp]
  \centering
  \includegraphics[width=\linewidth]{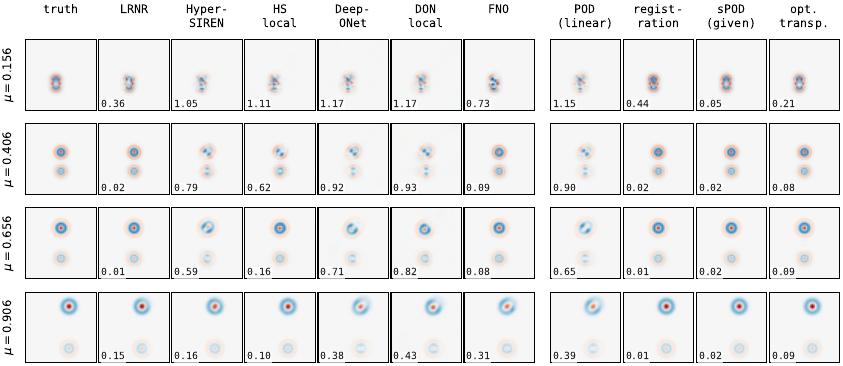}
  \caption{2D rings, $N^\ast = 17$, at four gap midpoints (rows): truth, final LRNR, the five neural baselines other than the CAE and four
  classical interpolants (seed 0 for the trained models); color only (signed fields), numbers are relative errors.  At
  $\mu = 0.156$ the copies have just separated and every model is least accurate.}
  \label{fig:rings}
\end{figure}

\begin{figure}[!htbp]
  \centering
  \includegraphics[width=\linewidth]{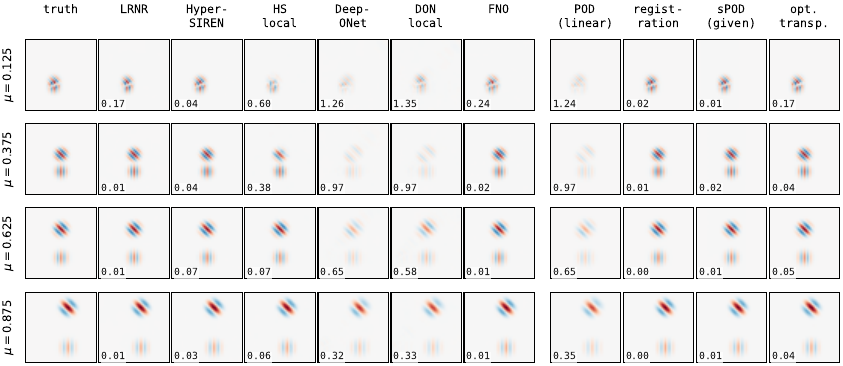}
  \caption{2D wave packet, $N^\ast = 21$, at four gap midpoints: truth, final LRNR, the five neural baselines other than the CAE and four
  classical interpolants (seed 0 for the trained models); numbers are relative errors.}
  \label{fig:wave}
\end{figure}

At the 1D and rings $N^\ast$ all 20 Hyper-SIREN, DeepONet and FNO runs fail, with held-out errors of 0.12--0.58 against the final LRNR's
0.062--0.083 (Table~\ref{tab:oscbaselines}).  The envelope test alone does not separate them: their envelope ratios
are 0.87--2.1, the one-sided behavior of \S\ref{sec:oscrule}; their gaps of 3--161 do, on all runs but one, which
fails on its envelope ratio (0.87).  Between the snapshots their packets lose amplitude or break into fragments
(Figures~\ref{fig:oscstrip1d} and~\ref{fig:rings}).  On the wave packet at $N=21$ Hyper-SIREN passes (both seeds whole,
seed 1 local) with 1.4--3.4 times the final LRNR's error (0.036 / 0.116 whole), and DeepONet fails, with held-out
errors of 0.62--0.65 and gaps of 3--15 (Figure~\ref{fig:wave}).  The FNO passes too (by the absolute gap) and is about
as accurate as the final LRNR (0.043 / 0.023 against 0.026 / 0.034): the family is a fixed-frequency carrier under a
moving envelope, and we posit that the Fourier layers represent the carrier directly.  Like the LRNR it fails at $N\le17$, with
held-out errors of 0.54--1.18; at $N = 29$ it reaches 0.014 / 0.013, against the final LRNR's 0.022 / 0.021 there.
The CAE, with the training recipe that fits every non-oscillatory family, does not leave its initial plateau on the
1D family or the wave packet: its output stays at zero.  On the rings it fits (0.065 / 0.005) and fails between
the snapshots (0.60 / 0.53).

\subsection{Errors between the snapshots and above $N^\ast$}\label{sec:oscbetween}

\begin{figure}[!htbp]
  \centering
  \includegraphics[width=\linewidth]{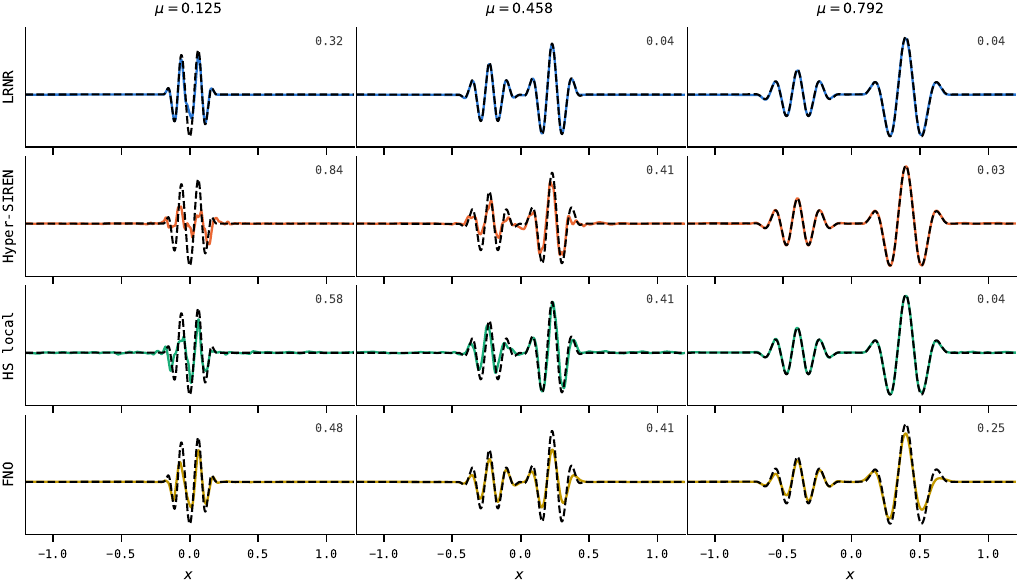}
  \caption{1D oscillatory split, $N=13$ (first-stage models), at three gap midpoints.  The family's $N^\ast$ is 9;
  Figure~\ref{fig:oscstrip1d} shows $N^\ast$.}
  \label{fig:strip_osc1d}
\end{figure}

\begin{figure}[!htbp]
  \centering
  \includegraphics[width=\linewidth]{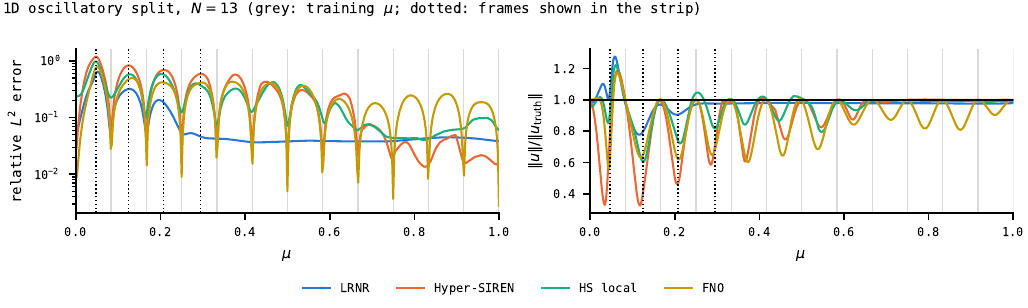}\\[0.3em]
  \includegraphics[width=\linewidth]{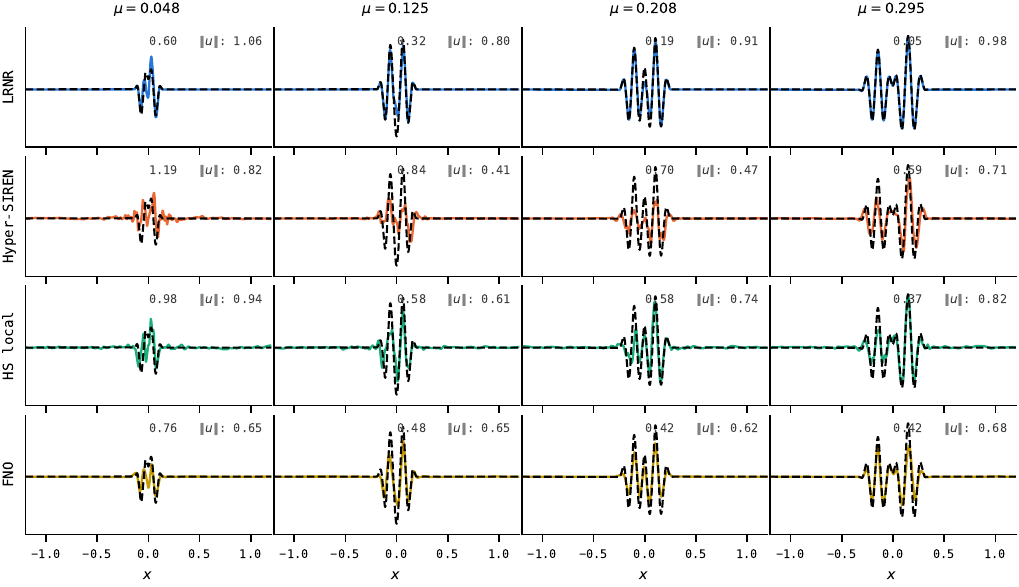}
  \caption{1D oscillatory split, $N = 13$.  Top: relative error and amplitude $\|u\|/\|u_{\rm truth}\|$ of each model
  on 401 values of $\mu$ (gray: training snapshots).  Bottom: the frames at the dotted $\mu$, chosen by rule (in each
  gap, the $\mu$ of Hyper-SIREN's largest error; the four gaps where it is largest); each panel gives the error and the
  amplitude.  Hyper-SIREN fits the snapshots but its packet loses amplitude and chirps in every gap up to
  $\mu\approx0.65$; the LRNR's error is flat in $\mu$ except in the first three gaps.  All frames: Figures~\ref{fig:surf_os1}
  and~\ref{fig:errfall_os1}.}
  \label{fig:between}
\end{figure}

At $N=13$, above its $N^\ast$, the 1D family was scanned and its frames chosen by the rule of \S\ref{sec:baselines}
(Figure~\ref{fig:between}; first-stage LRNR, Hyper-SIREN whole and local, FNO).  Over all $\mu$ the first-stage LRNR's
held-out error is 0.090 / 0.152 against Hyper-SIREN's 0.285 / 0.262 and local Hyper-SIREN's 0.241 (seed 1 diverged).
Hyper-SIREN's error averages 0.37 in the middle half of the gaps, the FNO's 0.32, against the LRNR's 0.11.  Mid-gap, Hyper-SIREN's wave packet
shrinks to as little as a third of the truth's norm (0.33 at $\mu \approx 0.12$), and its carrier
chirps as the packet re-forms toward the next snapshot.  The LRNR's own worst frames are in the first gaps, where the
carrier is finest ($\mu < 0.15$, error up to 0.65); the rule includes the gap of the LRNR's largest error.  Figure~\ref{fig:strip_osc1d} shows three
gap midpoints, and Appendix~\ref{app:movies} all frames.

\subsection{Classical methods}\label{sec:oscmeasure}

\begin{table}[!htbp]
\centering\small
\caption{Oscillatory families: mean / worst relative held-out error over $\mu\ge0.1$ at $N^\ast$ (LRNR: final models,
seed 0 and seed 1).  Underlined: smallest mean and smallest worst error per family.  $^\dagger$Given the exact shift of each
copy at every $\mu$, held-out ones included; an oracle, not a like-for-like competitor (\S\ref{sec:classical}).}
\label{tab:minimaxosc}
\begin{tabular}{lccc}
\toprule
 & 1D oscillatory & 2D rings & 2D wave packet \\
 & $N^\ast=9$ & $N^\ast=17$ & $N^\ast=21$ \\
\midrule
LRNR, s0          & \underline{.062} / \underline{.416} & .070 / .581 & .026 / .189 \\
LRNR, s1          & .083 / .479 & .081 / .661 & .034 / .408 \\
POD, linear       & .524 / 1.238 & .536 / 1.152 & .623 / 1.242 \\
POD-I (cubic)     & .495 / 1.378 & .543 / 1.276 & .560 / 1.414 \\
registration      & .575 / 1.801 & .060 / .916 & \underline{.018} / \underline{.038} \\
sPOD, fitted      & .836 / 1.365 & .703 / 1.397 & .804 / 1.602 \\
sPOD, provided$^\dagger$ & .323 / .585 & \underline{.023} / \underline{.051} & .022 / .039 \\
OT                & .250 / .454 & .111 / .455 & .056 / .170 \\
DMD               & .991 / 1.004 & .999 / 1.007 & .984 / 1.133 \\
\bottomrule
\end{tabular}
\end{table}

Table~\ref{tab:minimaxosc} compares the classical interpolants of \S\ref{sec:classical} with the final LRNR, and
Figures~\ref{fig:oscstrip1d}--\ref{fig:wave} show them at four gap midpoints.  The
results depend on the snapshot spacing.  On the 1D family ($N=9$) every classical method fails: the best are OT (mean
0.250) and sPOD with provided shifts (0.323, limited not by its shifts but by the split into frames,
Appendix~\ref{app:classical}), against the LRNR's 0.062 / 0.083; POD cross-fades (0.50--0.52), and
registration does no better (0.58).  On the rings ($N=17$) sPOD with provided shifts (0.023 mean, 0.051 worst) and
registration (0.060 mean) are more accurate on average than the LRNR (0.070 / 0.081), and OT (0.111) is close to it.  sPOD is
given each ring's exact shift at every $\mu$, which the LRNR has to find.  On
the wave packet ($N=21$) registration (0.018 mean, 0.038 worst) and sPOD with provided shifts (0.022, 0.039) are more
accurate than the LRNR (0.026 / 0.034 mean, 0.19 / 0.41 worst); OT is less accurate on average (0.056) but has a
smaller worst error (0.17).  Both families combine the translation of the copies with a smooth dilation of the
carrier: the rings' spacing grows two- to fourfold over the family, and the packet's crests spread apart along the
direction of travel.  A diffeomorphic warp between neighboring snapshots describes this case directly.  With 17--21 snapshots neighboring snapshots differ by a small
motion.  We posit that methods that build the motion in (given shifts, gap-by-gap warps) profit from this more than a
regression model.  The worst errors lie at the edge of the overlap region: the LRNR's 0.42--0.66 on the 1D family and the rings at
$\mu = 0.10$--0.17, where OT's worst error is 0.45, sPOD's with provided shifts 0.05--0.59, registration's 0.92--1.8
and the POD variants' 1.2--1.4.  DMD fits none of the three families (training error 0.95--1.00 over $\mu\ge0.1$, held-out
0.98--1.00): the carrier makes neighboring snapshots further from a linear map of each other than on the
non-oscillatory families.  Across the three families the LRNR's worst error is at most 0.66, against 1.1--1.8
for the classical methods that apply to all of them (POD, POD-I, registration, fitted sPOD, DMD), 0.45 for OT, and 0.59
for sPOD with the shifts provided.

\subsection{Hyper-modes}\label{sec:oschypermodes}

\begin{figure}[!htbp]
  \centering
  \includegraphics[width=\linewidth]{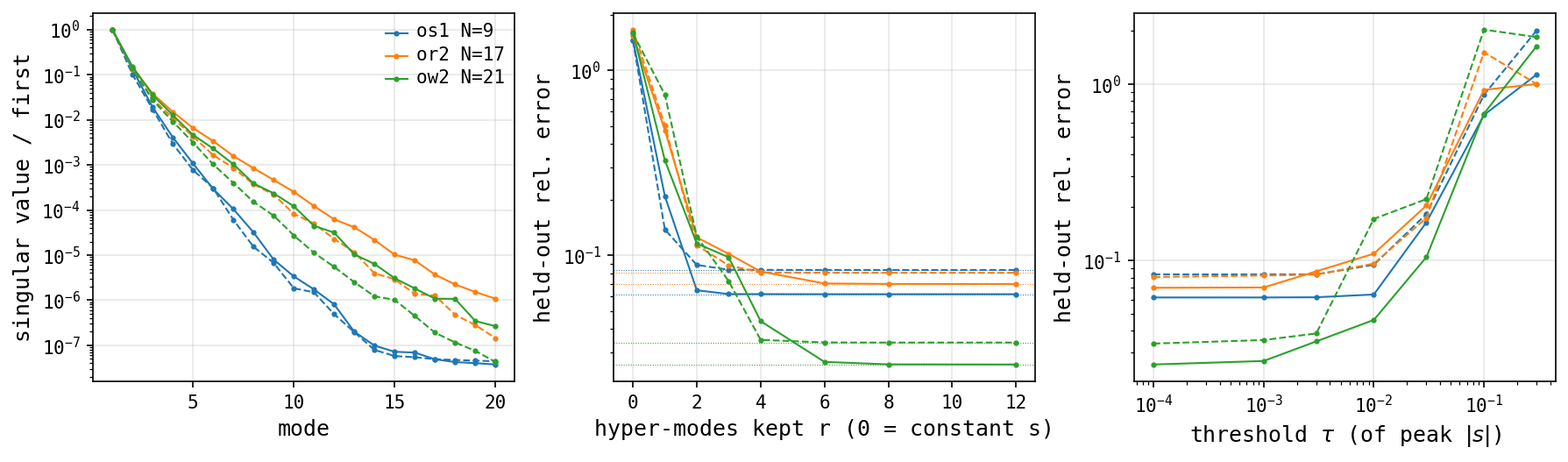}
  \caption{Oscillatory families: hyper-modes of the final LRNRs at $N^\ast$ (panels as in Figure~\ref{fig:hypermodes};
  errors over $\mu\ge0.1$).}
  \label{fig:hypermodesosc}
\end{figure}

The hyper-mode analysis of \S\ref{sec:latent} on the final models at $N^\ast$ (Table~\ref{tab:hypermodesosc},
Figure~\ref{fig:hypermodesosc}; errors over $\mu\ge0.1$) gives the same picture with more modes.  The coefficient path
is again one-dimensional to first order (effective dimension 1.02--1.05), but one hyper-mode leaves errors of 0.14--0.21
on the 1D family and 0.33--0.74 on the 2D families, many times the full models' 0.03--0.08.  The 1D family needs three
hyper-modes, the rings and the wave packet four to six, as many as the spiral and more than the split families;
their singular values decay about as slowly as the spiral's.  Truncation at $\tau = 10^{-3}$ keeps 764--1606 coefficients without loss (5\% more
error on the wave packet); the 2D families keep fewer than half.

\begin{figure}[!htbp]
  \centering
  \includegraphics[width=\linewidth]{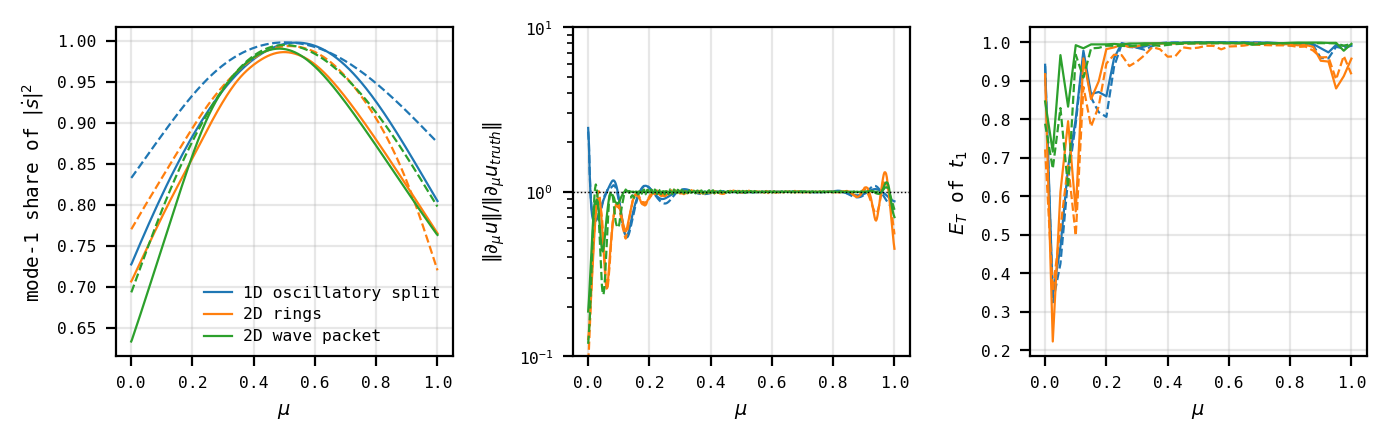}
  \caption{Oscillatory families: hyper-mode tangents of the final LRNRs at $N^\ast$ (panels as in
  Figure~\ref{fig:hypertangent}).}
  \label{fig:hypertangentosc}
\end{figure}

\begin{figure}[!htbp]
  \centering
  \includegraphics[width=\linewidth]{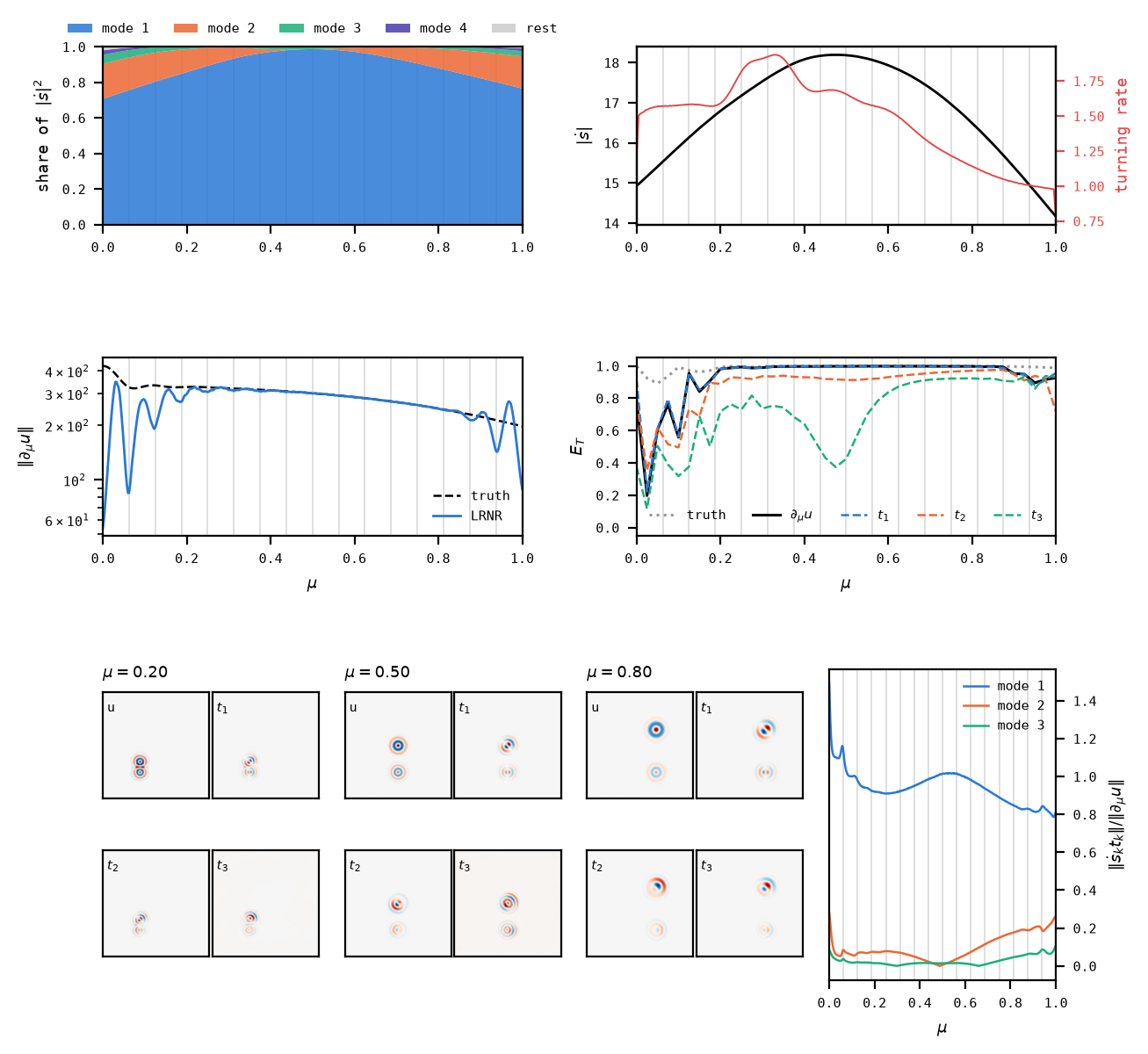}
  \caption{Hyper-mode tangents, 2D rings, $N^\ast=17$, final model, seed 0 (panels as in
  Figure~\ref{fig:hypertangentsp1}).  The field stalls at the early snapshots ($\mu = 1/16, 1/8$) while $|\dot\bs|$
  stays smooth.}
  \label{fig:hypertangentor2}
\end{figure}

The hyper-mode tangents (\S\ref{sec:latent}; Figures~\ref{fig:hypertangentosc} and~\ref{fig:hypertangentor2}) show
the same smooth, bent path: mode 1 accounts for 88--95\% of $|\dot\bs|^2$ on average, the mode-2 share follows
$(\mu-\frac12)^2$ (correlation 0.94--0.99), and $|\dot\bs|$ is the same at the snapshots and between them.  Unlike
on the other families, however, the field does not always follow it.  At the onset of the oscillation the field stalls at the
snapshots: on the rings $\|\partial_\mu u\|$ is 0.26 times the truth's at the snapshot $\mu = 1/16$ and 0.59 times at
$\mu = 1/8$, recovering in between, and on the 1D family 0.54 times at $\mu = 1/8$; from $\mu \approx 0.25$ on it is
within a few percent of the truth's.  On the rings the model's change there is also less transport-like than the
truth's ($E_T$ 0.20--0.33 at $\mu = 0.025$, the truth's $\ge 0.89$); on the 1D family both dip ($E_T$ 0.28--0.34, the
truth's 0.17).  The wave packet, whose fields are oscillatory from $\mu = 0$, shows a milder form of both: its field
speed is 0.24--0.45 times the truth's at the first snapshot ($\mu = 1/20$) and 0.62--0.87 times at the second, and its
$E_T$ dips to 0.62--0.71 before $\mu = 0.1$ (the truth's $\ge 0.99$) and is $\ge 0.91$ after.  The stall occurs at the $\mu$ where
these families have their largest errors.

\begin{table}[!htbp]
\centering\small
\caption{Oscillatory families: hyper-modes of the final LRNRs at $N^\ast$ (seed 0 / seed 1), errors over $\mu\ge0.1$;
columns as in Table~\ref{tab:hypermodes}.}
\label{tab:hypermodesosc}
\setlength{\tabcolsep}{3pt}
\resizebox{\linewidth}{!}{%
\begin{tabular}{lcccccccc}
\toprule
family & eff.\ dim & $r=1$ & $r=2$ & $r=3$ & $r=4$ & $r=6$ & full & $\tau=10^{-3}$: kept, error \\
\midrule
1D oscillatory, $N=9$ & 1.03 / 1.02 & .209 / .138 & .065 / .089 & .062 / .084 & .062 / .084 & .062 / .083 & .062 / .083 & 1606 / 1493, .062 / .083 \\
2D rings, $N=17$      & 1.05 / 1.04 & .474 / .505 & .125 / .113 & .102 / .088 & .082 / .081 & .071 / .081 & .070 / .081 & 800 / 764, .070 / .082 \\
2D wave packet, $N=21$ & 1.05 / 1.04 & .327 / .738 & .116 / .127 & .098 / .073 & .044 / .035 & .027 / .034 & .026 / .034 & 819 / 798, .027 / .035 \\
\bottomrule
\end{tabular}}
\end{table}

\subsection{Summary}

The oscillatory families need more snapshots than the others (9 for 1D, 17 for the rings and 21 for the wave packet, against 5 for
the same splits without a carrier) because the carrier's phase is undersampled in $\mu$ until neighboring snapshots
are less than about half a wavelength apart, and they need a different yardstick, because relative errors and field
fade ratios are dominated by the carrier's phase.  With that yardstick the LRNR passes on all three, the baselines
fail at every $N^\ast$ up to 17, and the classical methods with built-in transport are competitive only at the 2D
families' larger $N^\ast$.  Two open problems remain: the envelope ratio dips below 0.9 again above $N^\ast$ on the 1D
family and the rings, and at the 1D and rings $N^\ast$ the second stage fits the snapshots more closely than it
interpolates between them.

\section{The interpolants as dynamical systems in $\mu$}\label{sec:dynamics}

The analyses of this section use the final models at $N^\ast$ where available and otherwise the first-stage models, in which
the displacement is decided (\S\ref{sec:training}); the second stage changes their fade ratios by at most 0.06.

\subsection{Transport decomposition}\label{sec:transport}

Treating $\mu$ as time, we fit, at points inside every gap, smooth fields $\bv(\bx)$ (Gaussian radial basis
functions, $16\times16$ in 2D, 40 in 1D) to
\begin{equation}
  \partial_\mu u \approx -\bv\cdot\nabla u
\end{equation}
by least squares, with $\partial_\mu u$ from central differences ($h = 2\cdot10^{-3}$) and $\nabla u$ on the grid.  We
report $E_T$, the fraction of $\|\partial_\mu u\|^2$ explained by transport; $E_F$, the fraction explained by the
model's own straight cross-fade velocity $(u(\mu_{g+1})-u(\mu_g))/(\mu_{g+1}-\mu_g)$; and
$e_{du} = \|\partial_\mu u_{\rm model}-\partial_\mu u_{\rm truth}\|/\|\partial_\mu u_{\rm truth}\|$, which checks that
the model changes the way the truth does (comparing fitted velocities instead is ill-posed: the component of
$\bv$ along level sets is not determined by the data).  A growth term $ru$ is not tabulated: a smooth growth
field imitates the translation of a smooth positive hump almost exactly ($r = -cu'/u$), so it explains 90--100\% for
every model and the truth.

\begin{table}[!htbp]
\centering\small
\caption{Transport decomposition (seed 0; LRNR: final model at the splits' $N^\ast$, first stage elsewhere).  $E_T$: fraction of $\partial_\mu u$ explained by transport (mean, worst
gap); $E_F$: by the model's cross-fade; $e_{du}$: error of $\partial_\mu u$ against the truth's.}
\label{tab:transport}
\setlength{\tabcolsep}{4pt}
\begin{tabular}{llcccc}
\toprule
case & model & $E_T$ & $E_T$ worst gap & $E_F$ & $e_{du}$ \\
\midrule
\multirow{4}{*}{1D split, $N=5$} & truth & 0.94 & 0.78 & 0.84 & 0 \\
 & LRNR & 0.96 & 0.87 & 0.83 & \underline{0.07} \\
 & Hyper-SIREN / local & 0.56 / 0.69 & 0.38 / 0.51 & $-3.1$ / $-0.65$ & 1.01 / 0.86 \\
 & DeepONet / local & 0.79 / 0.78 & 0.43 / 0.45 & $-1.05$ / 0.31 & 0.44 / 0.55 \\
\midrule
\multirow{4}{*}{2D split, $N=5$} & truth & 0.99 & 0.99 & 0.46 & 0 \\
 & LRNR & 1.00 & 0.99 & 0.46 & \underline{0.07} \\
 & Hyper-SIREN / local & 0.77 / 0.81 & 0.74 / 0.78 & 0.05 / 0.05 & 0.84 / 0.88 \\
 & DeepONet / local & 0.73 / 0.69 & 0.63 / 0.62 & 0.11 / 0.40 & 0.76 / 0.76 \\
\midrule
\multirow{3}{*}{2D spiral, $N=9$} & truth & 0.99 & 0.99 & 0.60 & 0 \\
 & LRNR & 0.97 & 0.76 & 0.72 & \underline{0.30} \\
 & Hyper-SIREN / local & 0.78 / 0.83 & 0.67 / 0.73 & $-6.9$ / $-3.9$ & 0.91 / 0.81 \\
2D spiral, $N=21$ & LRNR & 0.99 & 0.99 & 0.88 & 0.29 \\
\midrule
\multirow{3}{*}{1D oscillatory, $N=13$} & truth & 0.94 & 0.38 & 0.83 & 0 \\
 & LRNR & 0.95 & 0.52 & 0.82 & \underline{0.12} \\
 & Hyper-SIREN / local & 0.64 / 0.81 & 0.04 / 0.38 & 0.73 / 0.76 & 0.46 / 0.38 \\
\bottomrule
\end{tabular}
\end{table}

\begin{figure}[!htbp]
  \centering
  \includegraphics[width=\linewidth]{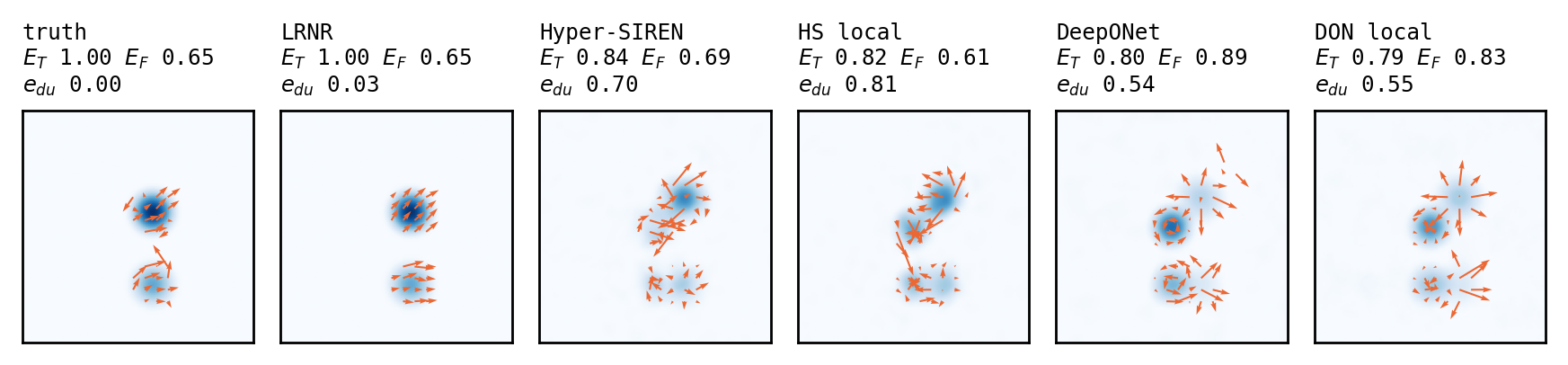}\\[0.4em]
  \includegraphics[width=\linewidth]{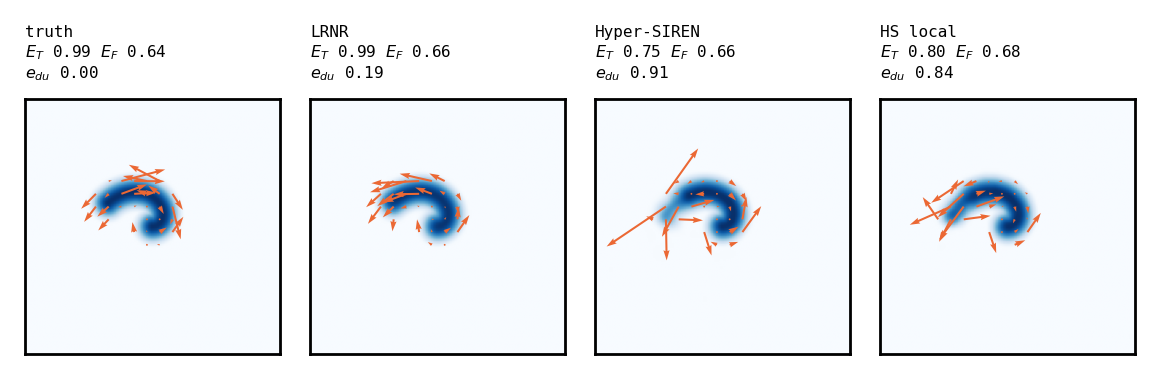}
  \caption{Fitted transport velocity (arrows; the component along level sets is arbitrary) at one unseen $\mu$,
  2D split $N=5$ (top) and spiral $N=9$ (bottom).  The LRNR's arrows follow the truth's; the baselines'
  arrows point radially in and out of fading and growing copies.}
  \label{fig:transport}
\end{figure}

The LRNR's $\mu$-derivative is transport to the same degree as the truth's, and it is the closest to the truth's
derivative (7--30\% against 38--101\%; Table~\ref{tab:transport}).  The baselines' change is explained by transport
only 56--83\%.  We posit that even this fraction comes from a smooth velocity field that imitates a change of
amplitude by spreading or squeezing (the radial arrows in Figure~\ref{fig:transport}).  Hyper-SIREN's strongly negative $E_F$ on the 1D split and the spiral indicates a change
concentrated inside the gap rather than spread across it, consistent with memorizing the snapshots and switching
between them.  At fine spacing the truth's own $E_F$ approaches 0.9, so fade and transport become indistinguishable
to first order (the same effect makes the fade ratio noisy at large $N$), whereas $e_{du}$ remains
discriminating.  In this sense the LRNR interpolant behaves as a non-autonomous transport system
$\partial_\mu u + \bv(\bx,\mu)\cdot\nabla u \approx 0$ with a velocity close to the truth's.

\subsection{The latent path}\label{sec:latent}

\begin{figure}[!htbp]
  \centering
  \includegraphics[width=\linewidth]{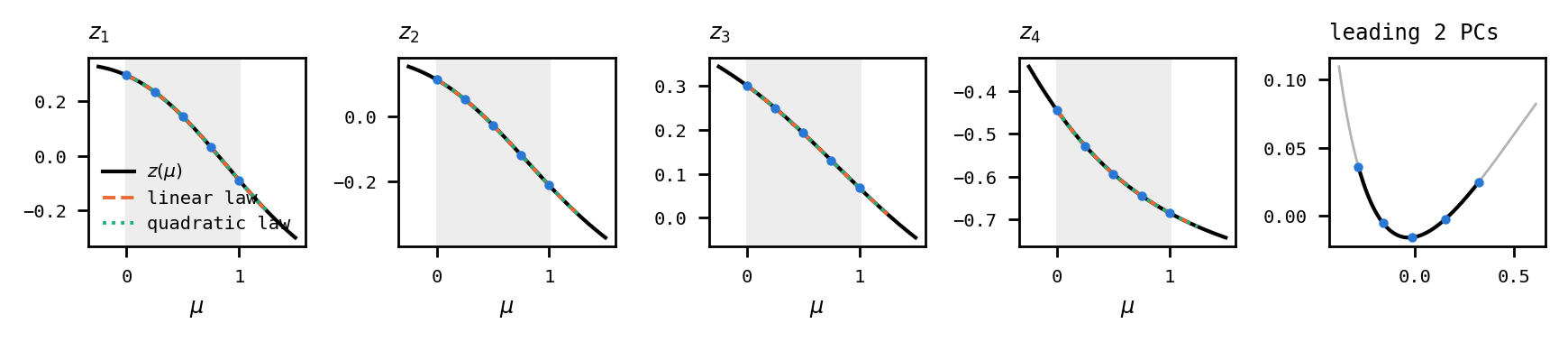}\\[0.3em]
  \includegraphics[width=\linewidth]{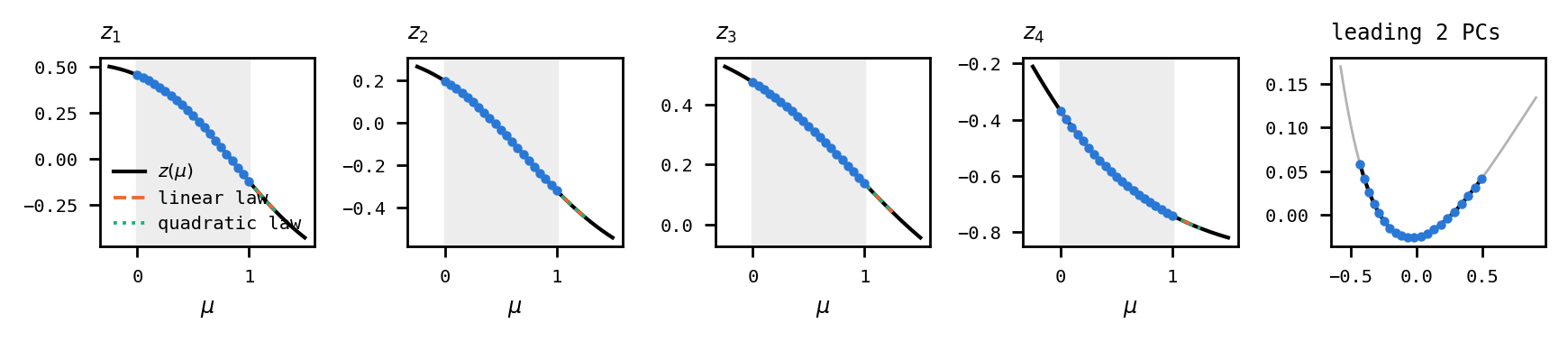}
  \caption{Latent path $\bz(\mu)$ of the LRNR (black; shaded: training range; dots: training snapshots) with the
  trajectories of the fitted autonomous linear and quadratic laws (dashed, dotted; indistinguishable), and the path
  in its two leading principal directions.  2D split $N=5$ (top), spiral $N=21$ (bottom).}
  \label{fig:latent}
\end{figure}

All $\mu$-dependence of the LRNR passes through $\bz(\mu)\in\R^4$~\eqref{eq:hyper}.  For every trained model we examined,
the latent path is almost a straight line traversed at constant speed (Table~\ref{tab:latent},
Figure~\ref{fig:latent}): $\bz$ is affine in $\mu$ to $R^2 \ge 0.979$, the effective dimension of the path is
1.00--1.02 (and 1.00--1.06 for the 2000 coefficients $\bs$), and autonomous laws $\dot\bz = \mathbf{A}\bz+\bb$ or quadratic
$\dot\bz = \mathbf{G}(\bz)$ fitted on $[0,1]$ reproduce it to $10^{-3}$.  The pre-softplus coefficients are affine in $\mu$ as
well, and since $\softplus_{10}$ is close to a ReLU, each coefficient is effectively
$s_k(\mu)\approx\max(0, a_k+b_k\mu)$: 20--40\% are active throughout and change linearly, 47--64\% are inactive
throughout, and 13--17\% switch on or off once (rarely twice) at kinks spread over $\mu\approx0.1$--$0.93$.

\begin{table}[!htbp]
\centering\small
\caption{Latent path of the LRNR (seed 0).  Effective dimension $(\sum\sigma_i^2)^2/\sum\sigma_i^4$ of the centered path
over $[0,1]$; coefficient activity under $\softplus_{10}\approx$ ReLU; model error inside and beyond the
training range.  At the splits' $N^\ast = 5$ the model is the final one; its coefficient path differs from the
first stage's by under 1\% of its peak, and $R^2$ and the activity split are those of the first-stage path.}
\label{tab:latent}
\resizebox{\linewidth}{!}{%
\begin{tabular}{lccccc}
\toprule
case & $R^2$ ($\bz$ affine in $\mu$) & eff.\ dim $\bz$ / $\bs$ & on / off / switching & error $[0,1]$ & error $(1,1.25]$ \\
\midrule
1D split, $N=5$ & 0.997 & 1.00 / 1.00 & 40 / 47 / 13\% & 0.022 & 0.010 \\
2D split, $N=5$ & 0.991 & 1.01 / 1.02 & 29 / 57 / 14\% & 0.028 & 0.054 \\
2D spiral, $N=9$ & 0.990 & 1.02 / 1.06 & 21 / 64 / 15\% & 0.069 & 0.465 \\
2D spiral, $N=21$ & 0.990 & 1.01 / 1.06 & 20 / 63 / 17\% & 0.133 & 0.419 \\
1D oscillatory, $N=13$ & 0.979 & 1.02 / 1.04 & 24 / 61 / 16\% & 0.086 & 0.291 \\
\bottomrule
\end{tabular}}
\end{table}

The straight path has two consequences.  First, in latent coordinates the LRNR is an autonomous system of the simplest kind, a constant
velocity; the apparent non-autonomy of $\partial_\mu u = D_{\bs}\Phi(\bs)\dot\bs$ in function space comes from the decoder,
whose tangent directions change along the line and whose rank-one terms are switched on and off by the gating.  A
straight latent line mapped by a linear decoder gives a cross-fade.  This is DeepONet's situation, since its
output is linear in its spatial modes $\varphi_k$.  The displacement is therefore produced by the LRNR's nonlinear decoder, not by its
latent dynamics.  Second, the straight line extends past $\mu = 1$ unchanged, but the decoder has learned motion only
in the training range: errors beyond it grow to 42--47\% on the spiral and 29\% on the 1D oscillatory family (on the
splits' final models they stay at 1--5\%).

\subsubsection*{Hyper-modes and small coefficients}  
\begin{figure}[!htbp]
  \centering
  \includegraphics[width=\linewidth]{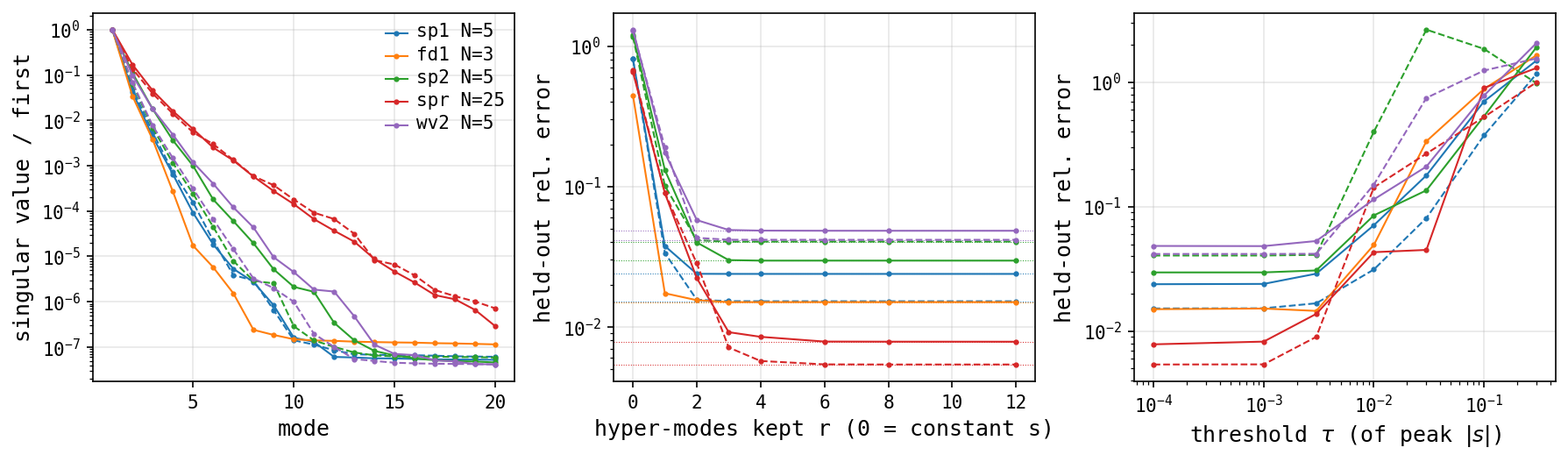}
  \caption{Hyper-modes of the final LRNRs at $N^\ast$ (solid seed 0, dashed seed 1).  Left: singular values of the
  centered coefficient path.  Middle: held-out error with the leading $r$ hyper-modes ($r=0$: constant coefficients;
  dotted: full model).  Right: held-out error after zeroing coefficients below $\tau$ times the peak.}
  \label{fig:hypermodes}
\end{figure}

The same structure shows in the final models at $N^\ast$.
We traced the 2000 coefficients $\bs(\mu)$ on 401 values of $\mu$ and took the SVD of the
centered path, $\bs(\mu) = \bar\bs + \sum_i \sigma_i \alpha_i(\mu) \be_i$; we call the $\be_i$ the \emph{hyper-modes}.  Replacing
$\bs(\mu)$ by its projection $\bar\bs + \sum_{i\le r} \be_i \be_i^\top (\bs(\mu)-\bar\bs)$ on the leading $r$ hyper-modes and
decoding gives the field error of a model with $r$ hyper-modes (Table~\ref{tab:hypermodes},
Figure~\ref{fig:hypermodes}).  The path is one-dimensional to first order: the effective dimension is 1.00--1.06, and
the second singular value is 3--17\% of the first.  One hyper-mode is not enough, however: with $r = 1$ the error is 1.2--17
times the model's.  Two to three hyper-modes recover the model's accuracy on the split families, the fade control and
the 2D wave, four to six on the spiral; the singular values fall to $10^{-6}$ of the first by mode 7--11, on the spiral
only by mode 19.  Constant
coefficients ($r = 0$, a $\mu$-independent model) have errors of 0.45--1.3, so the $\mu$-dependence is essential on every family,
the fade control included.  Truncating small coefficients (zeroing every $s_k$ whose peak over $\mu$ is below
$\tau$ times the largest) has a larger effect on the error: at $\tau = 10^{-3}$ 717--1783 of the 2000 coefficients remain and the
error changes by at most 5\% (the spiral keeps fewer than half), but at $\tau = 10^{-2}$, with 553--1331 kept, it
rises 2- to 26-fold.  Coefficients at 0.3--1\% of the peak still contribute to the displacement.  These are projections of
trained models, without retraining; a model trained with few hyper-modes or few coefficients could be more accurate,
which we did not test.

\subsubsection*{Hyper-mode tangents}  
\begin{figure}[!htbp]
  \centering
  \includegraphics[width=\linewidth]{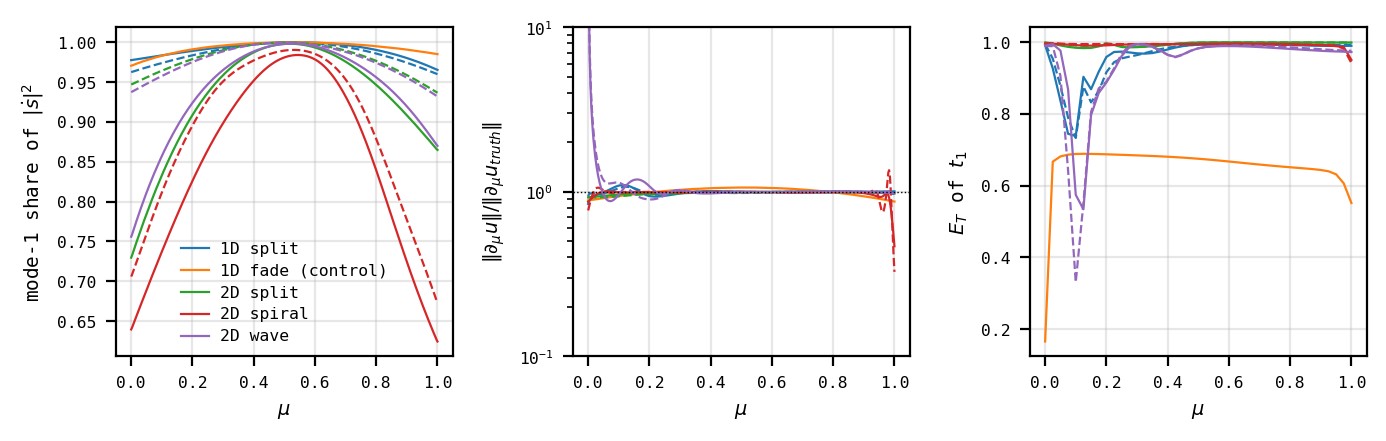}
  \caption{Hyper-mode tangents of the final LRNRs at $N^\ast$ (solid seed 0, dashed seed 1).  Left: share of
  $|\dot\bs|^2$ in hyper-mode 1.  Middle: $\|\partial_\mu u\|$ of the model over the truth's.  Right: transport
  fraction $E_T$ of the mode-1 tangent $t_1$.  Oscillatory families: Figure~\ref{fig:hypertangentosc}.}
  \label{fig:hypertangent}
\end{figure}

\begin{figure}[!htbp]
  \centering
  \includegraphics[width=\linewidth]{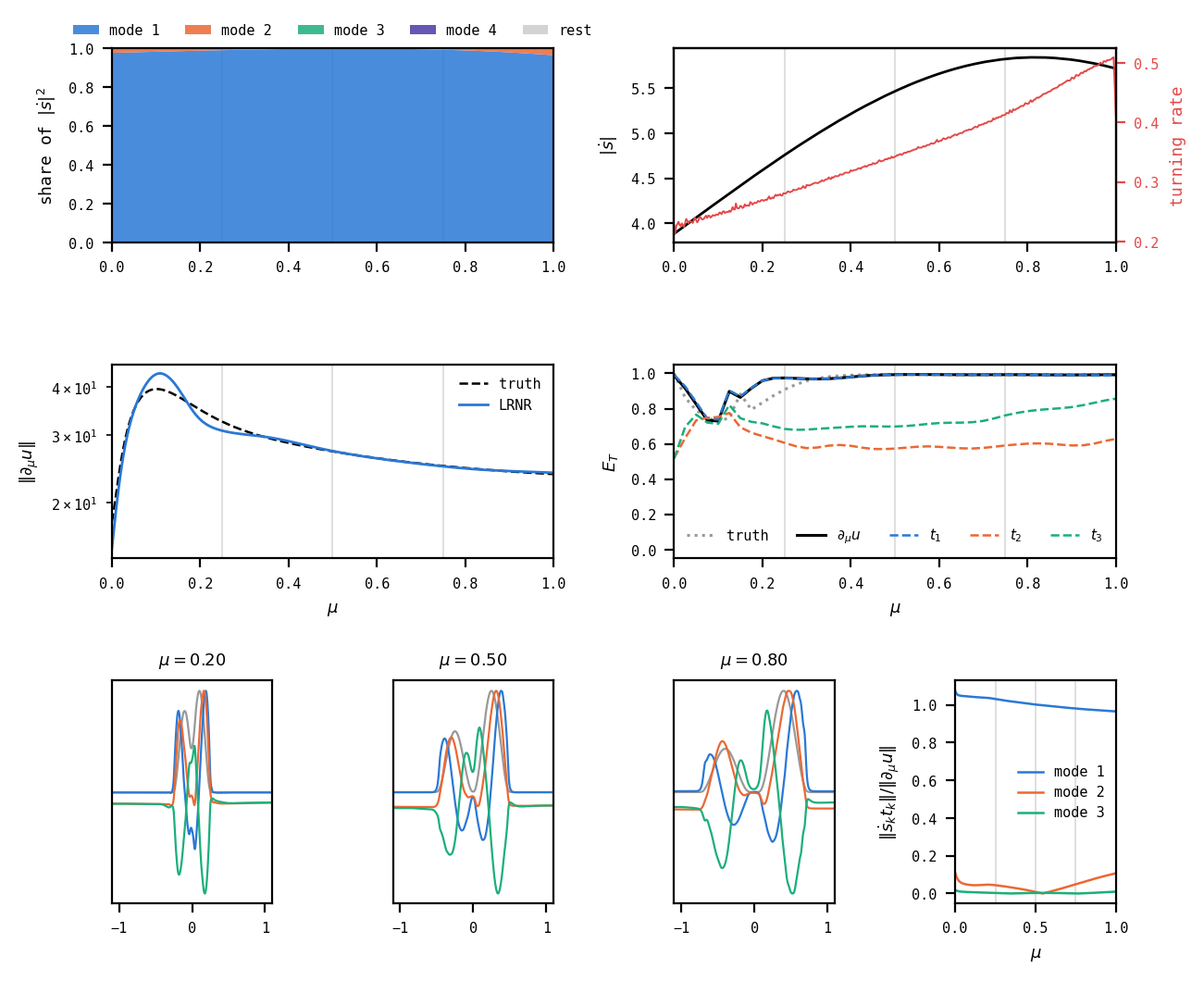}
  \caption{Hyper-mode tangents, 1D split, $N^\ast=5$, final model, seed 0 (vertical lines: training snapshots).  Top:
  shares of $|\dot\bs|^2$ in the hyper-modes; speed $|\dot\bs|$ and turning rate $|d(\dot\bs/|\dot\bs|)/d\mu|$.  Middle:
  $\|\partial_\mu u\|$ against the truth's; $E_T$ of $\partial_\mu u$, of the truth's $\partial_\mu u$ and of the mode
  tangents.  Bottom: $u$ (gray) and $t_1,t_2,t_3$ (normalized) at three $\mu$; $\|(\be_k\cdot\dot\bs)t_k\|/\|\partial_\mu u\|$.}
  \label{fig:hypertangentsp1}
\end{figure}

Along the path we followed the velocity $\dot\bs = d\bs/d\mu$, its shares
$(\be_k\cdot\dot\bs)^2/|\dot\bs|^2$ in the hyper-modes, and in field space the velocity $\partial_\mu u = \bJ\dot\bs$ ($\bJ$: the
decoder's Jacobian in $\bs$) and the \emph{mode tangents} $t_k = \bJ \be_k$, the change of the field when $\bs$ moves along
hyper-mode $k$ (401 values of $\mu$, $E_T$ on 41; Figures~\ref{fig:hypertangent}
and~\ref{fig:hypertangentsp1}).  We make four observations.  (i) The path shows no feature at the training snapshots: $|\dot\bs|$
at the interior snapshots is 1.00--1.04 times its value at the gap midpoints, and $\|\partial_\mu u\|$ is 0.98--1.06
times the truth's at both.  (ii) The path is a gently bent arc.  Hyper-mode 1 accounts for 85--99\% of $|\dot\bs|^2$ on
average; its share peaks ($\ge 0.98$) at $\mu=\frac12$ and falls toward both ends (to 0.62 on the spiral), and the
share of mode 2 follows $(\mu-\frac12)^2$ (correlation 0.91--1.00), as for $\bs \approx \bar\bs + a\bv_1\mu + b\bv_2(\mu-\frac12)^2$.
The shares therefore single out no particular $\mu$; the split, in particular, is not a turn of the path.  (iii) Mode 1
is the transport direction: $E_T$ of $t_1$ is within 0.02 of $E_T$ of $\partial_\mu u$ on every family but the 2D wave,
while $t_2$ and $t_3$ are less transport-like on most families (0.61--1.00) and differ between seeds, whereas the leading
direction is reproducible.  On the fade control every tangent has $E_T \approx 0.6$--0.65, the truth's.
(iv) Where the truth itself is not pure transport the model follows it with a smooth path: at the 1D split
($\mu\approx0.1$) $E_T$ of $\partial_\mu u$ falls to 0.72--0.73, as the truth's (0.72), with $\dot\bs$ smooth and mode 1
still dominant; the non-transport part of the change comes from the decoder's Jacobian along an almost straight path.
The one systematic misfit is the 2D wave at $\mu=0$: the true field starts from rest ($\partial_\mu u = 0$) but the
model's moves at about half its later speed, since $\dot\bs \ne 0$ there; the tangents of modes 1 and 2 partly cancel
(their norms sum to about twice $\|\partial_\mu u\|$) without removing it.

\begin{table}[!htbp]
\centering\small
\caption{Hyper-modes of the final LRNRs at $N^\ast$ (2D wave: $N=5$; seed 0 / seed 1; fade control: one seed): effective dimension of
the coefficient path, held-out error with $r$ hyper-modes and of the full model, and coefficients kept and error after
truncation at $\tau = 10^{-3}$ and $10^{-2}$ of the peak.  Oscillatory families: Table~\ref{tab:hypermodesosc}.}
\label{tab:hypermodes}
\setlength{\tabcolsep}{3pt}
\resizebox{\linewidth}{!}{%
\begin{tabular}{lcccccccc}
\toprule
family & eff.\ dim & $r=0$ & $r=1$ & $r=2$ & $r=3$ & full & $\tau=10^{-3}$: kept, error & $\tau=10^{-2}$: kept, error \\
\midrule
1D split, $N=5$ & 1.00 / 1.01 & .81 / .81 & .038 / .034 & .024 / .016 & .024 / .015 & .024 / .015 & 1718 / 1726, .024 / .015 & 1326 / 1285, .071 / .031 \\
1D fade, $N=3$  & 1.00        & .45        & .017        & .016        & .015        & .015        & 1686, .015                & 1331, .049 \\
2D split, $N=5$ & 1.02 / 1.01 & 1.20 / 1.17 & .131 / .101 & .040 / .041 & .030 / .041 & .030 / .041 & 1561 / 1522, .030 / .041 & 1081 / 1150, .085 / .400 \\
2D spiral, $N=25$ & 1.06 / 1.04 & .65 / .68 & .090 / .090 & .022 / .028 & .009 / .007 & .008 / .005 & 937 / 717, .008 / .005 & 740 / 553, .043 / .143 \\
2D wave, $N=5$  & 1.02 / 1.01 & 1.31 / 1.28 & .174 / .190 & .058 / .043 & .049 / .042 & .048 / .042 & 1716 / 1783, .048 / .042 & 1162 / 1218, .114 / .151 \\
\bottomrule
\end{tabular}}
\end{table}

\subsection{Displacement by biases and by weights}\label{sec:biases}

The inverse-bias trick (\S\ref{sec:invbias}) says a LRNR can move a feature by changing biases alone.  To see
whether trained LRNRs do, we split the coefficients of~\eqref{eq:lrnr} into the weight group $\bs^W(\mu)$ and the bias
group $\bs^b(\mu)$ and probed the models in two ways.

\subsubsection*{Derivative split}  To first order $\partial_\mu u = \bJ_W\dot\bs^W + \bJ_b\dot\bs^b$; each part is a central
difference ($h = 2\cdot10^{-3}$) in which only that group moves (the sum reproduces $\partial_\mu u$ to 0.2\%).  We
report, averaged over $\mu$ at $\tfrac14,\tfrac12,\tfrac34$ of every gap, the bias part's share of the energy
$|\bJ_b\dot\bs^b|^2/(|\bJ_b\dot\bs^b|^2+|\bJ_W\dot\bs^W|^2)$, the transport-only explained fraction $E_T$ of each part
(\S\ref{sec:transport}), the error $e_{du}$ of each part alone against the truth's $\partial_\mu u$, and how the
bias part divides among the five layers.

\subsubsection*{Frozen coefficients}  We evaluated hybrid models with one group held at its value at $\mu=\tfrac12$ and the other
left free, and scored them like every other model (training and held-out error, filtered fade ratio of the hybrid against its own
cross-fade).

\begin{table}[!htbp]
\centering\small
\caption{Biases against weights in trained LRNRs (seed 0 / seed 1).  Left: split of $\partial_\mu u$ into the parts due
to the bias and the weight coefficients: energy share of the bias part, error of each part alone against the truth's
$\partial_\mu u$ ($e_{du}$ of the full model for reference), and the first layer's share of the bias part.  Right:
held-out error / filtered fade ratio with one group frozen at $\mu=\tfrac12$.  Splits and spiral: final models at
$N^\ast$; 1D oscillatory $N=13$: first stage.  Frozen columns: first row seed 0, second row seed 1; 1D oscillatory: one
seed.}
\label{tab:biases}
\resizebox{\linewidth}{!}{%
\begin{tabular}{lcccccccc}
\toprule
 & \multicolumn{5}{c}{derivative split} & \multicolumn{3}{c}{frozen at $\mu=\tfrac12$: held / fade} \\
\cmidrule(lr){2-6}\cmidrule(lr){7-9}
case & bias share & $e_{du}$ full & $e_{du}$ bias part & $e_{du}$ weight part & first-layer share & full & biases frozen & weights frozen \\
\midrule
1D split, $N=5$   & 0.90 / 0.89 & 0.07 / 0.05 & 0.26 / 0.26 & 0.81 / 0.80 & 0.92 / 0.88 & 0.02 / 0.95 & 0.71 / 0.02 & 0.21 / 0.66 \\
                  &             &             &             &             &             & 0.02 / 0.96 & 0.70 / 0.02 & 0.22 / 0.62 \\
2D split, $N=5$   & 0.98 / 0.98 & 0.07 / 0.11 & 0.17 / 0.18 & 0.93 / 0.94 & 0.92 / 0.96 & 0.03 / 0.95 & 1.18 / 0.02 & 0.30 / 0.76 \\
                  &             &             &             &             &             & 0.04 / 0.94 & 1.17 / 0.02 & 0.30 / 0.61 \\
1D oscillatory, $N=13$ & 0.75 & 0.12 & 0.39 & 0.65 & 0.95 & 0.09 / 0.84 & 1.39 / 0.04 & 1.16 / 0.23 \\
2D spiral, $N=25$ & 0.32 / 0.51 & 0.10 / 0.07 & 0.68 / 0.68 & 0.45 / 0.73 & 0.54 / 0.56 & 0.01 / 0.98 & 0.38 / 0.68 & 0.61 / 0.02 \\
                  &             &             &             &             &             & 0.01 / 1.00 & 0.70 / 0.23 & 0.73 / 0.06 \\
\bottomrule
\end{tabular}}
\end{table}

\begin{figure}[!htbp]
  \centering
  \includegraphics[height=0.4\textheight]{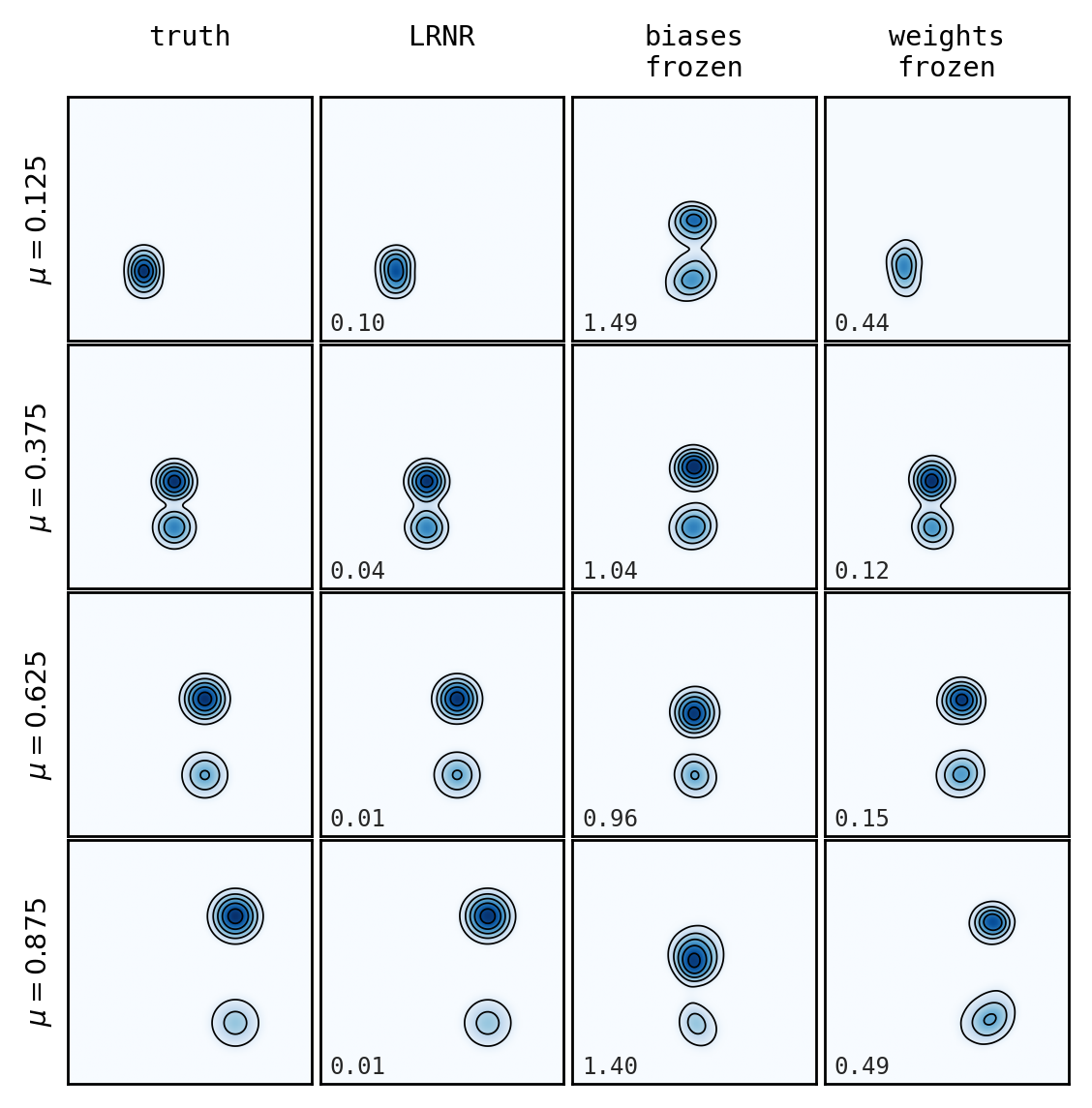}\\[0.3em]
  \includegraphics[height=0.4\textheight]{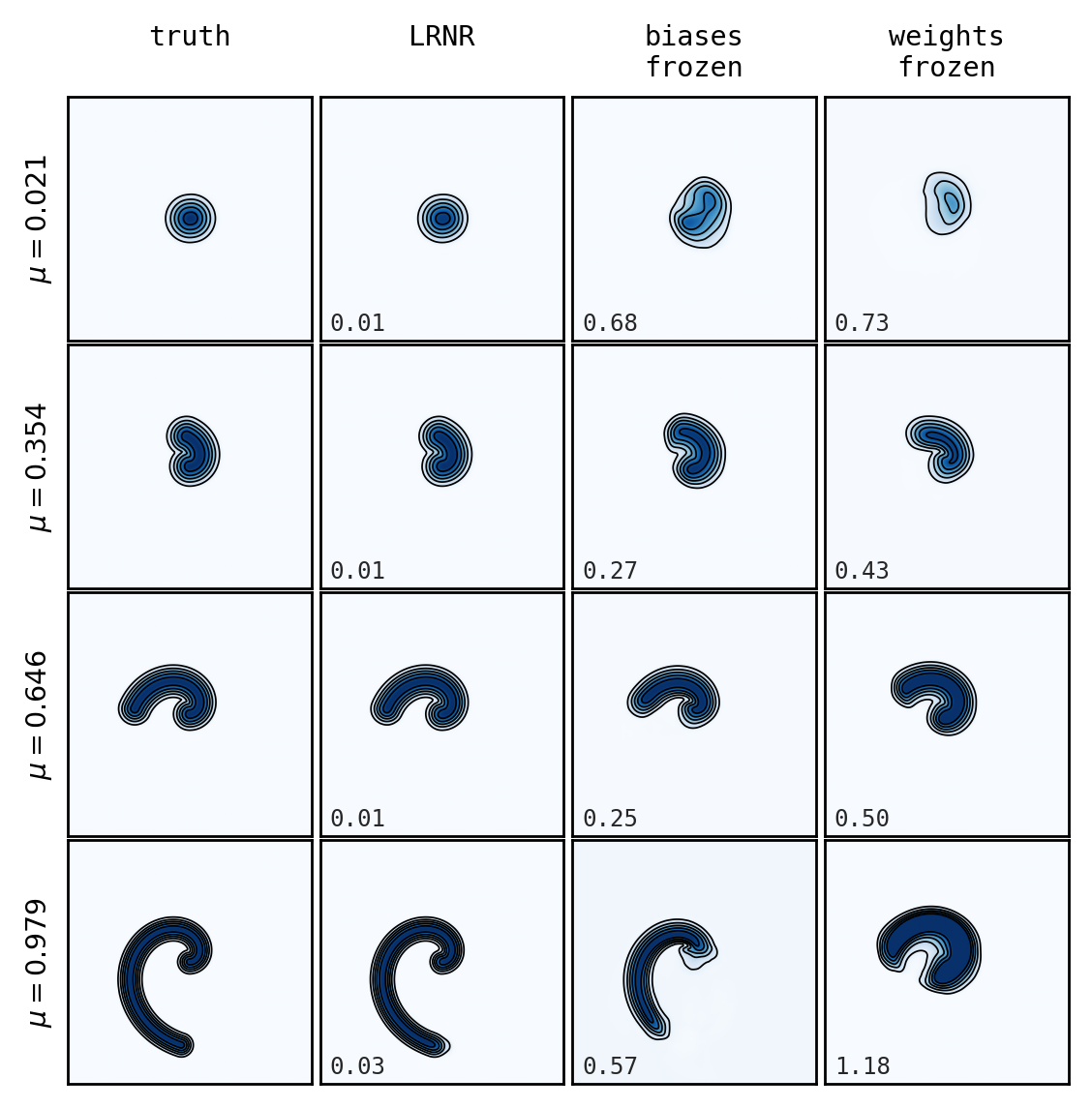}
  \caption{Coefficients frozen at $\mu=\tfrac12$ (final models, seed 0; mid-gap frames in rows, contoured at the
  truth's levels; numbers are relative errors).  2D split
  $N=5$ (top): with the biases frozen the two copies stay where they are at $\mu=\tfrac12$ and only change in height,
  a cross-fade; with the weights frozen they still travel to the right places.  Spiral $N=25$ (bottom): the roles are
  reversed in part: with the weights frozen the arc stops lengthening and winding, with the biases frozen it still
  grows but is misplaced near the ends of the range.}
  \label{fig:biases}
\end{figure}

The results (Table~\ref{tab:biases}, Figure~\ref{fig:biases}) separate the families by the kind of motion they contain.

\subsubsection*{Split families: first-layer biases}  On the 1D and 2D splits, whose features translate,
89--98\% of the energy of $\partial_\mu u$ comes from the bias coefficients, and 88--96\% of that from the first layer,
the one acting directly on the input features and the layer where~\eqref{eq:invbias} places the transport.  The bias part
alone reproduces the truth's $\partial_\mu u$ nearly as well as the whole model ($e_{du}$ 0.17--0.18 against 0.07--0.11 in
2D), and it is pure transport ($E_T = 0.96$--$1.00$); the weight part alone does not ($e_{du}$ 0.80--0.94).  Freezing the
biases turns the model into a cross-fade (fade ratio 0.02 on both families and seeds): the copies stay at their
mid-range positions and only change in height.  Freezing the weights instead leaves the copies moving (fade ratio
0.61--0.76) to roughly the right places; the error grows mainly through their heights and widths.  The two parts are
positively correlated (cosine 0.38--0.78), which suggests that the weights refine the bias-driven motion rather than
oppose it.

\subsubsection*{1D oscillatory split: carrier motion and frequency change}  The carrier motion is in the biases and the
frequency change in the weights.  The bias part still
dominates (75\%, 95\% of it in the first layer), and freezing the biases again gives a cross-fade (0.04).  But freezing
the weights also fails (fade ratio 0.23, error 116\%): the carriers' frequencies change with $\mu$, a change of scale
that a shift of biases cannot produce.

\subsubsection*{Spiral: deformation and the weights}  The spiral's arc lengthens and winds up.  This deformation differs from a
translation, and here the weights account for half or more of $\partial_\mu u$ (bias share 0.32--0.51, only about half of it in the
first layer).  Freezing the weights stops the motion (fade ratio 0.02--0.06), while freezing the biases leaves part of it
(0.23--0.68).

Freezing at one $\mu$ moves the model off its training path, and the derivative split is local.  Within these limits
of the probe, the results agree with the inverse-bias trick.  Translation, the transported-subspace case, is
realized by the trained LRNRs as a first-layer bias shift, as in~\eqref{eq:invbias}, while changes of scale and shape
use the weights, which the classical construction does not have.

\section{Discussion}\label{sec:discussion}

\subsubsection*{Source of the displacement}  Three results indicate that displacement is a property of the
architecture: the explicit regularizers can be
removed without losing the behavior (\S\ref{sec:ablation}); Hyper-SIREN and DeepONet with the same parameter budget
and the same spatial sampling, and an FNO and a convolutional autoencoder of the same size, do not have it at the families' small $N^\ast$
(\S\ref{sec:baselines}); and the LRNR's latent path is a straight line, so
the motion is produced by the decoder (\S\ref{sec:latent}).  We posit that the decoder's low-rank weights with fixed bases and gated
coefficients let a uniform change of $\mu$ move a feature rather than blend two copies of it.  The
inverse-bias trick (\S\ref{sec:invbias}) gives a possible mechanism: in a LRNR a displacement of fronts is a change of
biases that is affine in the parameter, so a straight latent path suffices to move a feature, whereas in a linear
decoder the same straight path can only cross-fade.  The trained models use this mechanism at the level each family
needs: on the split families 89--98\% of the energy of $\partial_\mu u$ comes from the bias coefficients, most of it (88--96\%)
from the first layer (planar fronts, single-layer displacement), and freezing the biases reduces the LRNR to a cross-fade (\S\ref{sec:biases});
changes of scale and shape (the oscillatory carrier's frequency, the spiral's winding) use the weight coefficients as
well, and on the spiral about half of the bias part lies in the deeper layers, whose fronts are curved.

\subsubsection*{Learned structure and the theory}  For scalar conservation laws the theory
gives LRNRs of small rank whose coefficients are linear or constant in time and are switched by the
activation~\cite[Thm.~4.6]{rim2026entropy}, with the transport represented by the biases~\cite[\S4]{rim2026entropy}; such
representations of hyperbolic wave dynamics are learnable from data~\cite{cho2025hyperbolic}.  The trained
interpolants have this form: the latent path is affine in $\mu$ and each coefficient is effectively
$\max(0, a_k + b_k\mu)$ (\S\ref{sec:latent}).  Of the $2rL = 2000$ coefficients (far more than the
rank-3 construction of the theory needs, and a small fraction of the 2.4M basis parameters), 940--1280 are inactive
for every $\mu$, 400--800 are active and change linearly, and 260--340 switch on or off once (rarely twice), so
720--1060 are nonzero anywhere in $[0,1]$; the path they trace has effective dimension 1.00--1.06
(Table~\ref{tab:latent}).  The theory gives an existence result: it shows that a small LRNR with such coefficients
exists, and it does not show that gradient descent on snapshot data finds one.  In the experiments, training finds a
representation of the predicted form, though with far more active coefficients than the explicit construction.  The translation is represented by the first layer's bias coefficients
(\S\ref{sec:biases}).  These properties were not imposed, since the LRNR is a generic architecture trained on snapshots.  The
families also go beyond the theory's setting: copies that split off, a spiral whose speed and winding change, a carrier
whose frequency changes.  We posit that the LRNR exploits the low-dimensional structure that the theory establishes
for hyperbolic problems and, being general, finds structure of the same kind in these more complicated families.
Where that structure involves more than translation (scale, shape, winding), the weight coefficients also contribute.

\subsubsection*{Classical interpolants}  On families made of a few translating, non-negative features, classical
methods can be as accurate as the LRNR or more accurate.  At the LRNR's own $N^\ast$, registration and sPOD with provided shifts (given the arc's motion)
reach held-out errors of 0.006--0.008 on the spiral ($N = 25$) against the final LRNR's 0.005--0.008.  However, they
need information that the LRNR does not need.  sPOD is very accurate when
the user provides the number of frames and their exact shifts at every $\mu$, and fails when it has to estimate
them; registration and optimal transport interpolate gap by gap from two snapshots at a time, cannot use the rest of
the family, produce folds and tears at changes of topology, and produce kinks in $\mu$ at every snapshot (\S\ref{sec:classical}).
The LRNR learns the displacement from the snapshots, as one smooth model of the whole family; in this sense it extends
the classical transported-subspace idea~\cite{rim2023transported,rim2026entropy} from prescribed to learned transports.
Each classical method also suits one kind of motion (shifts for sPOD, a diffeomorphism for registration, mass
conservation for optimal transport), whereas the same LRNR, unchanged, transports translating, splitting and
spiraling features; on a linear wave and on shocks the fade ratio does not decide this, and we compare held-out
errors there (Appendix~\ref{app:nofade}).  The difference shows in the
worst case (Table~\ref{tab:minimax}): every classical method either has a family on which its worst error is 0.76--1.3,
because its assumption fails there, or, like sPOD with provided shifts, cannot be applied without information the LRNR
does not need, while the LRNR's stays at or below 0.12 on all of them.  The LRNR is also the most accurate on average,
or level with the most accurate, on each of the three families with moving fronts, and it can be used on all four families without knowing in advance what kind of change the family undergoes.  For hyperbolic problems, shocks included, the theory guarantees
that a small LRNR with simple coefficients exists~\cite{rim2026entropy}; the experiments indicate that training finds
such representations, and that the same mechanism applies to families the theory does not cover.

\subsubsection*{The oscillatory families}  A carrier that travels with its copy changes the problem in three ways.
The snapshots undersample the carrier's phase until neighboring snapshots are less than about half a
wavelength apart, a small misplacement of the crests is a large relative error, and the field fade ratio tracks the
phase rather than the motion (\S\ref{sec:oscchallenge}).  With the envelope test away from the initial overlap the LRNR
passes on all three families, at $N^\ast = 9$, 17 and 21, and the baselines fail at the first two; the classical
methods with built-in transport are competitive at the larger $N^\ast$ of the 2D families, and are more accurate
there on average.  The numbers of the oscillatory families are not comparable with those of the other families and are
reported only in \S\ref{sec:osc2d}.

\subsubsection*{Number of snapshots}  Transport is learned once neighboring snapshots overlap enough for the motion between
them to be determined: $N=5$ for the split families, where a copy moves about a third of its width per gap; $N = 25$
for the spiral's fast tail; $N = 9$--21 for the oscillatory families, whose carrier moves several wavelengths per gap
at small $N$ (at $N^\ast$ about half a wavelength or less in 2D).  Placing
snapshots where the family moves fastest helps up to a point (tail12), but the last gaps of the spiral needed a
uniformly finer grid.

\subsubsection*{Spatial mini-batching}  Training on cells of a quarter of the domain keeps the fade ratio close to that of
whole-snapshot training (\S\ref{sec:ndomains}), though the held-out error of one seed rises from 0.079 to 0.132; very small cells reduce the accuracy at a fixed budget.

\subsubsection*{Limitations}  The families are synthetic and, except the 2D Burgers family, known in closed form; two seeds per setting (one for the
fade control); GPU sharing makes
wall-clock comparisons unreliable; the fade ratio is a worst-cell statistic that becomes noisy at fine spacing, so
the transport decomposition and $e_{du}$ are reported alongside it; for oscillatory carriers it is replaced by
an envelope version that is one-sided (a high ratio does not prove displacement) and is evaluated only where the copies
no longer overlap ($\mu\ge0.1$); and extrapolation in $\mu$ has large errors.

\subsubsection*{Follow-ups}  Training a LRNR with $\mu$-dependent first-layer biases only (all other coefficients
constant) would test whether the bias mechanism of \S\ref{sec:biases} suffices for translating families.  An affine hypernetwork ($\bz = \bz_0 + \mu\bw$ by construction) would test whether the
hypernetwork's nonlinearity is needed at all.  On the 2D oscillatory families the fit worsens as $N$ grows at a fixed
budget per snapshot (rings: 0.06--0.10 at $N=25$, 0.24--0.36 at $N=33$), which a larger budget at large $N$ would test.  On the 2D wave and
2D Burgers families, testing whether the LRNR displaces needs a motion test that does not depend on changes of
amplitude or on how sharply the shock is resolved; this is an open problem (Appendix~\ref{app:nofade}).

\section*{Acknowledgments}
The authors acknowledge the Research Infrastructure Services (RIS) group at Washington University in St.~Louis for
providing the computational resources and services needed to generate the research results presented in this paper.
Claude (Anthropic) was used to assist in conducting the experiments and preparing the manuscript; all results were
supervised by the authors, who take responsibility for them.

\appendix
\section{Additional tables}\label{app:tables}

\begin{table}[h]
\centering\small
\caption{2D split, $N=5$: models saved along a 166,667-step run.}
\label{tab:splitckpt}
\begin{tabular}{lccccc}
\toprule
seed & step & train & held-out & gap & fade (filtered) \\
\midrule
0 & 40k & 0.151 & 0.160 & 1.06 & 0.91 \\
0 & 80k & 0.141 & 0.148 & 1.05 & 0.94 \\
0 & 120k & 0.122 & 0.130 & 1.07 & 0.98 \\
0 & 160k & 0.048 & 0.064 & 1.33 & 0.95 \\
0 & final & 0.152 & 0.132 & 0.87 & 0.91 \\
1 & 40k & 0.288 & 0.282 & 0.98 & 0.79 \\
1 & 80k & 0.274 & 0.272 & 0.99 & 0.92 \\
1 & 120k & 0.086 & 0.098 & 1.14 & 0.94 \\
1 & 160k & 0.205 & 0.210 & 1.03 & 0.95 \\
1 & final & 0.143 & 0.138 & 0.96 & 0.91 \\
\bottomrule
\end{tabular}
\end{table}

\begin{table}[h]
\centering\small
\caption{Point placement at $N=5$ (seed 0 / seed 1): uniform grid, gradient-concentrated points (one set), adaptive-mesh-refinement-like (AMR-like)
points (per snapshot).}
\label{tab:sampling}
\begin{tabular}{llccc}
\toprule
family & sampling & train & held-out & fade (filtered) \\
\midrule
1D oscillatory & uniform & 0.078 / 0.059 & 0.413 / 0.385 & 0.49 / 0.60 \\
 & gradient & 0.105 / 0.065 & 0.288 / 0.655 & 0.63 / 0.17 \\
 & AMR-like & 0.123 / 0.082 & 0.292 / 0.372 & 0.67 / 0.60 \\
2D spiral & uniform & 0.058 / 0.134 & 0.176 / 0.215 & 0.31 / 0.34 \\
 & gradient & 0.072 / 0.106 & 0.181 / 0.198 & 0.27 / 0.30 \\
 & AMR-like & 0.057 / 0.074 & 0.185 / 0.189 & 0.23 / 0.29 \\
\bottomrule
\end{tabular}
\end{table}

\begin{table}[h]
\centering\small
\caption{1D split, $N=5$: peak deficit of the taller copy (seed 0 / seed 1).  cos: cosine learning-rate schedule;
cheb: Chebyshev input features.}
\label{tab:peak}
\begin{tabular}{lcccc}
\toprule
variant & peak short & train & held-out & fade (filtered) \\
\midrule
baseline & 7.8 / 5.7\% & 0.083 / 0.074 & 0.089 / 0.077 & 0.96 / 0.96 \\
cos & 6.2 / 5.6\% & 0.059 / 0.058 & 0.065 / 0.062 & 0.97 / 0.94 \\
cheb & 6.3 / 6.5\% & 0.062 / 0.062 & 0.077 / 0.073 & 0.89 / 0.96 \\
cos + cheb & 5.7 / 5.7\% & 0.057 / 0.058 & 0.068 / 0.067 & 0.90 / 0.94 \\
\bottomrule
\end{tabular}
\end{table}

\begin{figure}[h]
  \centering
  \includegraphics[width=\linewidth]{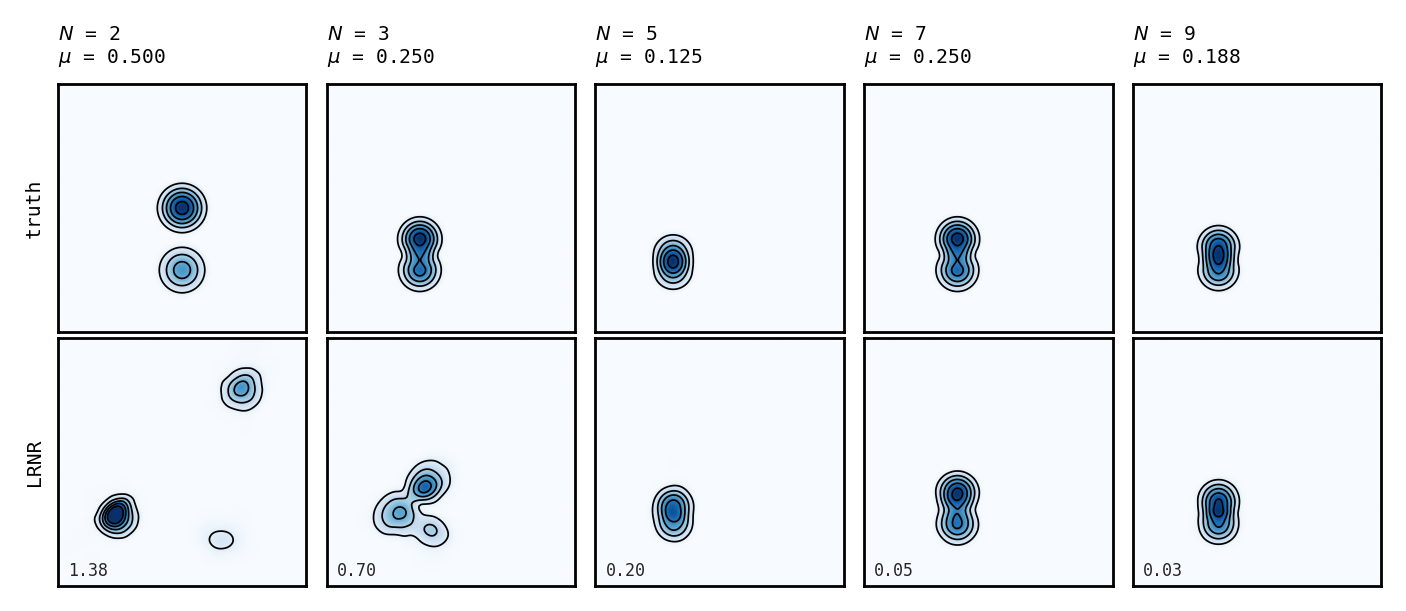}
  \caption{2D split: truth and first-stage LRNR at the unseen gap midpoint nearest $\mu = 0.25$, for each $N$ (seed
  0), contoured at the truth's levels; numbers are relative errors.}
  \label{fig:panels2d}
\end{figure}

\section{Other attempts}\label{sec:other}

\subsubsection*{Point placement}  Datasets whose points concentrate where the snapshots are steep (one point set for all
snapshots, or an AMR-like set per snapshot) did not raise the fade ratio of the spiral or the 1D oscillatory
family to 0.9 at $N=5$ (at most 0.67, Table~\ref{tab:sampling}).

\subsubsection*{The 1D split's peak deficit}  On the 1D split the LRNR moves both copies but leaves the taller one
about 6--8\% short of its peak.  A cosine learning-rate schedule improves the fit by 20--30\% (held-out 0.062--0.065 against
0.077--0.089) without changing the fade ratio; the peak remains about 6\% short (Table~\ref{tab:peak}).

\section{The 2D wave and 2D Burgers families}\label{app:nofade}

Two families are reported here rather than in \S\ref{sec:transition}--\S\ref{sec:classical}: the 2D wave and the
2D Burgers family.  On both, the fade ratio does not separate the motion of a front from other changes of the field.
We compare the LRNR with Hyper-SIREN, DeepONet and the FNO by held-out error instead, and the families are not counted
in the $N^\ast$ comparisons of the main text.

\subsection{The 2D wave}\label{app:wave}
The family is described in \S\ref{sec:families} (Figure~\ref{fig:familiesnofade}).

\begin{figure}[!htbp]
  \centering
  \includegraphics[width=0.8\linewidth]{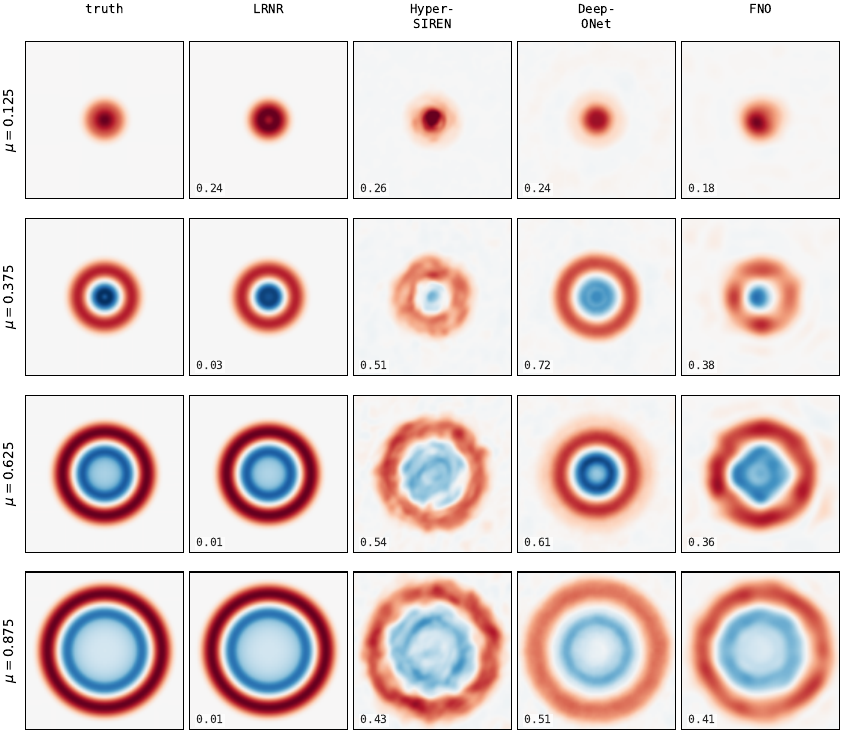}
  \caption{2D wave, $N = 5$, at the four gap midpoints: truth, final LRNR, Hyper-SIREN, DeepONet and the FNO (seed 0);
  numbers are relative errors.}
  \label{fig:wavecmp5}
\end{figure}

Between two snapshots the ring moves outward, loses amplitude and leaves a growing wake, and the field starts from
rest at $\mu = 0$.  We posit that the fade ratio, which compares an interpolant with the cross-fade of its two
neighboring snapshots, mixes the displacement of the front with these changes of amplitude and shape, and we do not
use it on this family.  Table~\ref{tab:wavecmp} compares the
models at $N = 5$ and 7 (Figure~\ref{fig:wavecmp5}: $N = 5$), with the LRNR trained in both stages (\S\ref{sec:l1}).  At both $N$ the final LRNR has the
lowest mean held-out error (0.048 / 0.042 at $N = 5$, 0.044 / 0.046 at $N = 7$).  At $N = 5$ the baselines'
mean errors are five to eight times larger (FNO 0.23 / 0.24, Hyper-SIREN and DeepONet 0.33--0.35), and the LRNR also
has the lowest worst-case error (0.24 / 0.18 against 0.41--0.73).  At $N = 7$ the FNO comes closer: its mean error
is 1.7--1.8 times the LRNR's and its worst-case error is similar (0.18 / 0.21 against 0.19 / 0.19), while the mean errors of
Hyper-SIREN and DeepONet are still four to five times the LRNR's.  In the frames at $\mu \ge 0.375$ the LRNR
reproduces the ring and its wake (errors 0.00--0.03, seed 0).  At $N = 5$ the baselines blur them or superpose them
with the neighboring snapshots (0.36--0.72); at $N = 7$ (gap midpoints, not shown) the FNO is close to the truth (0.06--0.13)
and Hyper-SIREN and DeepONet are not (0.26--0.35).  At $\mu \le 0.25$, where the field starts from rest and the wake forms, all
models have errors of 0.12--0.33; the FNO is the most accurate in two of these three frames (one at $N = 5$, two at $N = 7$) and the LRNR in the
third.

\begin{table}[!htbp]
\centering\small
\caption{2D wave: training error and mean and worst held-out relative error (seed 0 / seed 1).  LRNR: first stage
and final model.}
\label{tab:wavecmp}
\begin{tabular}{llccc}
\toprule
$N$ & model & train & held-out (mean) & held-out (max) \\
\midrule
5 & LRNR, stage 1 & 0.132 / 0.080 & 0.138 / 0.092 & \underline{0.209 / 0.150} \\
  & LRNR, final   & 0.008 / 0.009 & \underline{0.048 / 0.042} & 0.237 / 0.178 \\
  & Hyper-SIREN   & 0.041 / 0.011 & 0.334 / 0.344 & 0.546 / 0.599 \\
  & DeepONet      & 0.014 / 0.026 & 0.352 / 0.327 & 0.730 / 0.583 \\
  & FNO           & 0.006 / 0.006 & 0.227 / 0.241 & 0.408 / 0.483 \\
\midrule
7 & LRNR, stage 1 & 0.031 / 0.069 & 0.060 / 0.091 & 0.189 / 0.231 \\
  & LRNR, final   & 0.005 / 0.008 & \underline{0.044 / 0.046} & 0.186 / \underline{0.189} \\
  & Hyper-SIREN   & 0.011 / 0.018 & 0.208 / 0.224 & 0.409 / 0.606 \\
  & DeepONet      & 0.020 / 0.018 & 0.191 / 0.182 & 0.381 / 0.360 \\
  & FNO           & 0.005 / 0.008 & 0.073 / 0.084 & \underline{0.184} / 0.206 \\
\bottomrule
\end{tabular}
\end{table}

\subsection{The 2D Burgers family}\label{app:burgers}
The family is described in \S\ref{sec:families} (Figure~\ref{fig:familiesnofade}).

\begin{figure}[!htbp]
  \centering
  \includegraphics[width=0.8\linewidth]{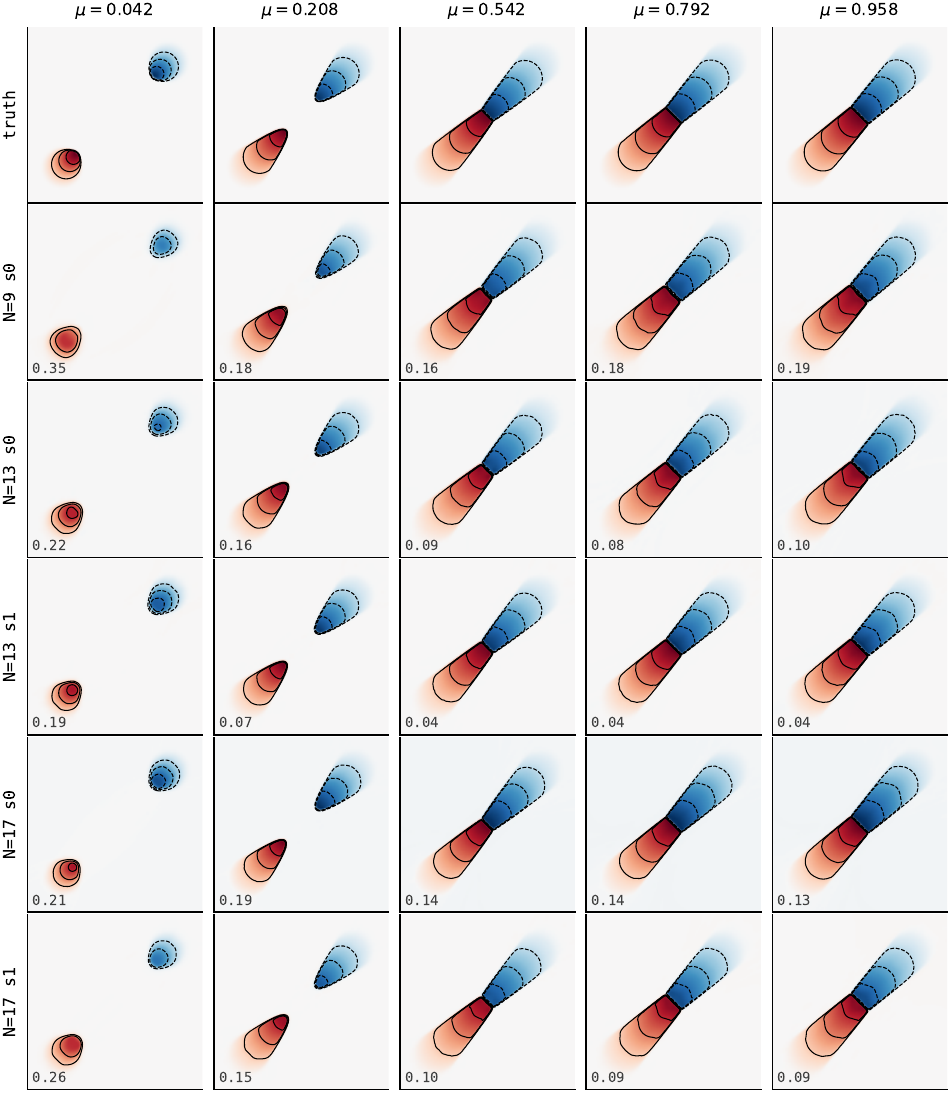}
  \caption{2D Burgers: truth and first-stage LRNR at $N = 9$, 13 and 17 (both seeds at 13 and 17) at five $\mu$ that
  are training values of none of them; every panel is contoured at the truth's levels ($\pm0.25, 0.5, 0.75$ of its
  maximum, negative dashed) and the numbers are relative errors.}
  \label{fig:bg2strip}
\end{figure}

The family was trained and scored like the others (first stage, two seeds, the
recipe of \S\ref{sec:training}) for $N = 2$ to 17, but its $N^\ast$ is an open problem.  From $N = 7$ on (and at $N = 5$ for seed 0) the fit and the
gap pass; the filtered fade ratio stays at 0.58--0.78 and does not reach 0.9 (Table~\ref{tab:bg2sweep}).  At unseen
$\mu$ the models place the humps, the shocks and the standing shock where the truth has them
(Figure~\ref{fig:bg2strip}), with errors of 0.04--0.19 except near $\mu = 0$ (0.19--0.35).  This subsection asks what
the fade ratio measures on this family: how a correct interpolant scores on it, how much of the models' shortfall is
the resolution of the shock, and how much is left when that is taken out.  The pass rule is not changed for this
family.

\begin{table}[!htbp]
\centering\small
\caption{2D Burgers: transition sweep (first-stage LRNR, seeds 0 / 1).  No $N$ passes: from $N=7$ the fit and the
gap pass, but the fade ratio does not.}
\label{tab:bg2sweep}
\begin{tabular}{rcccc}
\toprule
$N$ & train & held-out & gap & fade (filtered) \\
\midrule
2  & 0.181 / 0.310 & 0.997 / 0.868 & 5.52 / 2.80 & 0.67 / 0.68 \\
3  & 0.143 / 0.150 & 0.376 / 0.416 & 2.62 / 2.78 & 0.53 / 0.81 \\
5  & 0.248 / 0.089 & 0.262 / 0.158 & 1.06 / 1.77 & 0.74 / 0.76 \\
7  & 0.116 / 0.140 & 0.136 / 0.150 & 1.17 / 1.08 & 0.73 / 0.78 \\
9  & 0.181 / 0.115 & 0.185 / 0.125 & 1.02 / 1.08 & 0.69 / 0.77 \\
13 & 0.113 / 0.057 & 0.118 / 0.061 & 1.04 / 1.07 & 0.63 / 0.77 \\
17 & 0.157 / 0.124 & 0.158 / 0.124 & 1.01 / 0.99 & 0.72 / 0.58 \\
\bottomrule
\end{tabular}
\end{table}

\subsubsection*{Reference interpolants with exact motion}  We score interpolants that move exactly like the truth but are
not the truth, with the cells, regions and filter of the pass rule: blur $s$, the
truth at every $\mu$ convolved with a Gaussian of $s$ data cells (a smeared shock, exact motion), and lag $d$,
the truth at $\mu + d(1-\mu)$ (exact shape, slightly wrong timing, exact at both ends).  Table~\ref{tab:bg2ref} gives
their fade ratios.  Half a cell of smearing already fails the 0.9 threshold from $N = 9$ on, one cell scores
0.52--0.81, and the test gets harder as $N$ grows; a lag of 1\% of the remaining time scores 0.91--0.97.  On the 2D
split at $N=5$ the same references score 0.99 (one cell of blur) and 0.97 (lag 0.01).

\begin{table}[!htbp]
\centering\small
\caption{2D Burgers: filtered fade ratio of reference interpolants with exact motion (blur $s$: the truth smeared by a
Gaussian of $s$ data cells; lag $d$: the truth at $\mu + d(1-\mu)$), and the share of $\sum d_{\rm truth}^2$ that
comes from the shock neighborhood ($|\nabla u|$ above a quarter of its maximum at that $\mu$).}
\label{tab:bg2ref}
\begin{tabular}{rcccccccc}
\toprule
$N$ & blur $\tfrac12$ & blur 1 & blur 2 & blur 3 & lag 0.005 & lag 0.01 & lag 0.02 & shock share \\
\midrule
5  & 0.933 & 0.814 & 0.656 & 0.548 & 0.985 & 0.969 & 0.944 & 0.15 \\
9  & 0.888 & 0.701 & 0.480 & 0.346 & 0.977 & 0.949 & 0.913 & 0.24 \\
13 & 0.827 & 0.585 & 0.367 & 0.261 & 0.974 & 0.954 & 0.874 & 0.32 \\
17 & 0.809 & 0.523 & 0.266 & 0.168 & 0.956 & 0.913 & 0.867 & 0.39 \\
\bottomrule
\end{tabular}
\end{table}

\subsubsection*{Smeared shocks and the fade ratio}  The truth's departure from its own cross-fade, $d_{\rm truth}$,
has two parts.  The smooth parts of the field (the rarefaction behind each shock and the slow decay of the standing
shock) move by less than their width per gap, so they are close to their own cross-fade, and their contribution to
$d_{\rm truth}$ shrinks as the gaps get shorter.  The shock is a jump: it is displaced by more than its width in any
gap, and its contribution does not shrink.  Its share of $d_{\rm truth}^2$ grows from 0.15 at $N=5$ to 0.39 at $N=17$
(Table~\ref{tab:bg2ref}).  A shock smeared over a cell or two is no longer displaced by more than its width once the
gaps are short, so it departs from its cross-fade less than the sharp one does; as $N$ grows, the fade ratio is decided
more and more by how sharply the shock is resolved, at the scale of the data grid, rather than by whether it moves.

\subsubsection*{The models' resolution}  For each model, $\sigma^\ast$ is the blur at which the smeared truth is closest to
the model at its training snapshots (Table~\ref{tab:bg2models}).  The models resemble the truth smeared by 0.44--0.61
cells, about one cell for the less accurate fits, and smearing accounts for part of their error (0.050--0.225 against
0.057--0.248).  With exact motion at that resolution the fade ratio is 0.72--0.90 (\emph{blur $\sigma^\ast$ alone}), so
a model as sharp as these would not reach 0.9 even if it moved perfectly.  Scored against the smeared truth instead of
the truth, the models' ratios are 0.73--0.84.  Roughly, resolution explains between half and most of the gap to 1, and part of
the gap remains.

\begin{table}[!htbp]
\centering\small
\caption{2D Burgers, first-stage models: filtered fade ratio; $\sigma^\ast$, the blur (data cells) at which the
smeared truth best matches the model at its snapshots, and the model's training error against that smeared truth and
against the truth; the fade ratio of the truth smeared by $\sigma^\ast$ (exact motion); and the model's fade ratio
with that smeared truth as the reference ($d_{\rm model}/d_{{\rm blur}\sigma^\ast}$).}
\label{tab:bg2models}
\begin{tabular}{lcccccc}
\toprule
run & fade & $\sigma^\ast$ & error vs blur $\sigma^\ast$ & error vs truth & blur $\sigma^\ast$ alone & model / blur $\sigma^\ast$ \\
\midrule
$N=5$, s0  & 0.742 & 1.08 & 0.225 & 0.248 & 0.799 & 0.795 \\
$N=5$, s1  & 0.758 & 0.61 & 0.073 & 0.089 & 0.898 & 0.782 \\
$N=9$, s0  & 0.686 & 0.94 & 0.154 & 0.181 & 0.719 & 0.816 \\
$N=9$, s1  & 0.772 & 0.57 & 0.103 & 0.115 & 0.848 & 0.802 \\
$N=13$, s0 & 0.633 & 0.60 & 0.097 & 0.113 & 0.751 & 0.801 \\
$N=13$, s1 & 0.771 & 0.45 & 0.050 & 0.057 & 0.878 & 0.835 \\
$N=17$, s0 & 0.721 & 0.44 & 0.155 & 0.157 & 0.879 & 0.738 \\
$N=17$, s1 & 0.581 & 0.60 & 0.110 & 0.124 & 0.721 & 0.725 \\
\bottomrule
\end{tabular}
\end{table}

\subsubsection*{Equal-blur fade ratio}  To judge motion at a scale that both the model and the truth can represent, we
blur both by the same Gaussian of $s$ data cells ($s = 0$--2, a range that extends above the models' $\sigma^\ast$), and compute the usual filtered
fade ratio of the blurred pair: the cells, the energy filter and the filter $d_{\rm truth}\ge0.1$ are those of the
blurred truth (Table~\ref{tab:bg2equal}).  Two checks come first.  The
exact-motion references of Table~\ref{tab:bg2ref}, blurred alike, score close to 1 once $s$ is large enough: at $s = 2$,
0.98--0.99 (blur $\tfrac12$), 0.91--0.96 (blur 1) and 0.91--0.96 (lag 0.01) at every $N$; at $s = 1$ one cell of
smearing still lowers the ratio by up to 0.2.  On the 2D split at $N = 5$, which passes, the equal-blur ratio moves by at most 0.004
for $s\le2$ (first stage 0.904--0.910, second stage 0.936--0.951), so no verdict there changes.

On Burgers the models' ratios rise with $s$, but less than the references': at $s = 2$ they are 0.74--0.87, and no
run reaches 0.9.  The shortfall that remains lies mostly in the first gap, $\mu\in[0, 1/(N-1)]$, where the humps
steepen into shocks ($t\approx0.3$) and where the models' errors are largest (Figure~\ref{fig:bg2strip}): there every
run scores 0.68--0.87 at every $s$, while the exact-motion references score 0.88--1.02 unblurred and at least 0.95 at
$s=2$.  Over the later gaps, which hold the approach, the collision and the standing shock, six of the eight runs
score 0.95--1.03 at $s = 2$, the other two 0.81 ($N=9$, seed 0) and 0.86 ($N=17$, seed 1).  Blurring also removes
cells: at $s = 2$ the later gaps keep two to four cells with $d_{\rm truth}\ge0.1$, against two to ten unblurred.

\begin{table}[!htbp]
\centering\small
\caption{2D Burgers: equal-blur fade ratio, model and truth blurred by the same Gaussian of $s$ data cells (filtered,
cells of the blurred truth); last two columns: the minimum at $s = 2$ over the first gap and over the later gaps.
Below: exact-motion references (Table~\ref{tab:bg2ref}) in place of the model, at $N = 5$ / 9 / 13 / 17.  2D split,
$N = 5$: 0.904--0.910 (first stage) and 0.936--0.951 (second stage) for $s = 0$--2.}
\label{tab:bg2equal}
\setlength{\tabcolsep}{4pt}
\begin{tabular}{lcccccc}
\toprule
 & $s=0$ & $s=1$ & $s=1.5$ & $s=2$ & first gap & later gaps \\
\midrule
$N=5$, s0  & 0.742 & 0.766 & 0.775 & 0.783 & 0.783 & 1.031 \\
$N=5$, s1  & 0.758 & 0.773 & 0.778 & 0.783 & 0.783 & 0.952 \\
$N=9$, s0  & 0.686 & 0.756 & 0.785 & 0.813 & 0.826 & 0.813 \\
$N=9$, s1  & 0.772 & 0.799 & 0.808 & 0.814 & 0.814 & 0.948 \\
$N=13$, s0 & 0.633 & 0.797 & 0.814 & 0.828 & 0.828 & 0.962 \\
$N=13$, s1 & 0.771 & 0.847 & 0.858 & 0.866 & 0.866 & 0.968 \\
$N=17$, s0 & 0.721 & 0.753 & 0.762 & 0.767 & 0.767 & 0.949 \\
$N=17$, s1 & 0.581 & 0.715 & 0.727 & 0.736 & 0.736 & 0.856 \\
\midrule
blur $\tfrac12$  & .93/.89/.83/.81 & .98/.96/.95/.95 & .99/.97/.97/.97 & .99/.98/.98/.98 & & \\
blur 1          & .81/.70/.59/.52 & .91/.85/.82/.80 & .94/.89/.87/.88 & .96/.92/.92/.91 & & \\
lag 0.01        & .97/.95/.95/.91 & .97/.94/.94/.93 & .96/.94/.94/.92 & .96/.94/.93/.91 & & \\
\bottomrule
\end{tabular}
\end{table}

\subsubsection*{Summary}  On this family the fade ratio mixes two things: whether the shock moves, and how sharply it is
resolved.  Exact motion with half a cell of smearing fails the pass rule from $N=9$ on, and the trained models are
about that sharp.  Blurring model and truth alike removes the resolution effect from the references and from most of
the models' later gaps, but not from the first gap, where every model stays below 0.9.  The ratio does not show whether
that remaining shortfall is fading or another kind of error near shock formation.  Under the pass rule as
stated no $N\le17$ passes, and the family's $N^\ast$ is an open problem.

\subsubsection*{Comparison with the neural baselines}  
\begin{figure}[!htbp]
  \centering
  \includegraphics[width=0.8\linewidth]{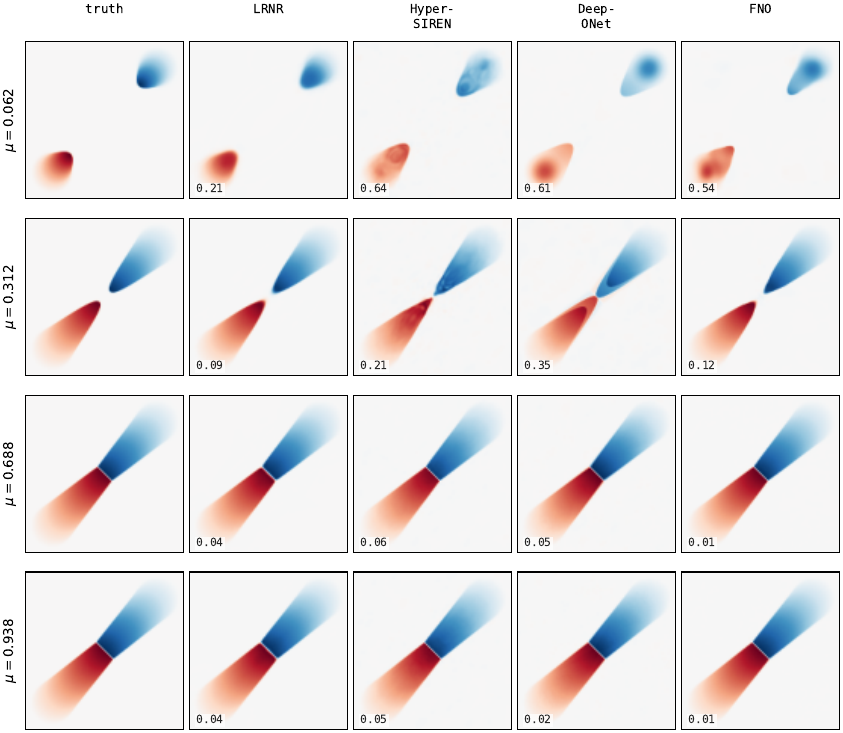}
  \caption{2D Burgers, $N = 9$, at four gap midpoints: truth, final LRNR, Hyper-SIREN, DeepONet and the FNO (seed
  0); numbers are relative errors.}
  \label{fig:bg2cmp9}
\end{figure}

Table~\ref{tab:bg2cmp} compares the LRNR with Hyper-SIREN, DeepONet and the FNO at $N = 9$ and 13
(Figure~\ref{fig:bg2cmp9}: $N = 9$).  For this comparison
the LRNR was trained in both stages (\S\ref{sec:l1}); the analysis above uses first-stage models.  At both $N$ the
final LRNR has the lowest mean and worst held-out error: at $N = 9$, 0.069 / 0.064 and 0.24 / 0.24 against
0.10--0.17 and 0.61--0.65 for the baselines; at $N = 13$, 0.051 / 0.041 and 0.18 / 0.17 against 0.08--0.16 and
0.50--0.55.  Every model has its largest error at $\mu = 0.02$--0.08, while the shocks form; there the LRNR's error is
about a third of the baselines' (Figure~\ref{fig:bg2cmp9}, $\mu = 0.062$: 0.21 against 0.54--0.64), and Hyper-SIREN
and DeepONet show visible defects.  After the collision all models place the standing shock correctly; the FNO's
error is then 0.01, against the LRNR's 0.03--0.04.

\begin{table}[!htbp]
\centering\small
\caption{2D Burgers: training error and mean and worst held-out relative error (seed 0 / seed 1).  LRNR: first stage
and final model.}
\label{tab:bg2cmp}
\begin{tabular}{llccc}
\toprule
$N$ & model & train & held-out (mean) & held-out (max) \\
\midrule
9  & LRNR, stage 1 & 0.181 / 0.115 & 0.185 / 0.125 & 0.346 / 0.276 \\
   & LRNR, final   & 0.050 / 0.045 & \underline{0.069 / 0.064} & \underline{0.243 / 0.240} \\
   & Hyper-SIREN   & 0.041 / 0.028 & 0.156 / 0.142 & 0.645 / 0.619 \\
   & DeepONet      & 0.014 / 0.025 & 0.170 / 0.174 & 0.615 / 0.618 \\
   & FNO           & 0.007 / 0.009 & 0.103 / 0.122 & 0.606 / 0.627 \\
\midrule
13 & LRNR, stage 1 & 0.113 / 0.057 & 0.118 / 0.061 & 0.222 / 0.191 \\
   & LRNR, final   & 0.043 / 0.032 & \underline{0.051 / 0.041} & \underline{0.182 / 0.168} \\
   & Hyper-SIREN   & 0.024 / 0.074 & 0.106 / 0.137 & 0.532 / 0.504 \\
   & DeepONet      & 0.037 / 0.069 & 0.144 / 0.160 & 0.510 / 0.511 \\
   & FNO           & 0.009 / 0.010 & 0.081 / 0.089 & 0.548 / 0.505 \\
\bottomrule
\end{tabular}
\end{table}

\section{Fine-frame comparisons}\label{app:movies}

The figures of the main text show a few values of $\mu$.  This appendix shows each comparison of
\S\ref{sec:baselines} at fine resolution in $\mu$ (seed 0); for the 1D and 2D
split \emph{LRNR} is the final model, shown next to the first-stage model (\emph{LRNR stage 1}); for the spiral at $N=9$ the
dashed \emph{LRNR N=25} is the final model at $N^\ast=25$.  In 1D we show
a 3D surface of $u(x,\mu)$ on 241 values of $\mu$ for the truth and every model, and a waterfall of the error
$u - u_{\rm truth}$ at four values of $\mu$ per gap plus the snapshots.  In 2D we show frame grids at four frames per gap, at
fractions $\tfrac18, \tfrac38, \tfrac58, \tfrac78$ of every gap, preceded by a plot that shows the
whole evolution at once: the half-maximum contour $u(\cdot,\mu) = \tfrac12 \max u_{\rm truth}(\cdot,\mu)$ every
$\Delta\mu = 0.025$, each drawn in its own $xy$ slice and stacked along a downward $\mu$ axis.  The contours of a
displaced field form one continuous stack; those of a cross-fade, whose value drops below half the maximum in
mid-gap, form separate clusters around the snapshots, with no contour in between.  The scans of error and amplitude on the fine grid,
and the frames chosen by the rule of \S\ref{sec:baselines}, are given for every family.

\subsubsection*{1D oscillatory split, $N = 13$} (scan and rule-chosen frames: Figure~\ref{fig:between}).  Hyper-SIREN's
crests pulse between the snapshots (the amplitude collapses in mid-gap and recovers at each snapshot), whereas the
truth's and the LRNR's crests are smooth ridges.  The error waterfall shows the baselines' errors throughout
$\mu < 0.65$ and the LRNR's confined to $\mu < 0.25$.

\begin{figure}[!htbp]
  \centering
  \includegraphics[width=0.92\linewidth]{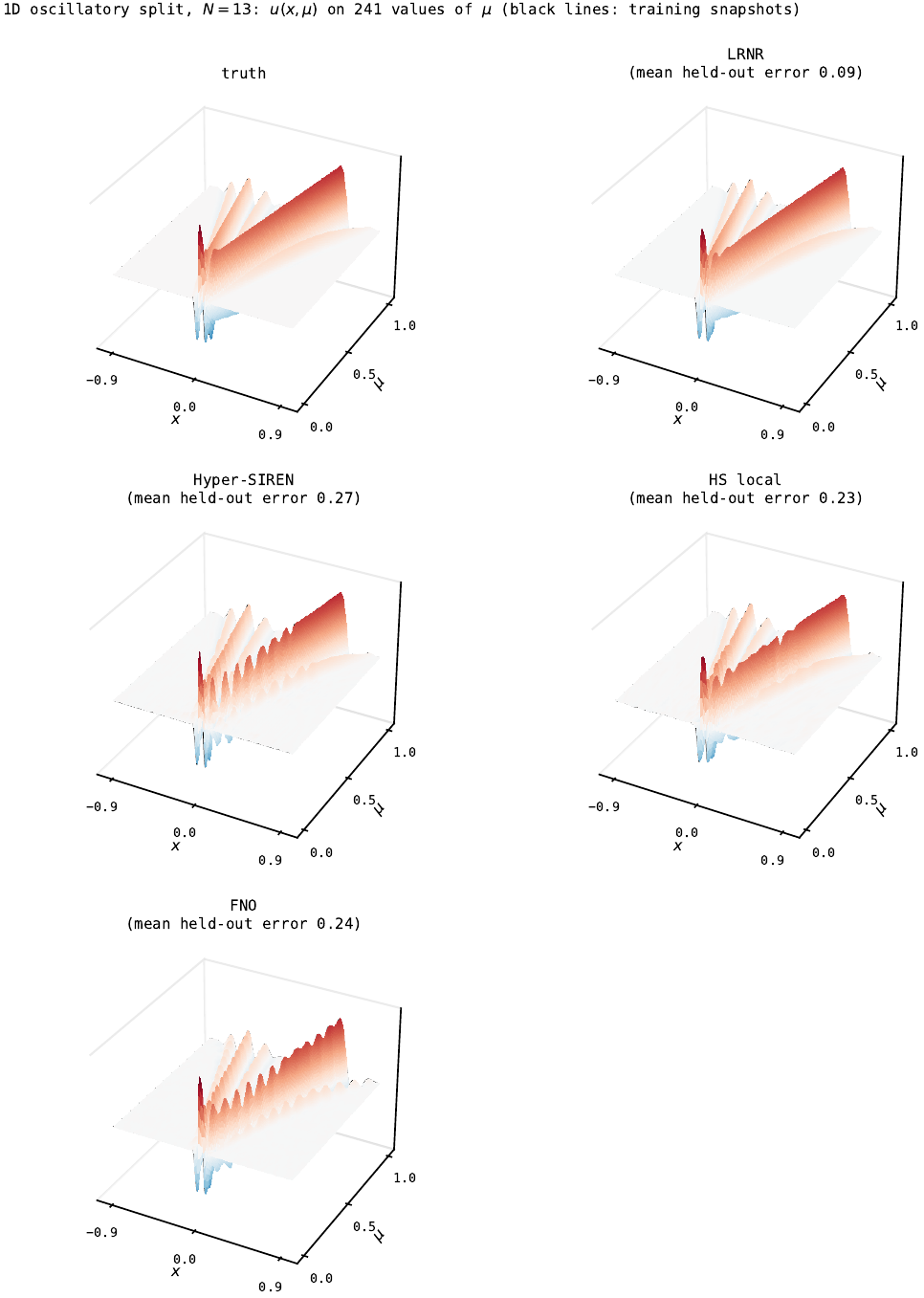}
  \caption{1D oscillatory split, $N=13$: $u(x,\mu)$ on 241 values of $\mu$.}
  \label{fig:surf_os1}
\end{figure}

\begin{figure}[!htbp]
  \centering
  \includegraphics[width=\linewidth]{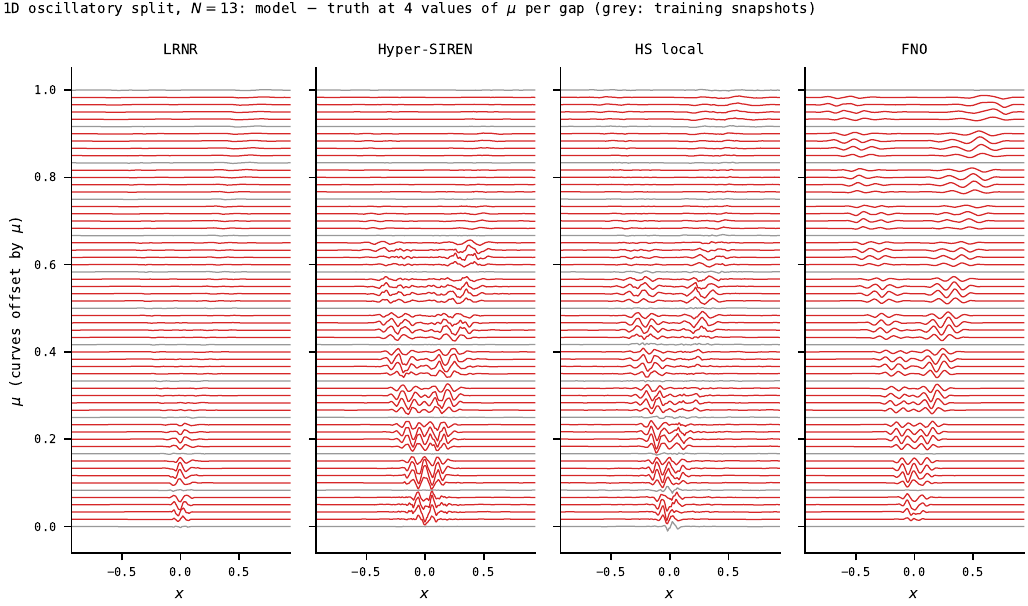}
  \caption{1D oscillatory split, $N=13$: error $u - u_{\rm truth}$ at four values of $\mu$ per gap (red) and at the
  training snapshots (gray), offset by $\mu$; same vertical scale for all models.}
  \label{fig:errfall_os1}
\end{figure}

\subsubsection*{1D split, $N = 5$}  
\begin{figure}[!htbp]
  \centering
  \includegraphics[width=0.85\linewidth]{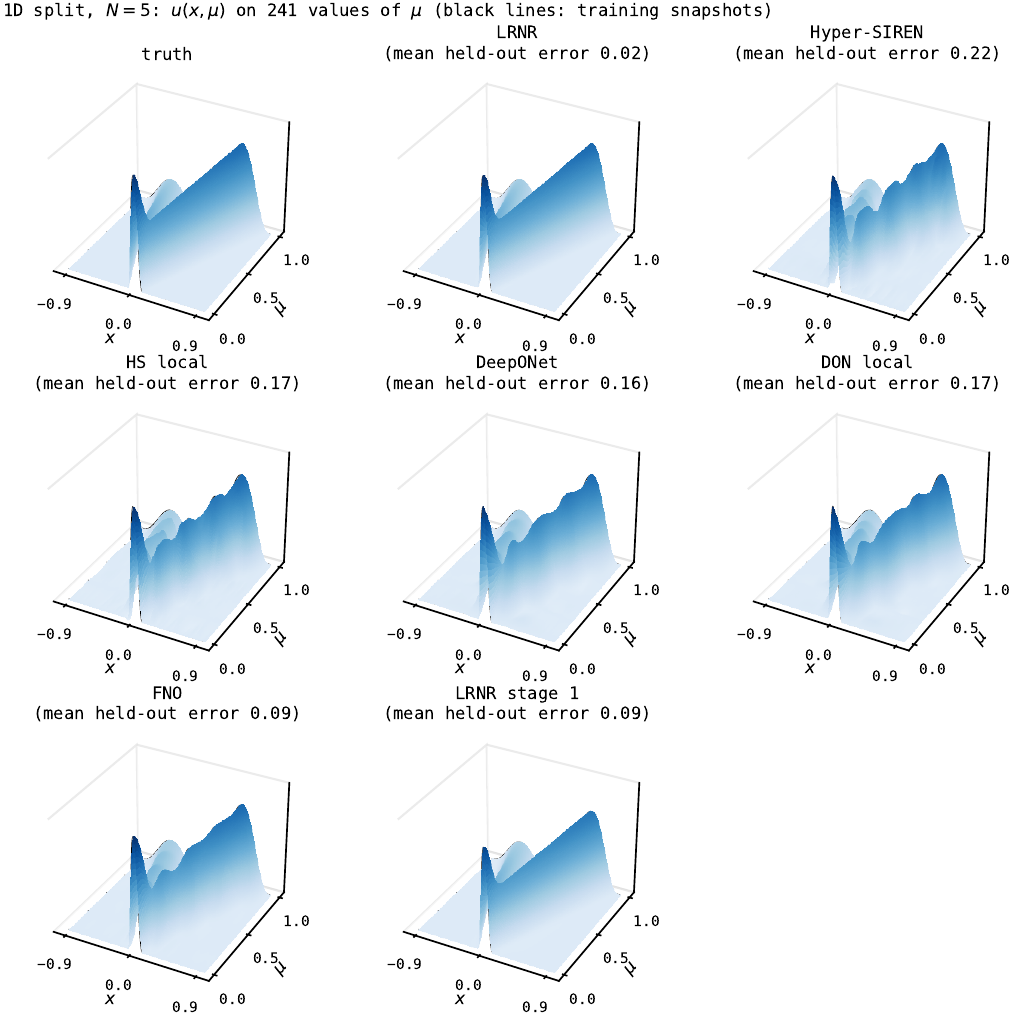}
  \caption{1D split, $N=5$: $u(x,\mu)$ on 241 values of $\mu$.}
  \label{fig:surf_sp1}
\end{figure}

\begin{figure}[!htbp]
  \centering
  \includegraphics[width=\linewidth]{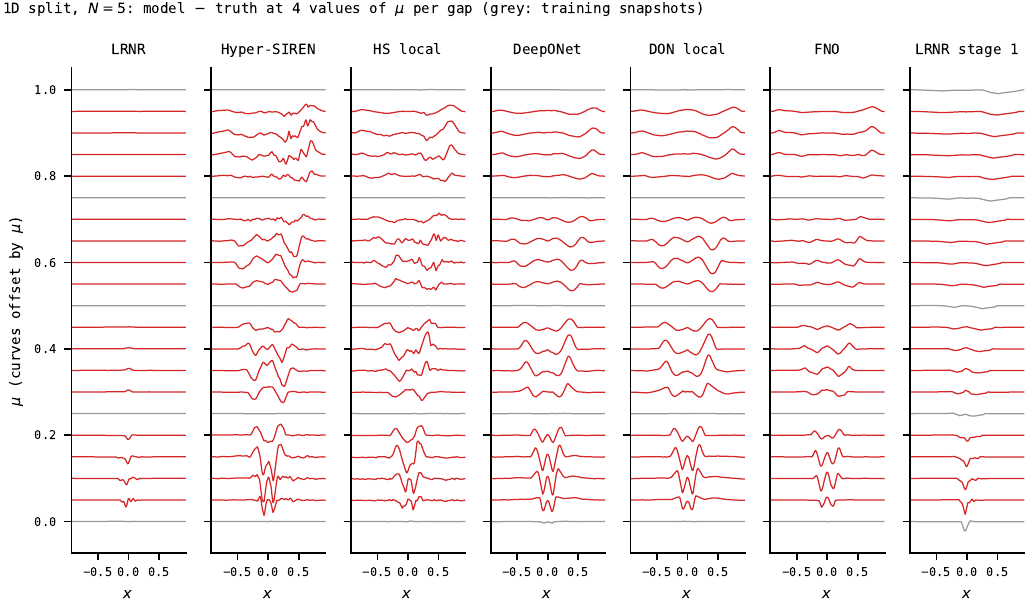}
  \caption{1D split, $N=5$: error waterfall, as in Figure~\ref{fig:errfall_os1}.}
  \label{fig:errfall_sp1}
\end{figure}

All five baselines in Figures~\ref{fig:surf_sp1} and \ref{fig:errfall_sp1} show the same pattern: exact at the snapshots, with a dip and
reshaping of the copies in every gap (bumps along the ridges of the surface plot); the LRNR's ridges are smooth and its
error is small and even in $\mu$, and the final model's is smaller still and also small at the snapshots.

\begin{figure}[!htbp]
  \centering
  \includegraphics[width=\linewidth]{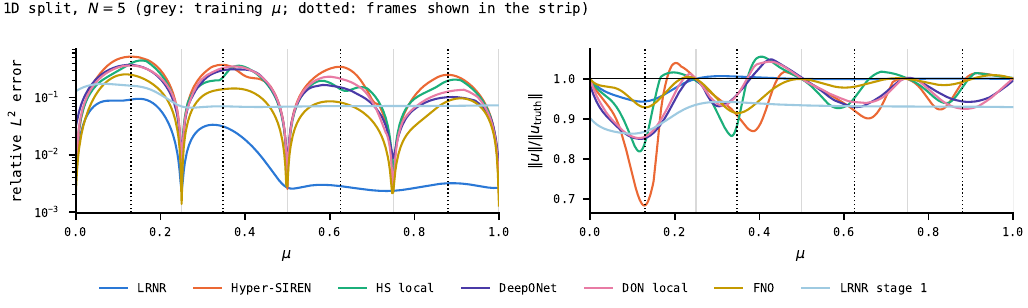}\\[0.3em]
  \includegraphics[width=\linewidth]{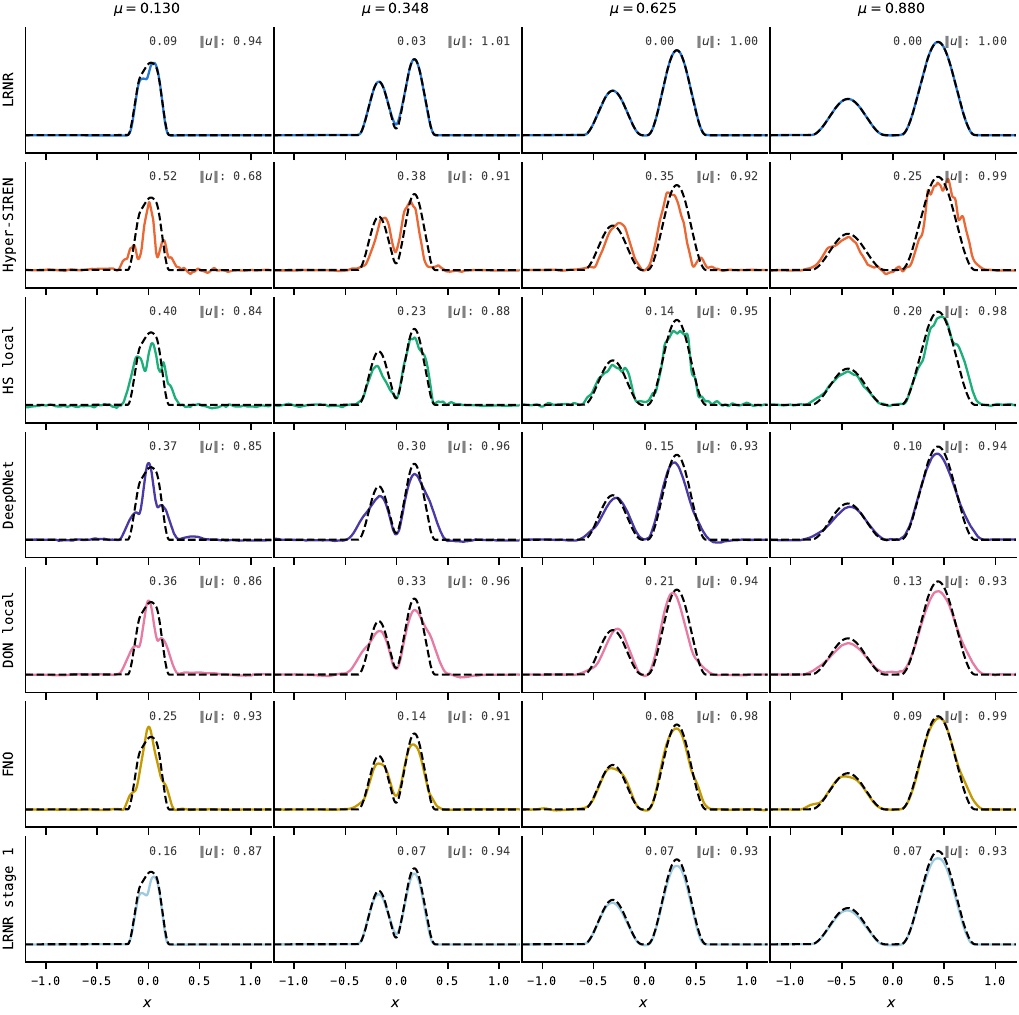}
  \caption{1D split, $N=5$: scan of error and amplitude, and the rule-chosen frames.}
  \label{fig:scan_sp1}
\end{figure}

\subsubsection*{2D split, $N = 5$}  
\begin{figure}[!htbp]
  \centering
  \includegraphics[width=\linewidth]{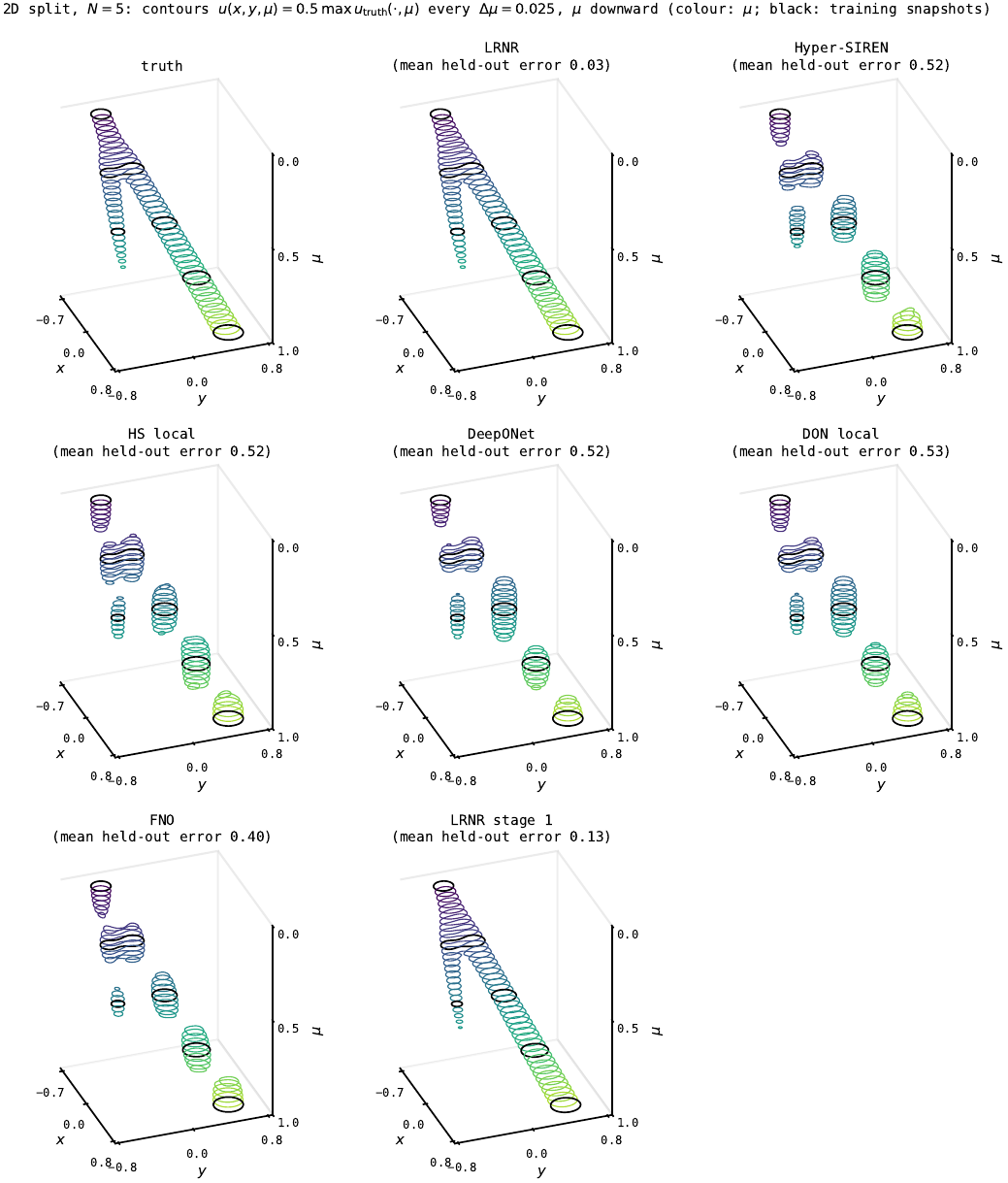}
  \caption{2D split, $N=5$: half-maximum contours every $\Delta\mu = 0.025$, stacked along a downward $\mu$ axis, each
  frame scaled by the truth's maximum at that $\mu$; color indicates $\mu$, black curves mark the training snapshots.}
  \label{fig:iso_sp2}
\end{figure}

\begin{figure}[!htbp]
  \centering
  \includegraphics[width=\linewidth]{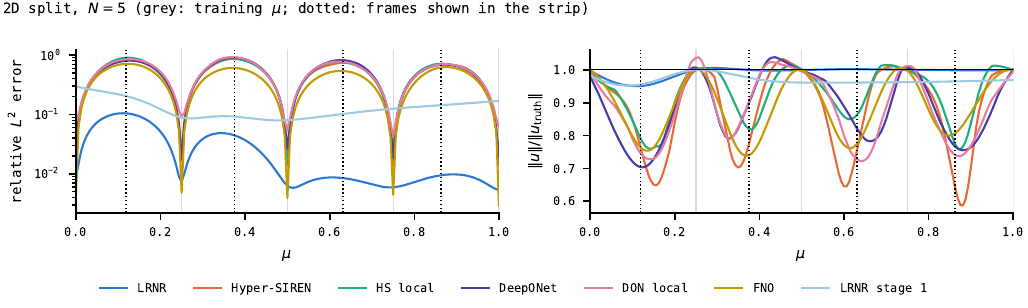}\\[0.3em]
  \includegraphics[height=0.6\textheight]{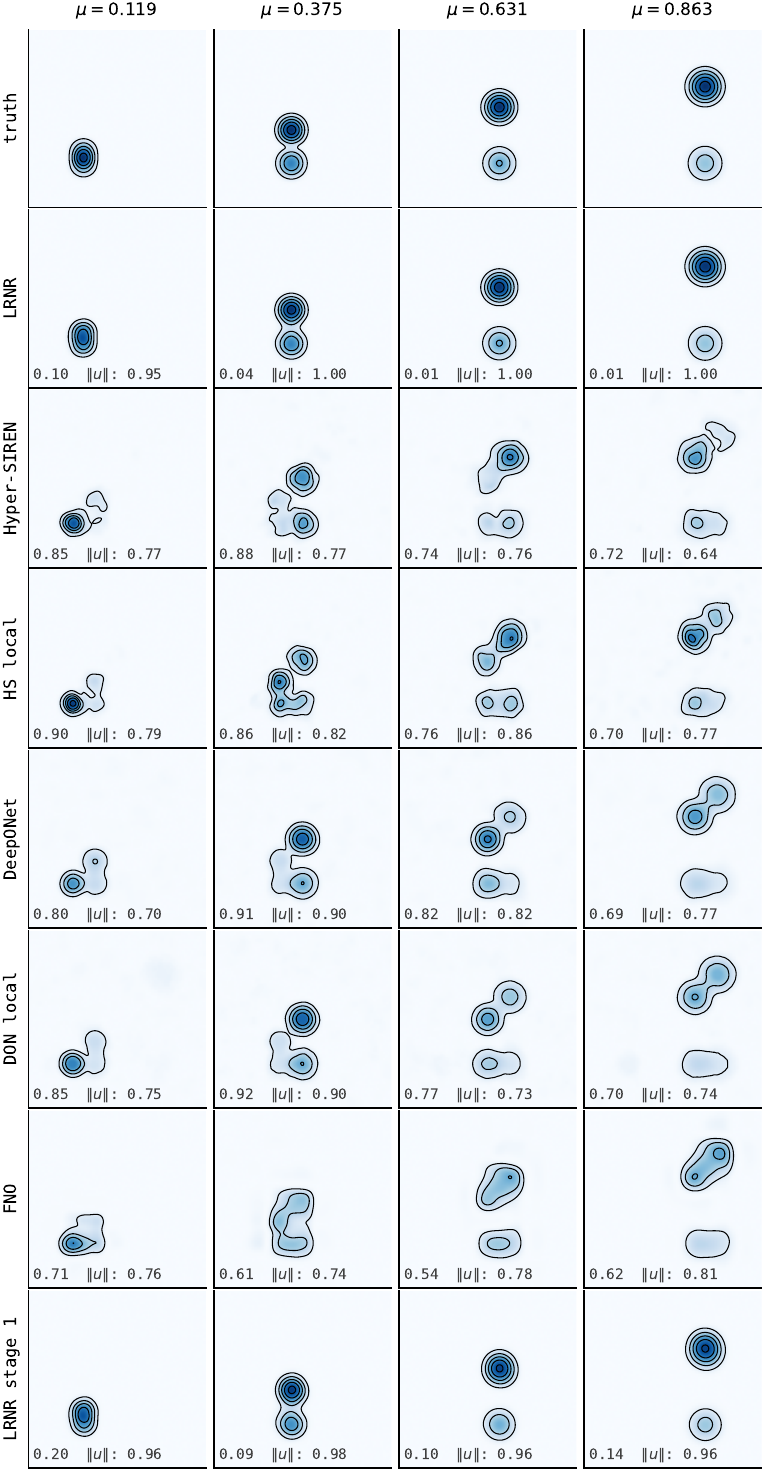}
  \caption{2D split, $N=5$: scan of error and amplitude, and the rule-chosen frames.}
  \label{fig:scan_sp2}
\end{figure}

\begin{figure}[!htbp]
  \centering
  \includegraphics[width=\linewidth]{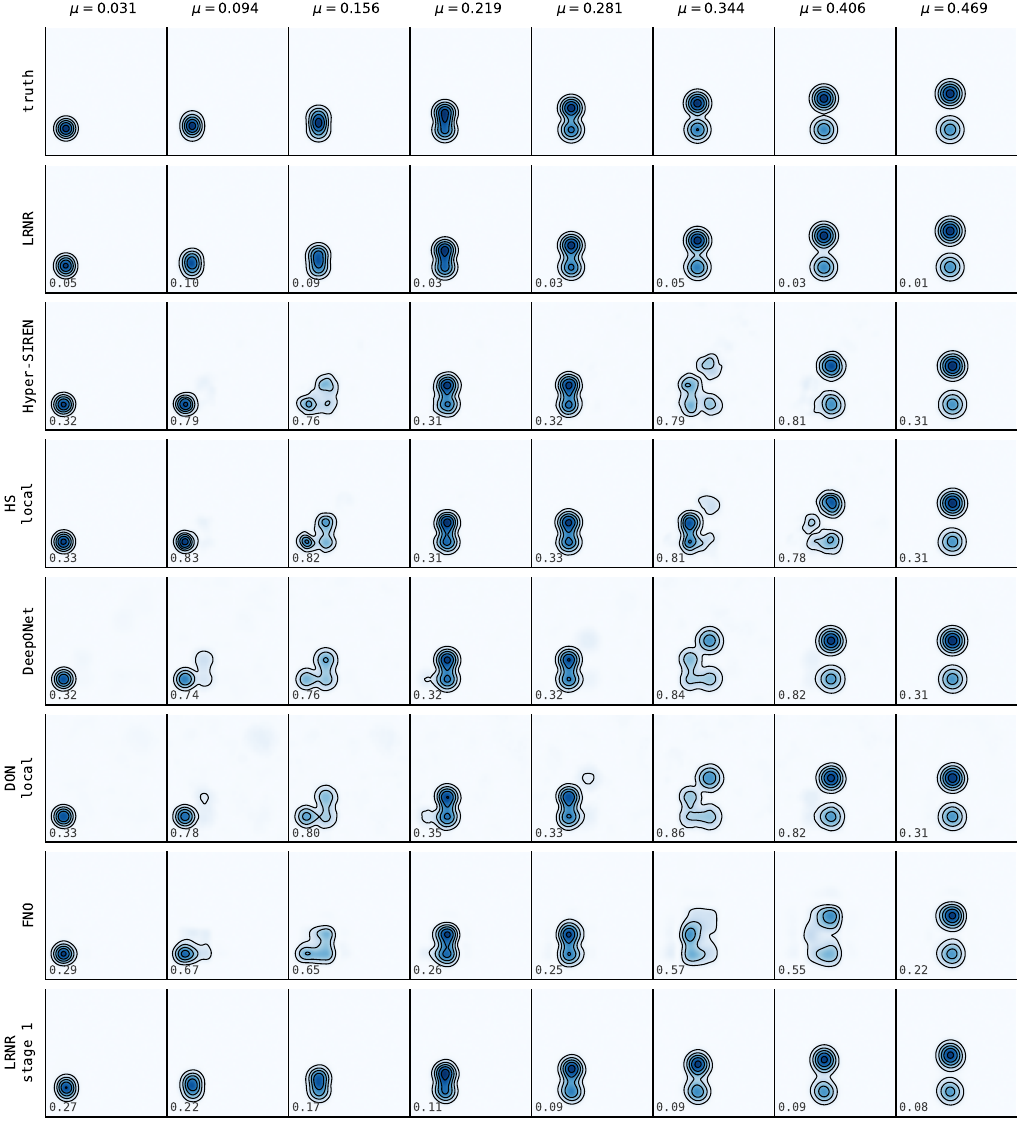}
  \caption{2D split, $N=5$, $\mu\in[0, 0.5]$: four frames per gap (relative error in each panel), contoured at the
  truth's levels.}
  \label{fig:movie_sp2}
\end{figure}

\begin{figure}[!htbp]
  \centering
  \includegraphics[width=\linewidth]{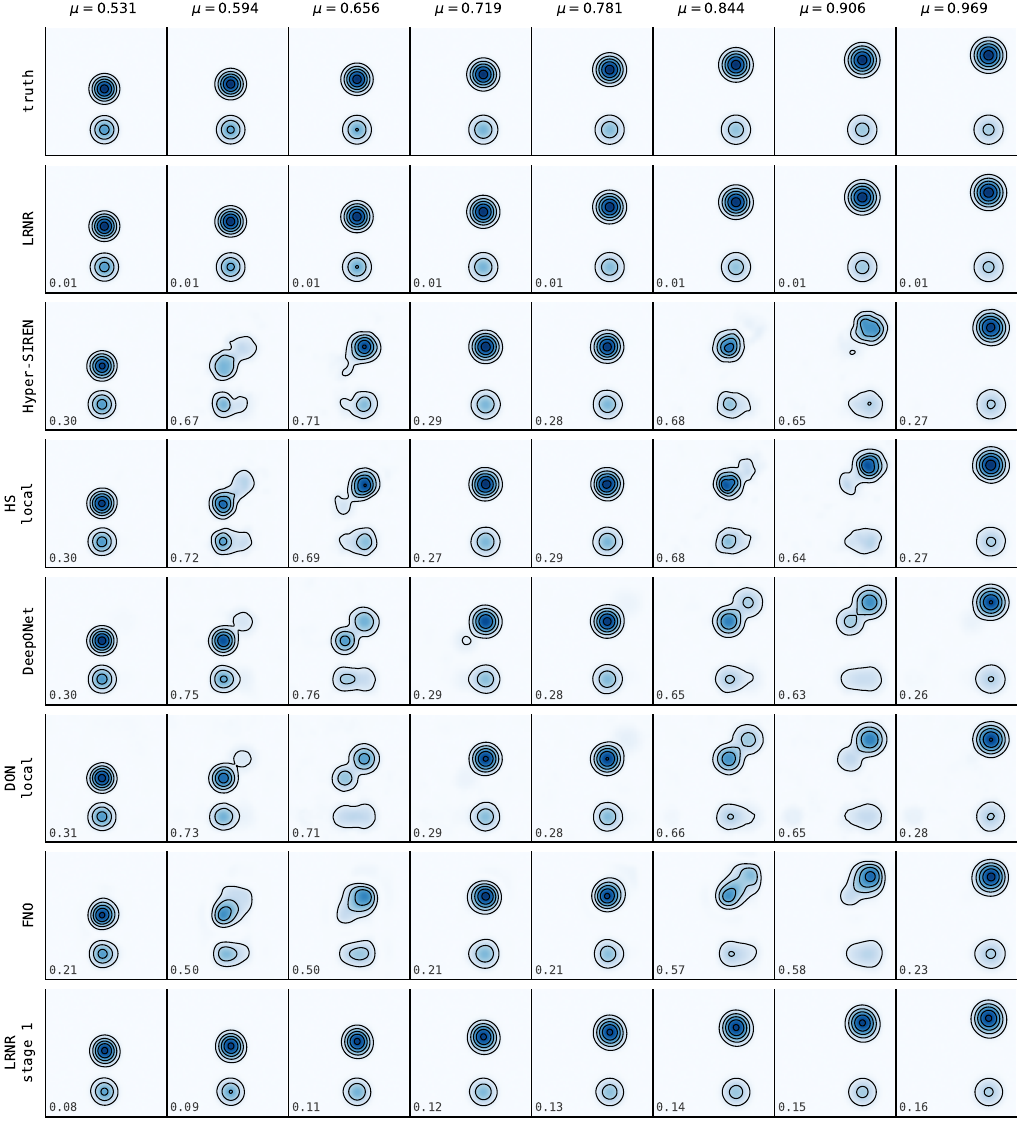}
  \caption{2D split, $N=5$, $\mu\in[0.5, 1]$, as in Figure~\ref{fig:movie_sp2}.}
  \label{fig:movie_sp2b}
\end{figure}

In every gap the baselines' copies break up into a fading copy at the old position and a
growing one at the new, joined by a bridge, and re-form at the next snapshot; the LRNR's copies move (Figures~\ref{fig:scan_sp2}--\ref{fig:movie_sp2b}).  The stacked contours (Figure~\ref{fig:iso_sp2}) show this for all of $\mu$ at once: the truth's and the
LRNR's form one continuous stack that branches into the copies, while every baseline's break into separate clusters,
one around each snapshot.

\subsubsection*{2D spiral, $N = 9$}  
\begin{figure}[!htbp]
  \centering
  \includegraphics[width=\linewidth]{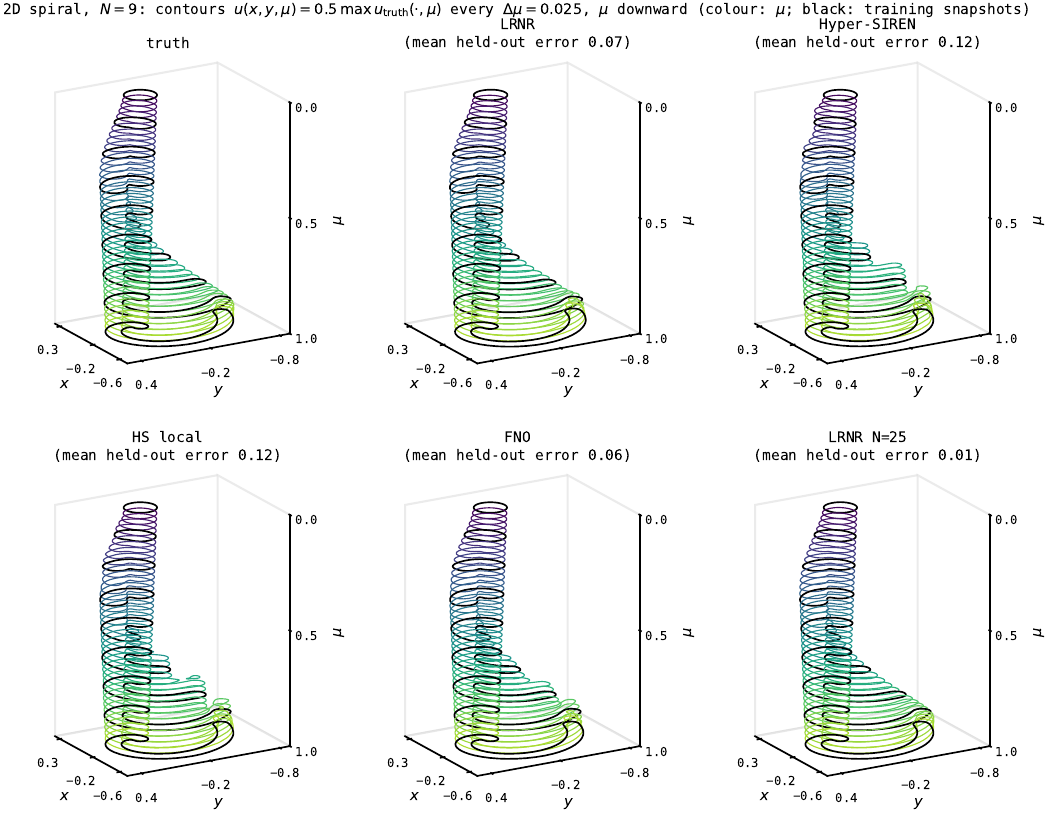}
  \caption{2D spiral, $N=9$ (and the LRNR at $N=25$): stacked half-maximum contours, as in Figure~\ref{fig:iso_sp2}.}
  \label{fig:iso_spr}
\end{figure}

\begin{figure}[!htbp]
  \centering
  \includegraphics[width=\linewidth]{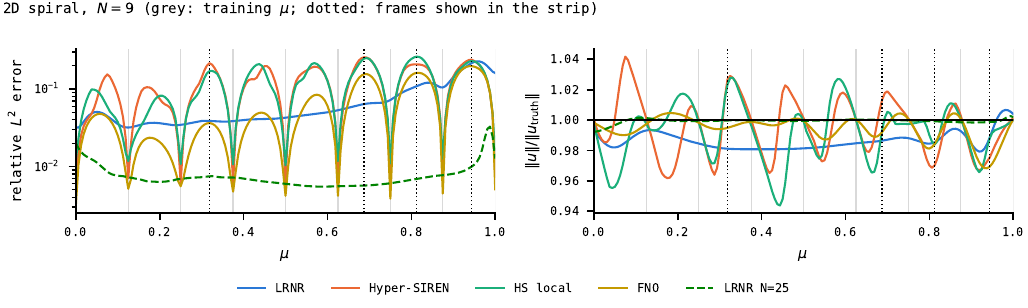}\\[0.3em]
  \includegraphics[width=0.6\linewidth]{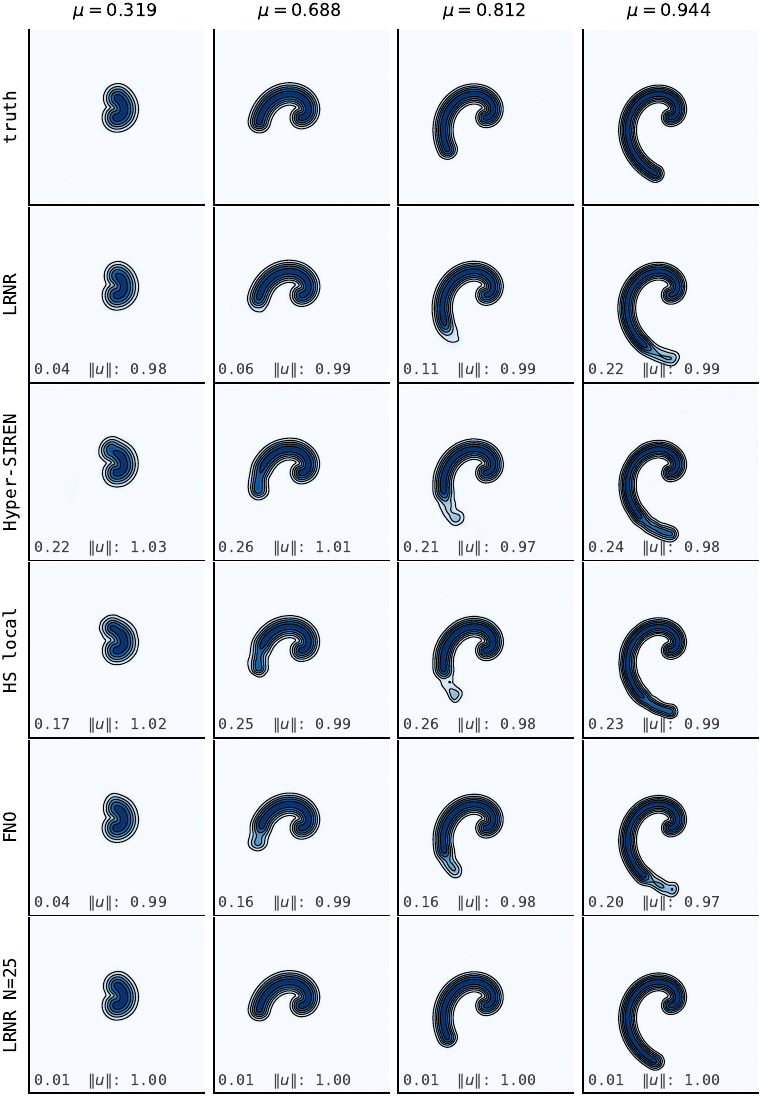}
  \caption{2D spiral, $N=9$ (and the LRNR at $N=25$, dashed): scan of error and amplitude, and the rule-chosen frames.}
  \label{fig:scan_spr}
\end{figure}

\begin{figure}[!htbp]
  \centering
  \includegraphics[height=0.85\textheight]{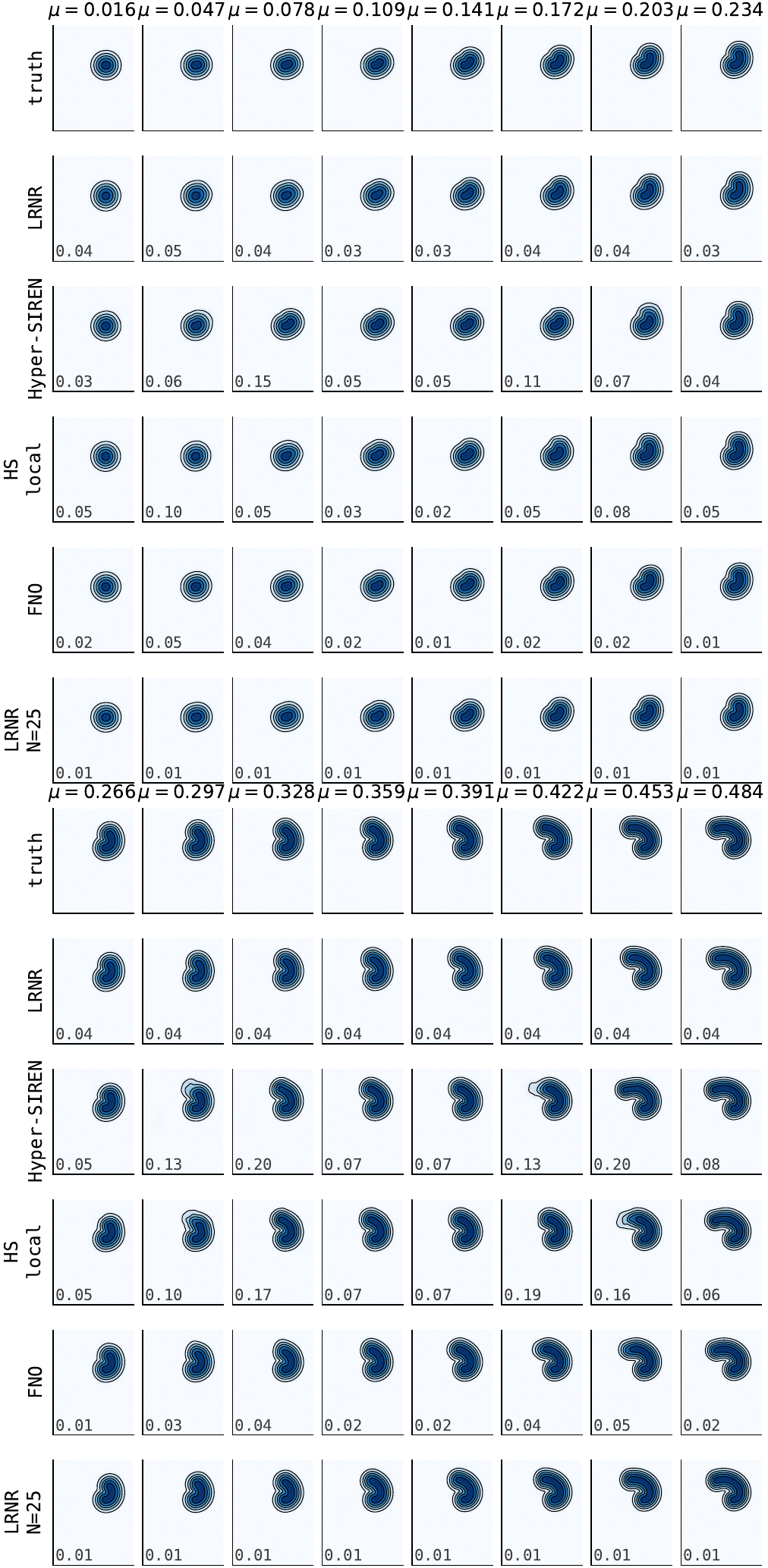}
  \caption{2D spiral, $N=9$, $\mu\in[0, 0.5]$: four frames per gap.}
  \label{fig:movie_spr1}
\end{figure}

\begin{figure}[!htbp]
  \centering
  \includegraphics[height=0.85\textheight]{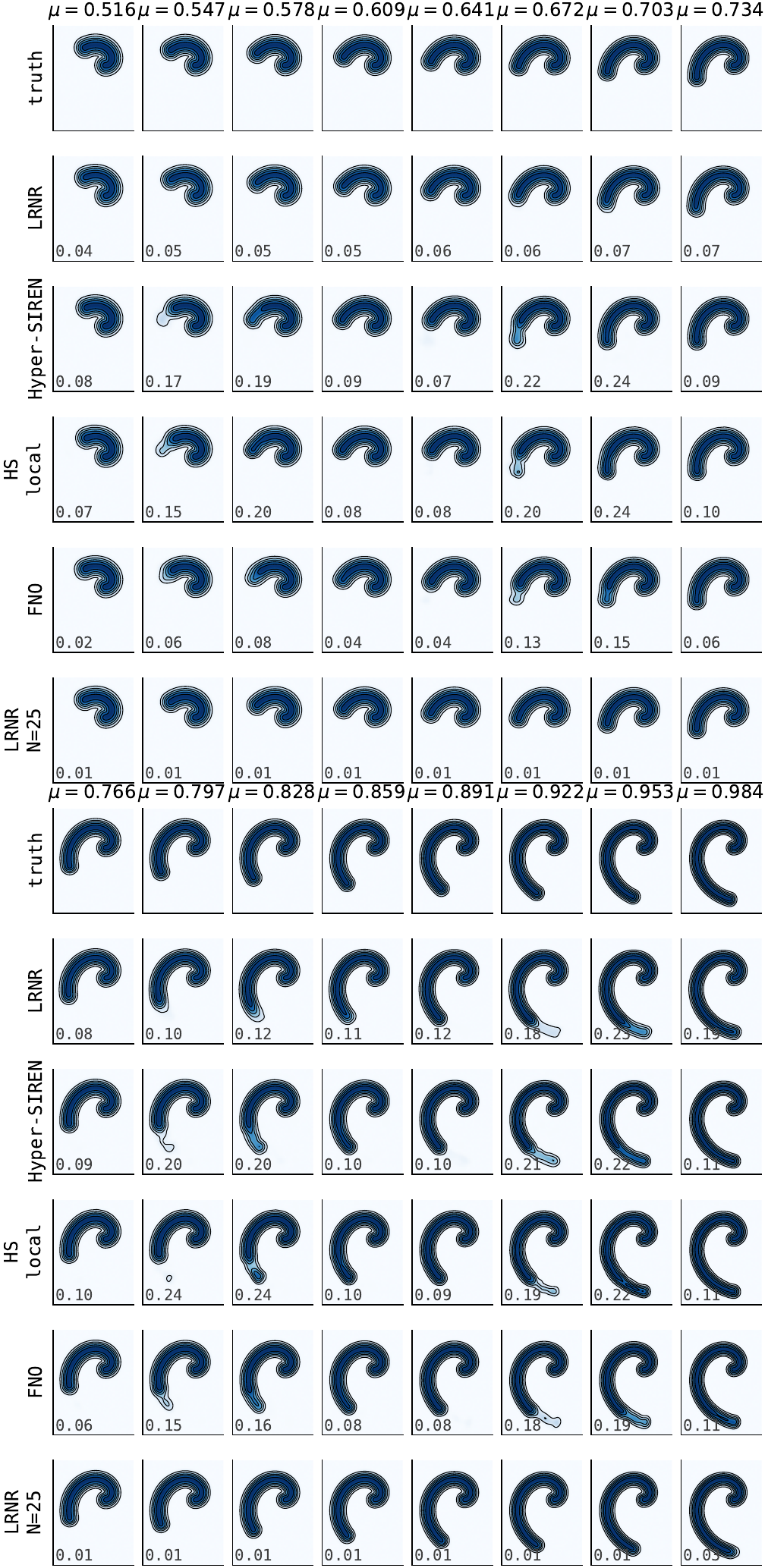}
  \caption{2D spiral, $N=9$, $\mu\in[0.5, 1]$: four frames per gap.}
  \label{fig:movie_spr2}
\end{figure}

Here no model passes; Hyper-SIREN's error peaks in every gap and the LRNR's grows
toward the tail ($\mu > 0.8$), where both cross-fade the end of the arc (Figures~\ref{fig:scan_spr}, \ref{fig:movie_spr1} and~\ref{fig:movie_spr2}); the final LRNR at $N = 25$ is shown for reference.  In the stacked contours (Figure~\ref{fig:iso_spr}) the truth's curves wind smoothly into the spiral arc,
and so do the LRNR's, while Hyper-SIREN's and local Hyper-SIREN's are ragged along the inner edge of the arc, where in every
gap the contour falls back toward the cross-fade.

\section{Classical baselines: details and full results}\label{app:classical}

\subsubsection*{Settings}  Registration: displacement on a $16\times16$ control grid (64 points in 1D) upsampled by cubic
interpolation, objective $\|u_i(x-d/2) - u_{i+1}(x+d/2)\|^2/\|u\|^2 + \alpha\|\nabla d\|^2$ minimized by Adam over
five Gaussian blur levels, $\alpha\in\{10^{-3},10^{-2},10^{-1},1\}$.  sPOD with fitted shifts: modes, amplitudes and
per-snapshot shifts fitted jointly by Adam over the same blur levels, three random starts, (frames, rank per frame)
$(n_f, r)\in\{(1,1),(1,2),(2,1),(2,2),(3,1),(3,2)\}$ with $n_f r \le \max(1, N-1)$.  sPOD with provided shifts: FISTA~\cite{beck2009fista} on
$\tfrac12\|\bQ - \sum_k \mathbf{T}_k \bQ_k - \mathbf{E}\|^2 + \lambda\sum_k\|\bQ_k\|_* + \lambda_E\|\mathbf{E}\|_1$, $\lambda = \beta\|\bQ\|_2$ with
$\beta\in\{10^{-3},10^{-2},10^{-1}\}$, the sparse term on or off, 400 iterations; the shifts are those of
\S\ref{sec:classical}.  OT: exact quantile interpolation in 1D; in 2D a log-domain Sinkhorn plan with
$\varepsilon$-scaling down to $\varepsilon = h^2$ on the supports, the displacement interpolant deposited bilinearly.
DMD: exact DMD~\cite{tu2014dmd} on the uniformly spaced snapshots, rank $r\in\{1,2,4,8,16,N-1\}$ with $r\le N-1$,
amplitudes by least squares over all $N$ snapshots~\cite{jovanovic2014sparsity}.  At $N^\ast$ the chosen rank is 1--8;
the full-rank variant, its eigenvalues fitted to snapshots that are not linearly related, grows or oscillates between
the snapshots (held-out error 0.6 to 444) and is chosen only on the fade control and the 2D wave ($N = 3$ and 5).  On the fade control, $u = (1-\mu)u_0 + \mu u_1$ is a
defective linear system (a double eigenvalue 1 with a Jordan block), which DMD's sum of exponentials represents only
in the limit of two nearly equal eigenvalues with large canceling amplitudes; whether the fit reaches this limit depends on
round-off (exact at $N = 3$ and $N = 7$--25, held-out error 0.45 at $N = 5$).
For each method, example and $N$ the setting with the lowest held-out error is reported.

\subsubsection*{sPOD with provided shifts on the 1D oscillatory split}  As throughout, this variant is an oracle rather
than a competitor (\S\ref{sec:classical}): it receives the shift of each frame at every $\mu$, held-out ones
included, which is most of the transport the LRNR has to learn.  On this family sPOD is given the exact shifts
$c(\mu) = (-\mu/2, +\mu/2)$ of the two copies, yet at $N = 9$ its held-out error (0.32, Table~\ref{tab:minimaxosc}) is
four to five times the LRNR's.  The error does not come from the shifts or from the linear interpolation in the co-moving frames.
Each copy is a carrier under a cosine hump in its own co-moving coordinate $\xi = x - c_k(\mu)$; across $\mu\in[0,1]$
the hump widens from half-width 0.1 to 0.4 and the carrier's wavenumber $k$ falls from 20 to 10 (left copy) or 5 (right copy),
so the frame content is far from constant.  Even so, applying sPOD's own prediction rule to the true
copies (each interpolated linearly between neighboring snapshots in its co-moving coordinate and shifted exactly)
gives a held-out error of 0.039 at $N=9$ (Table~\ref{tab:spoddiag}), below the LRNR's.  The error lies in the
split into frames.  A snapshot constrains only the sum $\sum_k T_{c_k(\mu_j)} q_k(\cdot;\mu_j)$, so any piece can be
moved from one frame into the other, shifted, without changing the fit; the copies overlap for every $\mu < 0.5$
(centers $\mu$ apart, half-widths $0.1 + 0.3\mu$), and at $\mu = 0$, where both shifts vanish, the split is
undetermined.  The nuclear norm chooses among these splits, and it does not choose the true one: at $N = 9$ the fitted
frames reproduce every snapshot to within 2.1\%, but each frame differs from its true copy by 18--61\% at the interior
snapshots, and at $\mu = 0.5$--0.75 the left frame contains only 56--58\% of the left copy's norm.  The defect lies in the
convex model rather than in its solution.  The true split has a larger objective than the fitted one (3.16 against
2.75), so it is not the minimizer, and a better solver would not recover it.  The true frames are high-rank in $\mu$ (numerical rank 4
and 5 at 99\% of the energy; fitted: 4 and 4), since the chirp and the dilation make their columns nearly orthogonal.  We
posit that the penalty lowers this rank at the price of a wrong assignment.  Between snapshots each misassigned piece moves with the other
frame's shift (at speed $\tfrac12$ in the wrong direction) and appears as a ghost oscillation or a cancellation;
the worst error (0.59) is at the edge of the overlap region.

The defect shrinks as snapshots are added (Table~\ref{tab:spoddiag}): the true split still has the larger objective
at every $N$, but with closely spaced snapshots the frames' content changes less between neighbors and the wrong
assignment raises the objective less.  At $N = 17$ sPOD reaches 0.050, about the LRNR's error at $N^\ast = 9$, and at $N = 25$
0.016.  Its frames remain wrong at the ends: at $\mu = 1$, where the carriers are slowest, the left frame's error is
0.6--0.9 for $N\ge13$, and at the first snapshot after $\mu = 0$, inside the overlap, 0.27--0.41 for $N\ge17$.  This is consistent with the rings and the wave packet,
where sPOD with provided shifts is the most accurate or close to it with 17--21 snapshots
(Table~\ref{tab:minimaxosc}).  The \emph{true frames} column is a further oracle.  It needs the copies themselves in addition to their shifts, and it
serves as a diagnostic rather than as a method.  Knowing the shifts is therefore not sufficient for sPOD when the copies
overlap and change shape.  sPOD also needs frames that are identifiable from the snapshots, which here takes about
twice the LRNR's $N^\ast$.

\begin{table}[!htbp]
\centering\small
\caption{sPOD with provided shifts on the 1D oscillatory split: held-out error over $\mu\ge0.1$ (mean / worst, 11
held-out $\mu$ per gap) of sPOD as fitted and of the same prediction rule applied to the true copies (\emph{true frames}); at
the training snapshots, the relative error of the fitted frames against the true copies (range over the interior
snapshots and both frames), the largest relative misfit of the frames' sum, and the convex objective
$\tfrac12\|\bQ - \sum_k \mathbf{T}_k \bQ_k\|^2 + \lambda\sum_k\|\bQ_k\|_*$ of the true and of the fitted split ($\lambda$ as fitted: $\beta = 10^{-2}$ at $N=9$,
$10^{-3}$ otherwise, so objectives compare within a row only).  LRNR at $N^\ast=9$:
0.062 / 0.083 (seed 0 / seed 1).}
\label{tab:spoddiag}
\begin{tabular}{rcccccc}
\toprule
$N$ & true frames & sPOD & frame error & sum misfit & objective, true & objective, fitted \\
\midrule
9 & 0.039 / 0.083 & 0.326 / 0.585 & 0.18--0.61 & 0.021 & 3.16 & 2.75 \\
13 & 0.018 / 0.039 & 0.127 / 0.318 & 0.08--0.33 & 0.003 & 0.447 & 0.424 \\
17 & 0.010 / 0.022 & 0.050 / 0.179 & 0.04--0.37 & 0.005 & 0.583 & 0.563 \\
25 & 0.004 / 0.010 & 0.016 / 0.088 & 0.02--0.44 & 0.007 & 0.853 & 0.827 \\
\bottomrule
\end{tabular}
\end{table}

\section{Compute}
All runs used one shared 24 GB GPU, typically 5--8 LRNR trainings at once alongside unrelated jobs.  A LRNR step
takes about 10--12 ms alone and 0.08--0.3 s under that load; a 2D run at $N=5$ took 5--9 hours.  Hyper-SIREN and
DeepONet runs took 3--20 minutes alone, FNO runs 10--60 minutes on an A100, CAE runs 15 minutes to 4 hours on four
CPU cores of an LSF cluster.  The oscillatory families above $N = 13$, their second stages, and the
baselines at every $N^\ast$ ran on A100 and H100 GPUs of a SLURM cluster, two trainings per GPU.

\clearpage
\end{document}